\documentclass[11pt]{article}

\usepackage[letterpaper,top=0.74in,bottom=0.76in,left=0.80in,right=0.80in,includeheadfoot]{geometry}
\usepackage[T1]{fontenc}
\usepackage[utf8]{inputenc}
\usepackage{amssymb}
\usepackage{newtxtext,newtxmath}
\usepackage[scaled=0.92]{helvet}
\usepackage{microtype}
\usepackage{xcolor}
\usepackage[round,authoryear]{natbib}

\usepackage{hyperref}
\usepackage{url}
\usepackage{graphicx}
\usepackage{tabularx}
\usepackage{array}
\usepackage{booktabs}    
\usepackage{amsmath} 
\usepackage{multirow}
\usepackage{enumitem}
\usepackage{caption}
\usepackage{titlesec}
\usepackage{titling}
\usepackage{abstract}
\usepackage{fancyhdr}

\title{Calibrating Teacher--Student Discrepancy for On-Policy Distillation}

\definecolor{ink}{HTML}{16243B}
\definecolor{accent}{HTML}{2D5D7B}
\definecolor{muted}{HTML}{66758A}
\selectfont
\setlist{topsep=0.3em,itemsep=0.2em,parsep=0pt}
\titleformat{\section}{\sffamily\Large\bfseries\color{ink}}{\color{accent}\thesection}{0.75em}{}
\titleformat{\subsection}{\sffamily\large\bfseries\color{ink}}{\color{accent}\thesubsection}{0.7em}{}
\titleformat{\subsubsection}{\sffamily\normalsize\bfseries\color{ink}}{\color{accent}\thesubsubsection}{0.65em}{}
\titlespacing*{\section}{0pt}{1.9ex plus 0.45ex minus 0.2ex}{0.7ex}
\titlespacing*{\subsection}{0pt}{1.45ex plus 0.35ex minus 0.15ex}{0.5ex}
\titlespacing*{\subsubsection}{0pt}{1.1ex plus 0.25ex minus 0.1ex}{0.35ex}

\pretitle{\begin{center}\sffamily\bfseries\color{ink}\fontsize{23}{27}\selectfont}
\posttitle{\par\end{center}\vspace{0.35em}\begin{center}\color{accent}\rule{0.84\textwidth}{0.75pt}\end{center}\vspace{0.85em}}
\preauthor{\begin{center}\sffamily\large\color{ink}}
\postauthor{\par\end{center}\vspace{0.65em}}
\predate{}\postdate{}
\renewcommand{\headrulewidth}{0.35pt}
\renewcommand{\headrule}{\hbox to\headwidth{\color{accent}\leaders\hrule height \headrulewidth\hfill}}
\fancypagestyle{plain}{\fancyhf{}\fancyfoot[C]{\footnotesize\sffamily\color{muted}\thepage}\renewcommand{\headrulewidth}{0pt}}

\author{%
Qiangqiang He\textsuperscript{1}\quad
Jin Li\textsuperscript{2}\quad
MingCai Chen\textsuperscript{3}\\[0.65em]
\small
\begin{tabular}{c}
\textsuperscript{1}State Key Laboratory for Novel Software Technology, Nanjing University, Nanjing, China\\
\textsuperscript{2}College of Software Engineering, Southeast University, Nanjing, China\\
\textsuperscript{3}Nanjing University of Posts and Telecommunications, Nanjing, China\\[0.2em]
\texttt{qqh@smail.nju.edu.cn}\quad
\texttt{jin\_li@seu.edu.cn}\quad
\texttt{chenmc@njupt.edu.cn}
\end{tabular}}
\date{}
\hypersetup{
  pdftitle={Calibrating Teacher--Student Discrepancy for On-Policy Distillation},
  pdfauthor={Qiangqiang He, Jin Li, MingCai Chen},
  colorlinks=true,
  linkcolor=accent,
  citecolor=accent,
  urlcolor=accent
}
\begin{document}

\maketitle

\begin{abstract}
On-policy distillation (OPD) improves reasoning models by learning the token-level discrepancy between a stronger teacher and an on-policy student. However, this discrepancy does not purely reflect the capability gap between the teacher and the student: it also contains deviations arising from the teacher itself, which are consequently mixed into the observed teacher--student discrepancy and indiscriminately learned by standard OPD during training. This issue is further exacerbated by privileged OPD, where privileged information induces larger teacher-side likelihood shifts, thereby encouraging the student to learn more of the teacher's own deviation. We introduce \textbf{Calibrated On-Policy Distillation (Cal-OPD)}, which estimates the teacher's self-deviation region through positive and negative privileged interventions and calibrates the original teacher--student discrepancy by retaining only the component that lies beyond this region. Experiments on mathematical reasoning benchmarks show that, while retaining only about 52--65\% of the original teacher--student discrepancy as the optimization signal, Cal-OPD consistently outperforms standard OPD and its variants across model scales.
\end{abstract}

\section{Introduction}
\label{sec:introduction}

Knowledge distillation (KD) transfers capabilities from a stronger teacher to a weaker student\citep{hinton2015distilling}, but conventional off-policy distillation suffers from distribution mismatch between fixed distillation data and the student's evolving policy distribution\citep{agarwal2024policy, gu2024minillm}. On-policy distillation (OPD) mitigates this mismatch by training directly on student-generated trajectories and using teacher--student token-level likelihood discrepancies as dense supervision\citep{agarwal2024policy, yang2025qwen3, jin2026entropy}. Compared with reinforcement learning with verifiable rewards (RLVR), which relies on sparse outcome-level rewards\citep{wen2026reinforcement, guo2025deepseek, yu2026dapo}, OPD provides token-level guidance throughout the trajectory, enabling more direct and efficient reasoning post-training\citep{yang2025qwen3, jin2026entropy}.

Standard OPD implicitly treats the teacher likelihood assigned to each student token as an equally reliable reference. Yet its stability varies substantially across tokens. Recent studies on reasoning models suggest that reasoning tokens tend to remain relatively stable, whereas stylistic or surface-form tokens, such as discourse markers and formatting choices, are substantially more sensitive to contextual interventions even when both the question and student rollout are held fixed\citep{pan2026rlcsd, he2026not}. Related work further shows that the reliability of teacher supervision can vary across tokens and reasoning positions\citep{liu2026teacher}. We refer to this token-specific variability in teacher likelihood as \textbf{Teacher Self-Deviation (TSD)}. Consequently, the observed teacher--student discrepancy reflects not only the underlying capability gap but also deviations arising from the teacher itself, which standard OPD indiscriminately incorporates into the learning signal.

\begin{figure}[t]
    \centering
    \includegraphics[width=0.95\linewidth]{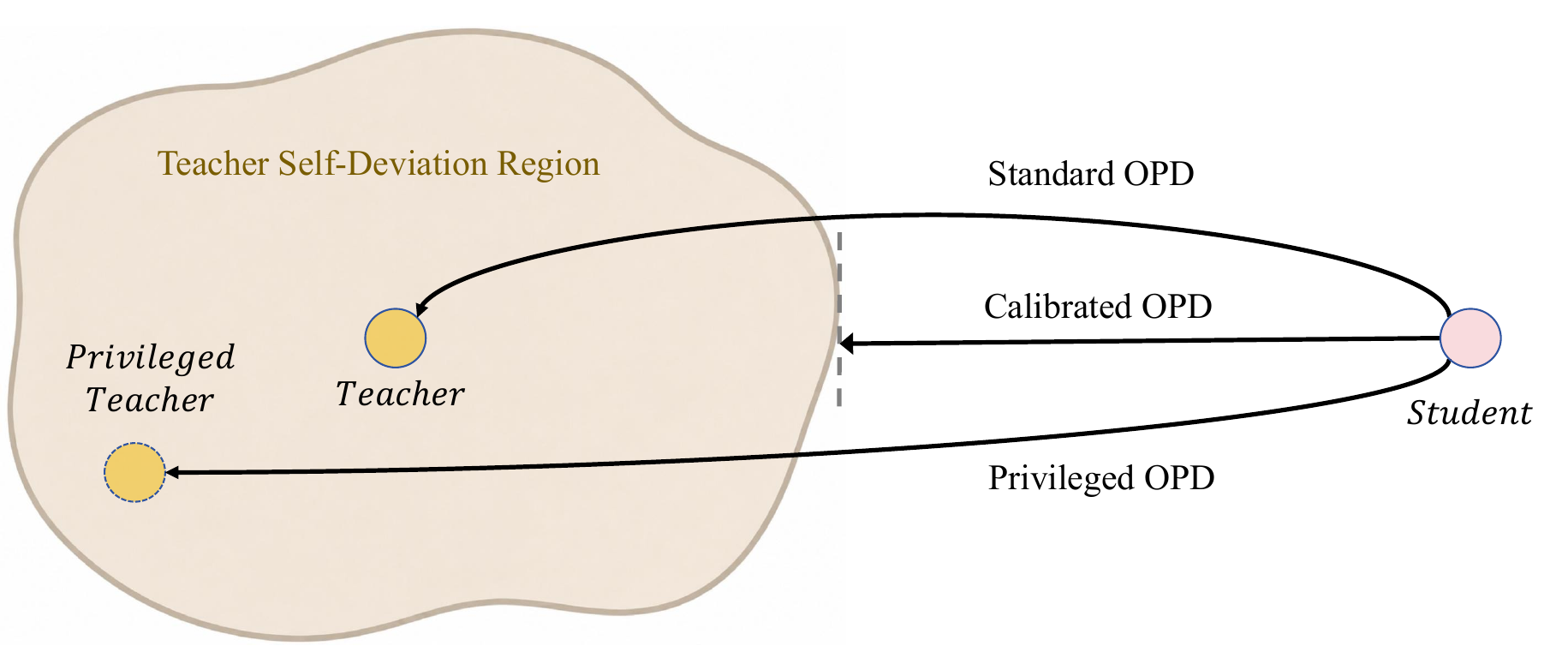}
    \caption{Conceptual comparison of standard OPD, privileged OPD, and Cal-OPD. Cal-OPD removes the TSD-explained discrepancy and retains only the residual beyond the TSD region.}
    \label{fig:cal_opd_intro}
\end{figure}

Privileged OPD extends standard OPD by conditioning a stronger teacher on additional training-time information, such as reference solutions, final answers, or hints\citep{ye2026policy, yu2026dopd, kaur2026rethinking}. While such privileged context is intended to improve teacher supervision, it also induces further shifts in the teacher distribution. Existing studies show that these shifts can introduce information-asymmetry effects\citep{yu2026dopd}, shortcut behavior\citep{tian2026vicur}, and even degrade performance in thinking models\citep{kaur2026rethinking}. From the perspective of teacher self-deviation, privileged OPD therefore amplifies the teacher-side deviations already embedded in the teacher--student discrepancy, causing the student to learn more context-induced teacher variation, as illustrated in Figure~\ref{fig:cal_opd_intro}.

To address this issue, we propose \textbf{Calibrated On-Policy Distillation (Cal-OPD)}, which first estimates the teacher's self-deviation region and then learns only the teacher--student discrepancy that lies beyond it. Specifically, Cal-OPD applies positive and negative privileged interventions, whose contrasting semantics provide complementary probes of the teacher's contextual variability and enable an approximation of its token-level self-deviation region. Unlike privileged OPD, these interventions are not directly distilled into the student, but are instead used to measure and calibrate the teacher reference. Cal-OPD then removes the portion of the original discrepancy covered by the estimated self-deviation region, retaining only the residual as the learning signal and setting it to zero when the student likelihood falls within this estimated region.

Our contributions are summarized as follows:
\begin{itemize}
    \item We identify and empirically characterize \textbf{Teacher Self-Deviation (TSD)}, showing that teacher likelihood is not an equally stable reference across tokens and contexts, and that learning these teacher-side deviations, which are further amplified by privileged context, can substantially degrade OPD for long-CoT reasoning.

    \item We introduce \textbf{Calibrated On-Policy Distillation (Cal-OPD)}, which uses positive and negative privileged interventions to estimate the teacher's self-deviation region and retains only the teacher--student discrepancy beyond it. Privileged information is thus used to calibrate the teacher reference rather than directly supervise the student.

    \item Extensive experiments on mathematical reasoning benchmarks show that Cal-OPD retains only about 52--65\% of the original teacher--student discrepancy for optimization, yet consistently outperforms standard OPD and its variants across model scales, while alleviating the response-length expansion observed in standard and privileged OPD.
\end{itemize}

\section{Understanding Teacher Self-Deviation}

\subsection{Preliminaries}
\label{sec:preliminaries}

\paragraph{On-Policy Distillation.}
Let $\pi_S=\pi_\theta$ denote the student policy parameterized by $\theta$, and let $\pi_T$ denote a fixed teacher policy. Given a problem $x\sim\mathcal{D}$, the student generates an on-policy rollout
\begin{equation}
    y=(y_1,\ldots,y_T)\sim\pi_S(\cdot\mid x),
\end{equation}
where $y_{<t}=(y_1,\ldots,y_{t-1})$ denotes the prefix of the student response preceding token $y_t$.
For each student-generated token, we evaluate the teacher and student under the same problem $x$ and student response prefix $y_{<t}$, and define their token-level log-likelihoods and discrepancy as
\begin{equation}
    \ell_t^T=\log\pi_T(y_t\mid x,y_{<t}),\qquad
    \ell_t^S=\log\pi_S(y_t\mid x,y_{<t}),\qquad
    A_t^{\mathrm{OPD}}=\ell_t^T-\ell_t^S.
\end{equation}
OPD uses $A_t^{\mathrm{OPD}}$ as the token-level advantage for optimizing the student:
\begin{equation}
    \mathcal{L}_{\mathrm{OPD}}(\theta)
    =
    -
    \mathbb{E}_{x\sim\mathcal{D},\,y\sim\pi_S(\cdot\mid x)}
    \left[
        \frac{1}{T}
        \sum_{t=1}^{T}
        \operatorname{sg}\!\left(A_t^{\mathrm{OPD}}\right)
        \log\pi_S(y_t\mid x,y_{<t})
    \right],
    \label{eq:opd_objective}
\end{equation}
where $\operatorname{sg}(\cdot)$ denotes the stop-gradient operator. Accordingly, $A_t^{\mathrm{OPD}}>0$ increases the student likelihood of $y_t$, whereas $A_t^{\mathrm{OPD}}<0$ decreases it. This formulation treats the teacher likelihood $\ell_t^T$ as the token-level reference against which the student is optimized.

\paragraph{Teacher Self-Deviation.}
For a fixed problem $x$, student rollout $y$, and token position $t$, we introduce a teacher-side contextual intervention $c$ while keeping the evaluated student trajectory unchanged. The intervention is provided only to the teacher, together with the problem $x$, and precedes the student rollout prefix $y_{<t}$ in the teacher context. We use $c_0=\varnothing$ to denote the original setting without additional context. The teacher likelihoods under intervention $c$ and the original setting $c_0$, together with the induced likelihood variation, are defined as
\begin{equation}
\begin{gathered}
    \ell_t^T(c)=\log\pi_T(y_t\mid x,c,y_{<t}),\qquad
    \ell_t^T(c_0)=\log\pi_T(y_t\mid x,y_{<t})=\ell_t^T,\\
    \Delta_t^T(c)=\ell_t^T(c)-\ell_t^T(c_0).
\end{gathered}
\end{equation}
Since $x$, $y_{<t}$, and $y_t$ remain fixed throughout the comparison, $\Delta_t^T(c)$ captures the change in teacher likelihood induced by the additional context $c$, rather than any change in the evaluated trajectory. We refer to this context-induced variation in teacher likelihood as \textbf{teacher self-deviation (TSD)}.

Given an intervention set $\mathcal{C}$, we use the induced likelihood variations to estimate the downward and upward magnitudes of TSD:
\begin{equation}
    \hat{d}_t^{\downarrow}
    =
    \max\left(
        0,\,
        -\min_{c\in\mathcal{C}}\Delta_t^T(c)
    \right),
    \qquad
    \hat{d}_t^{\uparrow}
    =
    \max\left(
        0,\,
        \max_{c\in\mathcal{C}}\Delta_t^T(c)
    \right).
\end{equation}
Both $\hat{d}_t^{\downarrow}$ and $\hat{d}_t^{\uparrow}$ are non-negative and measure the maximum downward and upward deviations in teacher log-likelihood, respectively. These quantities yield an empirical estimate of the \textbf{TSD region}:
\begin{equation}
    \hat{\mathcal{R}}_t^T(\mathcal{C})
    =
    \left[
        \ell_t^T-\hat{d}_t^{\downarrow},
        \;
        \ell_t^T+\hat{d}_t^{\uparrow}
    \right].
\end{equation}
Accordingly, $\hat{\mathcal{R}}_t^T(\mathcal{C})$ is a finite-intervention approximation to the underlying teacher self-deviation region $\mathcal{R}_t^T$, rather than an exhaustive characterization of all possible contextual variation. TSD characterizes variability in the teacher likelihood, while student-generated rollouts provide the trajectories on which this variability is measured.

\subsection{Teacher Self-Deviation Is Not Reliable Knowledge}

TSD captures contextual changes in the teacher likelihood assigned to a fixed token. If TSD \textbf{faithfully reflected task-relevant knowledge}, its variation should be systematically tied to the information introduced by the context. In particular, TSD should \textbf{depend on the presence of task-specific information} and \textbf{respond consistently to the correctness of that information}. Otherwise, the observed likelihood shift cannot be reliably interpreted as a knowledge signal. Table~\ref{tab:contextual_interventions} summarizes the contextual interventions used in our analysis, spanning task-agnostic instructions, evaluative feedback, and answer- and solution-level privileged information. We conduct this analysis with Qwen3-1.7B as the student and Qwen3-8B\citep{yang2025qwen3} as the teacher, both in thinking mode, over 6,528 questions sampled from DAPO-17k\citep{yu2026dapo}, with one response per question and approximately 60 million response tokens evaluated across all intervention conditions. Details of data collection and preparation for TSD analysis are provided in Appendix~\ref{app:tsd_data}.

\begin{table}[t]
\caption{
Contextual interventions used to probe TSD.
The shared problem context $x$ is omitted for brevity.
For answer- and solution-level interventions, the positive and negative variants share the same template and differ only in the inserted privileged content.
}
\label{tab:contextual_interventions}

\centering
\footnotesize
\setlength{\tabcolsep}{2pt}
\renewcommand{\arraystretch}{1.08}

\begin{tabularx}{\linewidth}{
    @{}
    >{\centering\arraybackslash}p{0.12\linewidth}
    >{\raggedright\arraybackslash}X
    @{}
}
\toprule
\textbf{Symbol}
&
\multicolumn{1}{c}{\textbf{Contextual Intervention}}
\\
\midrule

\addlinespace[1pt]
\multicolumn{2}{@{}c@{}}{\textbf{Task-Agnostic Instructions}}
\\[-1pt]

$c_{\mathrm{inst}}^{\mathrm{pos}}$
&
Please reason through the problem \textbf{carefully and thoroughly}. \textbf{Verify intermediate steps} and provide a complete, rigorous solution.
\\[1.5pt]

$c_{\mathrm{inst}}^{\mathrm{neg}}$
&
Please solve the problem \textbf{quickly and directly}. \textbf{Avoid unnecessary elaboration or extensive verification} and reach the final answer efficiently.
\\

\addlinespace[3pt]
\multicolumn{2}{@{}c@{}}{\textbf{Evaluative Feedback}}
\\[-1pt]

$c_{\mathrm{eval}}^{\mathrm{pos}}$
&
A gold-standard verifier has judged that the following solution \textbf{reaches the correct final answer}. The reasoning is \textbf{rigorous, coherent, and mathematically sound}.
\\[1.5pt]

$c_{\mathrm{eval}}^{\mathrm{neg}}$
&
A gold-standard verifier has judged that the following solution \textbf{does not reach the correct final answer}. The reasoning is \textbf{flawed, incoherent, and mathematically unreliable}.
\\

\addlinespace[3pt]
\multicolumn{2}{@{}c@{}}{\textbf{Answer-Level Privilege}}
\\[-1pt]

$c_{\mathrm{ans}}^{\mathrm{pos/neg}}$
&
A verified ground-truth answer is provided as a reliable reference:
\textbf{\emph{[correct / incorrect answer]}}. Use it to guide your reasoning while providing a complete and logically coherent solution.
\\

\addlinespace[3pt]
\multicolumn{2}{@{}c@{}}{\textbf{Solution-Level Privilege}}
\\[-1pt]

$c_{\mathrm{sol}}^{\mathrm{pos/neg}}$
&
A reference solution is provided as additional guidance:
\textbf{\emph{[correct / unrelated solution]}}. Use it while independently providing a complete and logically coherent solution.
\\

\bottomrule
\end{tabularx}
\end{table}

\paragraph{TSD Emerges Without Task Knowledge.}
We first find that \textbf{substantial TSD emerges even in the absence of external task-specific knowledge}, and that the affected token positions are largely preserved when richer privileged information is subsequently introduced. This indicates that part of the teacher-side variability amplified by privileged context is already present before answer- or solution-level knowledge is provided.

\begin{figure}[t]
    \centering

    \begin{tabular}{@{}c@{\hspace{3mm}}c@{}}
        \includegraphics[
            height=5.2cm
        ]{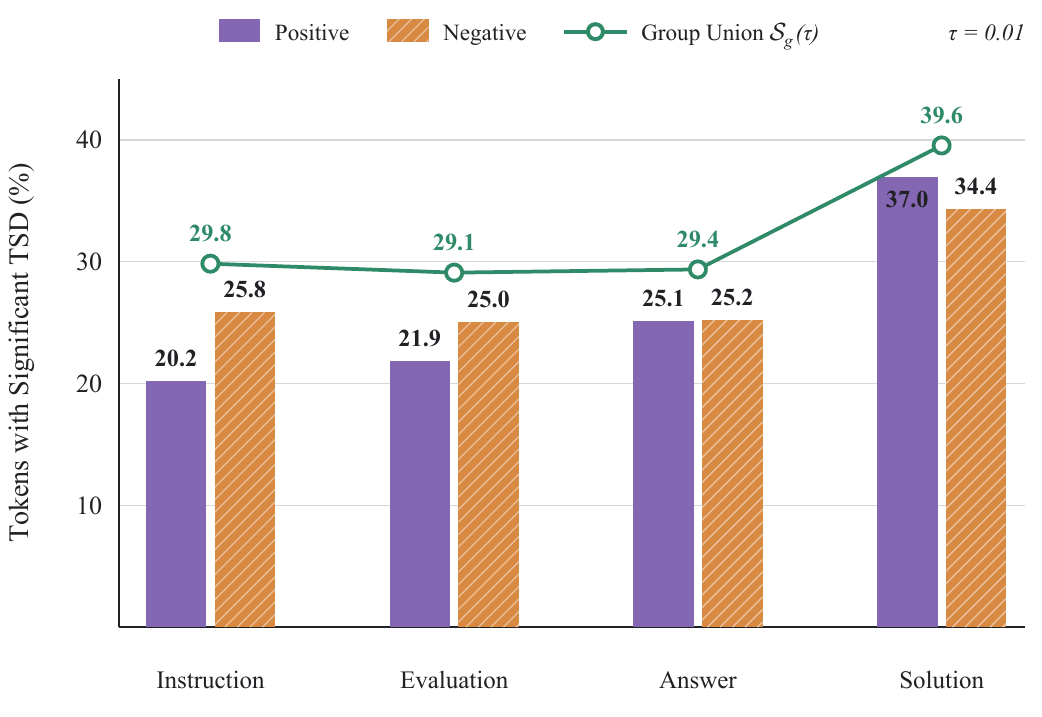}
        &
        \includegraphics[
            height=5.2cm
        ]{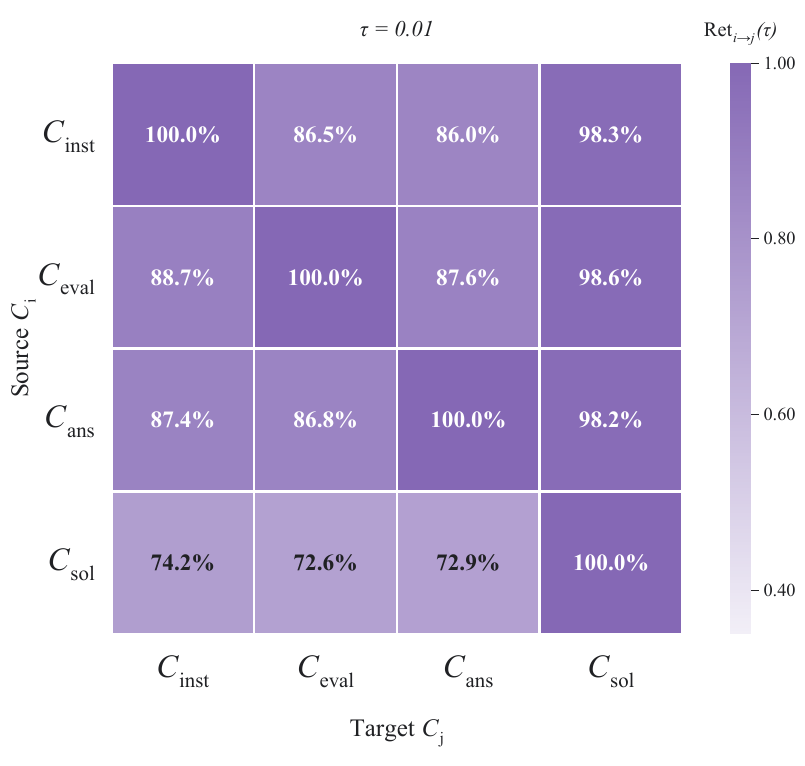}
        \\[-1mm]
        \small\textbf{(a) Prevalence of Significant TSD}
        &
        \small\textbf{(b) Retention of Significant TSD}
    \end{tabular}

    \vspace{-1mm}
    \caption{
        TSD across contextual interventions. Significant TSD emerges even under task-agnostic interventions, and the affected token positions are largely retained under richer privileged contexts.
    }
    \label{fig:tsd_significance_retention}
\end{figure}

To quantify this effect, we group the positive and negative variants of each intervention group as
\begin{equation}
    \mathcal{C}_{g}
    =
    \left\{
        c_{g}^{\mathrm{pos}},
        c_{g}^{\mathrm{neg}}
    \right\},
    \qquad
    g\in
    \{
        \mathrm{inst},
        \mathrm{eval},
        \mathrm{ans},
        \mathrm{sol}
    \}.
\end{equation}
Given a threshold $\tau$, we define the set of tokens exhibiting \textbf{significant TSD} under group $\mathcal{C}_{g}$ as
\begin{equation}
    \mathcal{S}_{g}(\tau)
    =
    \left\{
        t:
        \max_{c\in\mathcal{C}_{g}}
        \left|
            \Delta_t^T(c)
        \right|
        >
        \tau
    \right\}.
\end{equation}
Figure~\ref{fig:tsd_significance_retention}(a) reports the prevalence of significant TSD under each intervention, together with the union prevalence defined by $\mathcal{S}_{g}(\tau)$. At $\tau=0.01$, task-agnostic instructions induce significant TSD on $20.2\%$ and $25.8\%$ of tokens under the positive and negative variants, respectively, with their union covering $29.8\%$ of all tokens. This prevalence is comparable to the $29.1\%$ observed under evaluative feedback and the $29.4\%$ under answer-level privilege, despite task-agnostic interventions providing neither an answer nor a solution. Solution-level privilege further expands the affected set to $39.6\%$, showing that richer privileged context broadens TSD rather than creating it from scratch.

We measure whether TSD under less informative interventions persists under richer privileged contexts. For groups $\mathcal{C}_{i}$ and $\mathcal{C}_{j}$, we define the retention of significant TSD as
\begin{equation}
    \operatorname{Ret}_{i\rightarrow j}(\tau)
    =
    \frac{
        \left|
            \mathcal{S}_{i}(\tau)
            \cap
            \mathcal{S}_{j}(\tau)
        \right|
    }{
        \left|
            \mathcal{S}_{i}(\tau)
        \right|
    }.
\end{equation}
Figure~\ref{fig:tsd_significance_retention}(b) reveals a clear asymmetry. Under solution-level privilege, $98.3\%$, $98.6\%$, and $98.2\%$ of tokens exhibiting significant TSD under task-agnostic instructions, evaluative feedback, and answer-level privilege remain significant, respectively, whereas only $74.2\%$, $72.6\%$, and $72.9\%$ of solution-level significant-TSD tokens remain significant under these less informative interventions. Thus, richer privileged context largely preserves previously affected positions while extending TSD to additional positions. Together, these results show that \textbf{task knowledge is not necessary for TSD to emerge}, while richer privileged contexts mainly broaden the affected set rather than introducing an entirely new deviation pattern.

\begin{figure}[t]
    \centering
    \includegraphics[width=0.85\linewidth]{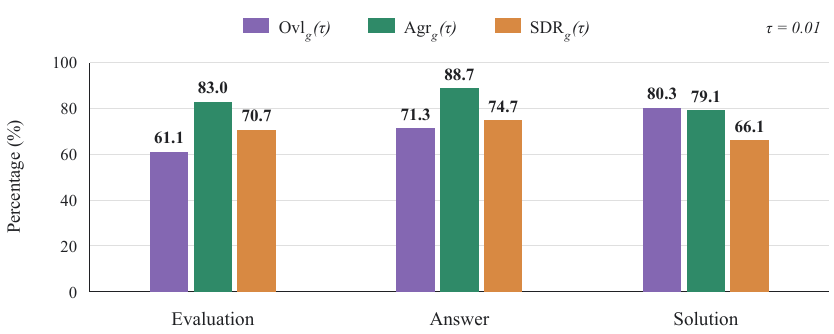}
    \caption{
        Consistency of TSD under contrasting interventions. TSD occurs at overlapping token positions, remains directionally aligned, and is dominated by a shared common-mode component.
    }
    \label{fig:tsd_semantic_consistency}
\end{figure}

\paragraph{TSD Is Largely Insensitive to Intervention Semantics.}
We find that TSD is largely \textbf{insensitive to both semantic polarity and correctness}. To quantify this consistency, we compare the positive and negative variants of evaluative, answer-level, and solution-level interventions using three metrics. For intervention group $g$, let $\Delta_t^{+}=\Delta_t^T(c_g^{\mathrm{pos}})$ and $\Delta_t^{-}=\Delta_t^T(c_g^{\mathrm{neg}})$, with corresponding significant-TSD sets $S_g^{+}(\tau)=\{t:|\Delta_t^{+}|>\tau\}$ and $S_g^{-}(\tau)=\{t:|\Delta_t^{-}|>\tau\}$. \textbf{Overlap} measures whether significant TSD occurs at the same token positions, while \textbf{Agreement} measures whether the two interventions shift teacher likelihood in the same direction at jointly significant positions:
\begin{equation}
    \operatorname{Ovl}_g(\tau)
    =
    \frac{|S_g^{+}(\tau)\cap S_g^{-}(\tau)|}
         {|S_g^{+}(\tau)\cup S_g^{-}(\tau)|},
    \qquad
    \operatorname{Agr}_g(\tau)
    =
    \frac{
        \sum_{t\in S_g^{+}(\tau)\cap S_g^{-}(\tau)}
        \mathbf{1}\!\left[\Delta_t^{+}\Delta_t^{-}>0\right]
    }{
        |S_g^{+}(\tau)\cap S_g^{-}(\tau)|
    }.
\end{equation}
We further measure how much paired variation is shared between the two interventions. Let $I_g(\tau)=S_g^{+}(\tau)\cap S_g^{-}(\tau)$ denote the set of jointly significant tokens. We decompose the paired deviations into shared and contrastive components and define the \textbf{Shared Deviation Ratio (SDR)} as
\begin{equation}
    M_t=\frac{\Delta_t^{+}+\Delta_t^{-}}{2},\qquad
    D_t=\frac{\Delta_t^{+}-\Delta_t^{-}}{2},\qquad
    \operatorname{SDR}_g(\tau)
    =
    \frac{\sum_{t\in I_g(\tau)}|M_t|}
         {\sum_{t\in I_g(\tau)}\left(|M_t|+|D_t|\right)}.
\end{equation}
A higher shared deviation ratio indicates that, among jointly significant tokens, paired TSD is dominated by variation shared across interventions, with less attributable to their semantic difference.

Figure~\ref{fig:tsd_semantic_consistency} reveals that TSD remains highly structured under contrasting intervention semantics. Reversing answer correctness still yields $88.7\%$ directional agreement and a $74.7\%$ shared deviation ratio, showing that paired likelihood shifts are dominated by a shared response rather than the correctness contrast itself. Solution-level interventions exhibit the highest positional overlap at $80.3\%$, yet the lowest shared deviation ratio at $66.1\%$, suggesting that richer context alters how TSD varies more than where it emerges. Across all groups, substantial overlap, high directional agreement, and predominantly shared variation persist. These results show that \textbf{TSD is largely insensitive to intervention semantics and correctness}, further indicating that such teacher likelihood shifts cannot be reliably interpreted as task knowledge. Additional analyses across multiple thresholds and model scales in Appendix~\ref{app:tsd_semantic_consistency} reproduce these findings.

\begin{table}[t]
\caption{
Token forms with the highest and lowest significant-TSD rates
$\rho_{\tau}(v)$ under $c_{\mathrm{sol}}^{\mathrm{pos}}$ at $\tau=0.01$,
restricted to forms occurring more than 20,000 times.
$\overline{|\Delta|}$ denotes the mean TSD magnitude over significant occurrences.
Values are rounded for display; rankings are based on the unrounded
$\rho_{\tau}(v)$.
For compactness, \texttt{altern.} abbreviates \texttt{alternatively}.
}
\label{tab:top_bottom_tsd_tokens}

\centering
\footnotesize
\setlength{\tabcolsep}{0pt}
\renewcommand{\arraystretch}{1.06}

\begin{tabularx}{\linewidth}{
    @{}
    r
    @{\hspace{4pt}}
    >{\raggedright\arraybackslash}p{0.105\linewidth}
    @{\hspace{2.5pt}}
    r
    @{\hspace{9pt}}
    r
    X
    r
    @{\hspace{4pt}}
    >{\raggedright\arraybackslash}p{0.105\linewidth}
    @{\hspace{2.5pt}}
    r
    @{\hspace{9pt}}
    r
    X
    r
    @{\hspace{4pt}}
    >{\raggedright\arraybackslash}p{0.105\linewidth}
    @{\hspace{2.5pt}}
    r
    @{\hspace{9pt}}
    r
    @{}
}
\toprule

\multicolumn{1}{c@{\hspace{4pt}}}{\#}
& \textbf{Token}
& \multicolumn{1}{c@{\hspace{9pt}}}{$\boldsymbol{\rho_{\tau}}$ (\%)}
& \multicolumn{1}{c}{$\mathbf{\overline{|\Delta|}}$}
&
&
\multicolumn{1}{c@{\hspace{4pt}}}{\#}
& \textbf{Token}
& \multicolumn{1}{c@{\hspace{9pt}}}{$\boldsymbol{\rho_{\tau}}$ (\%)}
& \multicolumn{1}{c}{$\mathbf{\overline{|\Delta|}}$}
&
&
\multicolumn{1}{c@{\hspace{4pt}}}{\#}
& \textbf{Token}
& \multicolumn{1}{c@{\hspace{9pt}}}{$\boldsymbol{\rho_{\tau}}$ (\%)}
& \multicolumn{1}{c}{$\mathbf{\overline{|\Delta|}}$}
\\
\midrule

\multicolumn{14}{@{}c@{}}{
    \makebox[\linewidth][c]{\textbf{Highest $\rho_{\tau}$}}
}
\\
\addlinespace[1pt]

1 & \texttt{maybe}     & 97.4 & 0.59
& & 7  & \texttt{seems}   & 93.6 & 0.41
& & 13 & \texttt{wait}    & 90.0 & 0.42 \\

2 & \texttt{however}   & 97.4 & 0.63
& & 8  & \texttt{another} & 92.9 & 0.50
& & 14 & \texttt{check}   & 89.6 & 0.28 \\

3 & \texttt{therefore} & 95.9 & 0.46
& & 9  & \texttt{since}   & 91.8 & 0.44
& & 15 & \texttt{let}     & 89.5 & 0.50 \\

4 & \texttt{consider}  & 95.4 & 0.38
& & 10 & \texttt{try}     & 91.8 & 0.40
& & 16 & \texttt{here}    & 89.4 & 0.36 \\

5 & \texttt{altern.}   & 94.3 & 0.58
& & 11 & \texttt{think}   & 91.6 & 0.40
& & 17 & \texttt{because} & 89.4 & 0.40 \\

6 & \texttt{earlier}   & 94.2 & 0.41
& & 12 & \texttt{says}    & 90.4 & 0.25
& & 18 & \texttt{now}     & 89.2 & 0.49 \\

\midrule

\multicolumn{14}{@{}c@{}}{
    \makebox[\linewidth][c]{\textbf{Lowest $\rho_{\tau}$}}
}
\\
\addlinespace[1pt]

1 & \texttt{\}\{}      & 2.8 & 0.25
& & 7  & $^\circ$      & 7.3 & 0.17
& & 13 & \texttt{frac} & 8.0 & 0.22 \\

2 & \texttt{\#\#\#}    & 3.2 & 0.13
& & 8  & \texttt{\_}   & 7.3 & 0.29
& & 14 & \texttt{\{}   & 8.0 & 0.23 \\

3 & \texttt{0}         & 5.0 & 0.36
& & 9  & \texttt{6}    & 7.4 & 0.40
& & 15 & \texttt{4}    & 8.8 & 0.37 \\

4 & \texttt{9}         & 6.8 & 0.38
& & 10 & \texttt{7}    & 7.7 & 0.39
& & 16 & \texttt{2}    & 8.8 & 0.32 \\

5 & \texttt{8}         & 7.2 & 0.39
& & 11 & \texttt{5}    & 7.8 & 0.38
& & 17 & \texttt{3}    & 9.2 & 0.36 \\

6 & $\surd$            & 7.2 & 0.28
& & 12 & $\theta$      & 7.9 & 0.28
& & 18 & \texttt{\_i}  & 9.2 & 0.28 \\

\bottomrule
\end{tabularx}
\end{table}

\subsection{TSD Concentrates on Surface-Form Tokens}

TSD is \textbf{strongly concentrated on surface-form tokens rather than tokens carrying mathematical content}. High-TSD forms are natural-language markers that organize, qualify, or redirect the reasoning text without directly encoding problem-specific mathematical information, whereas low-TSD forms are dominated by numbers, mathematical symbols, and notation. Under $c_{\mathrm{sol}}^{\mathrm{pos}}$, we merge tokenizer variants corresponding to the same surface form and restrict the analysis to forms occurring more than 20,000 times. For each token form $v$, we define the significant-TSD rate as $\rho_{\tau}(v)=\Pr\!\left(|\Delta_t^T(c_{\mathrm{sol}}^{\mathrm{pos}})|>\tau\mid y_t=v\right)$. Unlike occurrence counts, $\rho_{\tau}(v)$ measures how often a token exhibits significant TSD when it appears, thereby controlling for differences in token frequency.

Table~\ref{tab:top_bottom_tsd_tokens} reveals a clear separation between token forms with the highest and lowest significant-TSD rates at $\tau=0.01$. The 18 highest-ranked forms are dominated by natural-language surface expressions such as \texttt{maybe}, \texttt{however}, \texttt{therefore}, \texttt{consider}, and \texttt{alternatively}, with $\rho_{\tau}$ exceeding $89\%$ throughout, meaning significant TSD appears in about nine of ten occurrences. In contrast, the 18 lowest-ranked forms consist predominantly of digits, mathematical symbols, and notation such as \texttt{0}, $\surd$, $\theta$, and \texttt{frac}, all with $\rho_{\tau}$ below $9.3\%$. This nearly order-of-magnitude separation shows that significant TSD occurs far more frequently on surface-form tokens than on tokens directly expressing mathematical content. The large $\overline{|\Delta|}$ values in Table~\ref{tab:top_bottom_tsd_tokens} mainly reflect the stronger teacher-side shifts induced by solution-level privilege. Appendix~\ref{app:tsd_token_patterns} extends the analysis to the top-24 and bottom-24 token forms across different contextual interventions, showing that weaker interventions substantially reduce deviation magnitudes while preserving the same separation between surface-form tokens and mathematical or symbolic forms. Appendix~\ref{app:tsd_trajectory_case} provides a trajectory-level case study illustrating how TSD is distributed throughout a complete reasoning trace.

\section{Calibrated On-Policy Distillation}
\label{sec:cal_opd}

The preceding analysis shows that teacher likelihood is not an equally reliable pointwise reference: it exhibits substantial TSD that cannot be reliably interpreted as a knowledge signal. We therefore propose \textbf{Calibrated On-Policy Distillation (Cal-OPD)}, which decomposes the observed teacher--student discrepancy into a TSD-explained component and a calibrated residual, and distills only the discrepancy that remains beyond the teacher's estimated self-deviation region.

\begin{equation}
    \underbrace{A_t^{\mathrm{OPD}}}_{\text{Teacher--Student Discrepancy}}
    =
    \underbrace{A_t^{\mathrm{Cal}}}_{\text{Calibrated Teacher--Student Discrepancy}}
    +
    \underbrace{A_t^{\mathrm{TSD}}}_{\text{TSD-Explained Discrepancy}}.
\end{equation}

For each token $y_t$, Cal-OPD probes the teacher with two contrasting interventions $c^{\mathrm{pos}}$ and $c^{\mathrm{neg}}$, producing $\Delta_t^{\mathrm{pos}}=\Delta_t^T(c^{\mathrm{pos}})$ and $\Delta_t^{\mathrm{neg}}=\Delta_t^T(c^{\mathrm{neg}})$. We define their maximum downward and upward deviations as $\hat d_t^{\downarrow}=\max(0,-\Delta_t^{\mathrm{pos}},-\Delta_t^{\mathrm{neg}})$ and $\hat d_t^{\uparrow}=\max(0,\Delta_t^{\mathrm{pos}},\Delta_t^{\mathrm{neg}})$. Since two interventions provide only a finite probe of the underlying TSD, we introduce a relaxation factor $\lambda\geq1$ and estimate the TSD region as
\begin{equation}
    \hat{\mathcal{R}}_t^T
    =
    \left[
        \ell_t^T-\lambda\hat d_t^{\downarrow},
        \;
        \ell_t^T+\lambda\hat d_t^{\uparrow}
    \right]
    =
    [L_t^T,U_t^T].
\end{equation}

Cal-OPD then removes the portion of the teacher--student discrepancy covered by the estimated TSD region and retains only the residual beyond its boundary:
\begin{equation}
    A_t^{\mathrm{Cal}}
    =
    \left[L_t^T-\ell_t^S\right]_+
    -
    \left[\ell_t^S-U_t^T\right]_+,
    \qquad
    [z]_+=\max(z,0).
\end{equation}
When $\ell_t^S\in\hat{\mathcal{R}}_t^T$, the observed teacher--student discrepancy is fully covered by the estimated TSD region and $A_t^{\mathrm{Cal}}=0$; otherwise, $A_t^{\mathrm{Cal}}$ retains only the discrepancy beyond the nearest boundary. Accordingly, $A_t^{\mathrm{TSD}}=A_t^{\mathrm{OPD}}-A_t^{\mathrm{Cal}}$. Cal-OPD follows the standard OPD objective in Eq.~\ref{eq:opd_objective}, replacing $A_t^{\mathrm{OPD}}$ with $A_t^{\mathrm{Cal}}$ as the token-level advantage.

\begin{table}[t]
\caption{
Main results on mathematical reasoning benchmarks. The best result among distillation methods for each teacher--student configuration is shown in \textbf{bold}. Subscripts on Cal-OPD indicate gains over the corresponding student. For compactness, \textbf{4B-2507} and \textbf{30B-2507} denote Qwen3-4B-Thinking-2507 and Qwen3-30B-A3B-Thinking-2507, respectively.
}
\label{tab:main_results}

\centering
\small
\setlength{\tabcolsep}{2.2pt}
\renewcommand{\arraystretch}{1.08}

\begin{tabularx}{\linewidth}{
@{}
>{\raggedright\arraybackslash}p{0.23\linewidth}
>{\hsize=0.88\hsize\centering\arraybackslash}X
>{\hsize=0.88\hsize\centering\arraybackslash}X
>{\hsize=0.88\hsize\centering\arraybackslash}X
>{\hsize=0.88\hsize\centering\arraybackslash}X
>{\hsize=1.35\hsize\centering\arraybackslash}X
>{\hsize=1.15\hsize\centering\arraybackslash}X
>{\hsize=0.98\hsize\centering\arraybackslash}X
@{}
}
\toprule

\textbf{Model / Method}
& \textbf{AMC23}
& \textbf{AIME24}
& \textbf{AIME25}
& \textbf{AIME26}
& \textbf{HMMT26}
& \textbf{MATH500}
& \textbf{Avg.}
\\
\midrule

\multicolumn{8}{@{}c@{}}{
\makebox[\linewidth][c]{
\textbf{4B-2507 $\rightarrow$ Qwen3-1.7B}
}
}
\\
\addlinespace[1pt]

Student
& 80.8 & 40.6 & 32.5 & 29.2 & 21.8 & 90.2 & 49.2 \\

Teacher
& 94.8 & 56.5 & 53.1 & 54.2 & 25.2 & 95.4 & 63.2 \\

\addlinespace[1pt]

OPD
& 82.0 & 39.6 & 35.0 & 34.6 & 22.9 & 90.6 & 50.8 \\

ExOPD
& 81.7 & 43.1 & 35.4 & 34.2 & \textbf{25.6} & 90.6 & 51.8 \\

EOPD
& 81.6 & 44.0 & 36.7 & \textbf{38.8} & 19.7 & 90.6 & 51.9 \\

Uni-OPD
& 81.4 & 42.1 & 34.4 & 32.7 & 22.5 & 89.8 & 50.5 \\

Privileged-OPD
& 83.3 & 39.8 & 30.4 & 31.0 & 21.8 & 89.4 & 49.3 \\

\textbf{Cal-OPD}
& $\mathbf{84.5}_{+3.7}$
& $\mathbf{45.0}_{+4.4}$
& $\mathbf{37.1}_{+4.6}$
& $36.3_{+7.1}$
& $24.8_{+3.0}$
& $\mathbf{90.8}_{+0.6}$
& $\mathbf{53.1}_{+3.9}$ \\

\midrule

\multicolumn{8}{@{}c@{}}{
\makebox[\linewidth][c]{
\textbf{30B-2507 $\rightarrow$ Qwen3-4B}
}
}
\\
\addlinespace[1pt]

Student
& 95.2 & 64.8 & 56.3 & 57.9 & 30.1 & 95.4 & 66.6 \\

Teacher
& 97.8 & 74.8 & 65.4 & 66.0 & 33.9 & 96.9 & 72.5 \\

\addlinespace[1pt]

OPD
& 94.2 & 62.1 & 57.5 & 55.2 & 31.1 & 95.0 & 65.9 \\

ExOPD
& 92.0 & 66.9 & 59.4 & 56.3 & \textbf{33.7} & 95.6 & 67.3 \\

EOPD
& 95.3 & 62.7 & 54.6 & 60.0 & 33.5 & \textbf{96.7} & 67.1 \\

Uni-OPD
& 93.4 & \textbf{67.9} & 59.8 & 57.9 & 31.4 & 95.4 & 67.6 \\

Privileged-OPD
& 93.4 & 58.8 & 52.1 & 54.0 & 31.3 & 94.4 & 64.0 \\

\textbf{Cal-OPD}
& $\mathbf{96.3}_{+1.1}$
& $67.1_{+2.3}$
& $\mathbf{61.3}_{+5.0}$
& $\mathbf{60.2}_{+2.3}$
& $33.3_{+3.2}$
& $95.8_{+0.4}$
& $\mathbf{69.0}_{+2.4}$ \\

\bottomrule
\end{tabularx}
\end{table}

\section{Experiments}

\subsection{Experimental Setup}

\paragraph{Datasets.}
We use DAPO-17K~\citep{yu2026dapo} filtered by Qwen3-235B-A22B-Instruct-2507\citep{yang2025qwen3} as the training dataset. The filtering and solution-annotation procedure is described in Appendix~\ref{app:tsd_data}. For evaluation, we consider six reasoning benchmarks: AMC23, AIME24, AIME25, AIME26, HMMT26, and MATH500~\citep{lightman2024let}. These benchmarks span a range of difficulty, from standard problem solving to high-difficulty competition mathematics.

\paragraph{Models and Baselines.}
We evaluate Cal-OPD on two teacher--student configurations from the Qwen3 family~\citep{yang2025qwen3}: Qwen3-4B-Thinking-2507$\rightarrow$Qwen3-1.7B and Qwen3-30B-A3B-Thinking-2507$\rightarrow$Qwen3-4B, covering two distinct scale regimes. We compare against five representative OPD baselines: standard OPD~\citep{agarwal2024policy}, which directly optimizes teacher--student discrepancies; ExOPD~\citep{yang2026learning}, augmented with reward extrapolation; EOPD~\citep{jin2026entropy}, with entropy-aware forward-KL supervision; Uni-OPD~\citep{hou2026uni}, with exploration and outcome-guided calibration; and Privileged-OPD~\citep{ye2026policy,kaur2026rethinking}, where the teacher is additionally conditioned on the ground-truth reference solution. This evaluates if Cal-OPD remains effective across varying capacity gaps and diverse modifications of the standard OPD objective. We report Avg@16 accuracy, computed by sampling 16 independent responses per problem and averaging their binary correctness scores across the full evaluation set.

\paragraph{Implementation Details.}
We implement all methods with \texttt{verl}~\citep{sheng2025hybridflow} and train on 8 NVIDIA H20 GPUs, with 4 GPUs hosting the student and 4 hosting the teacher. All methods are trained for 100 steps, with 256 trajectories per step, one rollout per question unless otherwise specified, and a learning rate of $1\times10^{-6}$. During training, we set response length to 16,384, temperature to $1.0$, and top-$p$ to $1.0$; during evaluation, we use response length of 20,480, temperature $0.6$, and top-$p$ to $0.95$. For Cal-OPD, we use $c_{\mathrm{eval}}^{\mathrm{pos}}$ and $c_{\mathrm{eval}}^{\mathrm{neg}}$ as contrasting interventions and set the relaxation factor to $\lambda=5$. Method-specific hyperparameters are provided in Appendix~\ref{app:training_details}.

\begin{table}[t]
\caption{
Effect of contextual interventions used for TSD estimation on Cal-OPD performance for
Qwen3-4B-Thinking-2507$\rightarrow$Qwen3-1.7B.
}
\label{tab:intervention_ablation}

\centering
\small
\setlength{\tabcolsep}{2.6pt}
\renewcommand{\arraystretch}{1.08}

\begin{tabularx}{\linewidth}{
@{}
>{\raggedright\arraybackslash}p{0.235\linewidth}
>{\centering\arraybackslash}X
>{\centering\arraybackslash}X
>{\centering\arraybackslash}X
>{\centering\arraybackslash}X
>{\centering\arraybackslash}X
>{\centering\arraybackslash}X
>{\centering\arraybackslash}X
@{}
}
\toprule

\textbf{Intervention Set}
& \textbf{AMC23}
& \textbf{AIME24}
& \textbf{AIME25}
& \textbf{AIME26}
& \textbf{HMMT26}
& \textbf{MATH500}
& \textbf{Avg.}
\\

\midrule

$\mathcal{C}_{\mathrm{inst}}$
& 83.9 & 40.8 & 34.8 & 34.2 & \textbf{25.2} & 90.3 & 51.5 \\

$\mathcal{C}_{\mathrm{eval}}$
& \textbf{84.5} & \textbf{45.0} & \textbf{37.1} & 36.3 & 24.8 & 90.8 & \textbf{53.1} \\

$\mathcal{C}_{\mathrm{ans}}$
& 84.2 & 40.8 & 34.8 & \textbf{37.3} & 23.9 & \textbf{90.9} & 52.0 \\

$\mathcal{C}_{\mathrm{sol}}$
& 81.3 & 40.6 & 30.0 & 30.8 & 22.0 & 89.2 & 49.0 \\

\bottomrule
\end{tabularx}
\end{table}

\subsection{Main Results}

Table~\ref{tab:main_results} shows that Cal-OPD achieves the highest average performance in both teacher--student configurations: $53.1$ for Qwen3-4B-Thinking-2507$\rightarrow$Qwen3-1.7B and $69.0$ for Qwen3-30B-A3B-Thinking-2507$\rightarrow$Qwen3-4B. This corresponds to gains of $+3.9$ and $+2.4$ over the student baselines, and $+2.3$ and $+3.1$ over standard OPD. Cal-OPD also achieves the best result on 7 of 12 benchmark--configuration pairs, demonstrating consistent gains across model scales and reasoning benchmarks. In contrast, standard OPD improves the 4B$\rightarrow$1.7B student only modestly, from $49.2$ to $50.8$, and even degrades the 30B$\rightarrow$4B student from $66.6$ to $65.9$, despite both teachers being substantially stronger than their students. This shows that raw teacher--student discrepancy is \textbf{not uniformly beneficial supervision}: optimizing all discrepancies also learns teacher-side deviations, whereas Cal-OPD removes the TSD-explained component and retains a more effective signal.

Privileged-OPD exhibits the strongest degradation, obtaining the lowest average performance among distillation methods in both configurations, at $49.3$ and $64.0$, respectively. In the 4B$\rightarrow$1.7B setting, it nearly eliminates the gain from standard OPD, while in the 30B$\rightarrow$4B setting it falls $2.6$ points below the student. Together with our earlier finding that privileged context substantially amplifies TSD, these results suggest that directly distilling the privileged teacher can transfer teacher-side deviation alongside task-relevant information, offsetting the benefit of stronger supervision. Cal-OPD mitigates this effect by filtering out the TSD-explained component before distillation.

\subsection{Analysis and Ablations}

\begin{figure}[t]
    \centering
    \begin{tabular}{@{}c@{\hspace{2mm}}c@{\hspace{2mm}}c@{}}
        \includegraphics[width=0.315\linewidth]{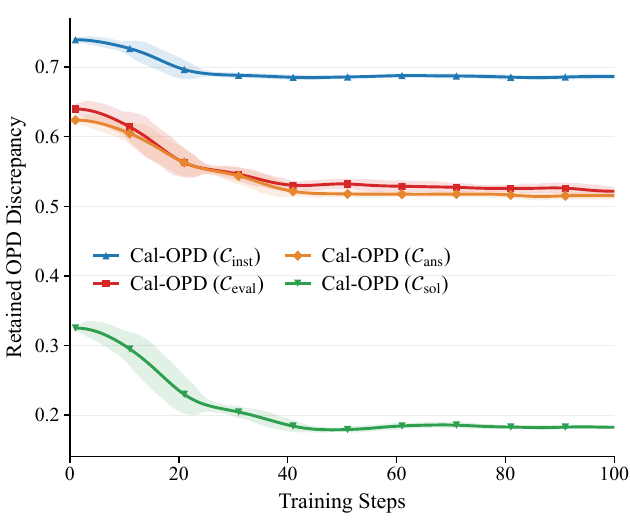}
        &
        \includegraphics[width=0.315\linewidth]{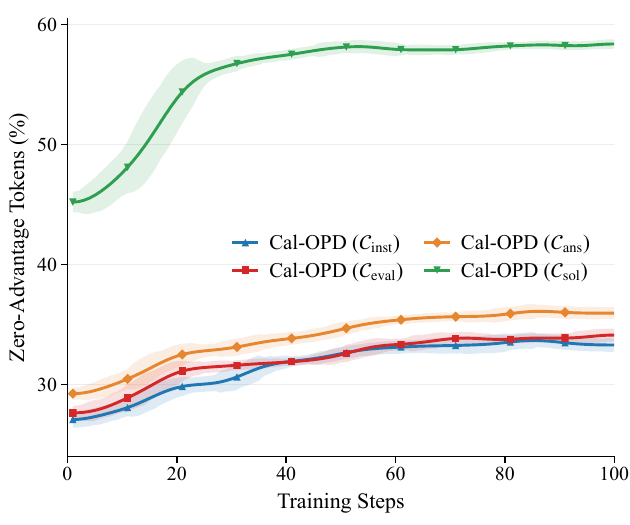}
        &
        \includegraphics[width=0.315\linewidth]{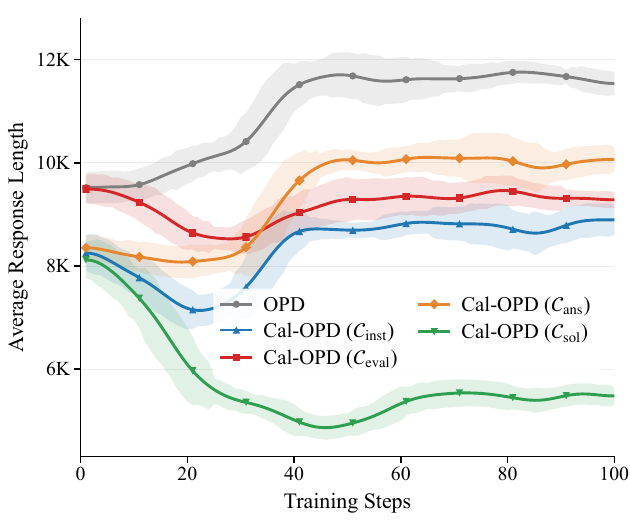}
        \\
        \small\textbf{(a) Retained OPD Discrepancy}
        &
        \small\textbf{(b) Zero-Advantage Token Ratio}
        &
        \small\textbf{(c) Average Response Length}
    \end{tabular}

    \vspace{-1mm}
\caption{
Training dynamics of Cal-OPD under different TSD-estimation interventions for
Qwen3-4B-Thinking-2507$\rightarrow$Qwen3-1.7B.
}
    \label{fig:cal_opd_signal_analysis}
\end{figure}

\paragraph{Effect of Contextual Interventions on Cal-OPD Performance.}
Table~\ref{tab:intervention_ablation} compares different intervention sets with $\lambda=5$ throughout, while Figure~\ref{fig:cal_opd_signal_analysis} shows their calibration dynamics. $\mathcal{C}_{\mathrm{eval}}$ achieves the best performance, reaching an average of $53.1$. Although evaluative feedback introduces external judgment, it provides no task-specific solution knowledge and yields stronger calibration than instruction interventions, which retain nearly $70\%$ of the teacher--student discrepancy. In contrast, $\mathcal{C}_{\mathrm{sol}}$ causes the largest degradation, with the average falling to $49.0$ while retaining only about $20\%$ of the discrepancy. This suggests that solution-induced TSD contains a larger task-relevant component, such that using it for calibration over-filters useful teacher supervision. Overall, $\mathcal{C}_{\mathrm{eval}}$ provides a stronger probe of TSD without the excessive filtering induced by solution-level privilege. Further training dynamics, including entropy and other measures, are provided in Appendix~\ref{app:training_dynamics}.

\paragraph{Effects of Calibration on OPD Training Dynamics.}
Figure~\ref{fig:cal_opd_signal_analysis} shows the training dynamics of the main $\mathcal{C}_{\mathrm{eval}}$ configuration. The retained-discrepancy ratio decreases from $65\%$ to $52\%$, while the zero-advantage token ratio increases from $27\%$ to $34\%$, indicating stronger filtering of the OPD signal. Calibration also alters response-length dynamics. Standard OPD expands the average response length from about $9.8$K to $11.8$K tokens, whereas Cal-OPD ends at only $9.3$K after an initial decrease to $8.5$K. Despite the additional teacher computation for estimating TSD, the shorter trajectories make Cal-OPD approximately $1.26\times$ faster to train. Together with the main results, these dynamics show that filtering the TSD-explained component yields a more effective and efficient supervision signal. A complete efficiency analysis is provided in Appendix~\ref{app:efficiency}.

\begin{figure}[t]
    \centering
    \begin{tabular}{@{}c@{\hspace{2mm}}c@{}}
        \includegraphics[width=0.46\linewidth]{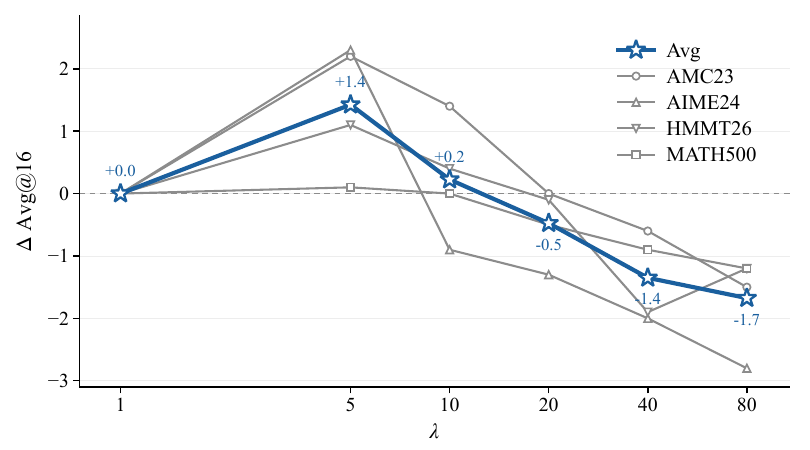}
        &
        \includegraphics[width=0.46\linewidth]{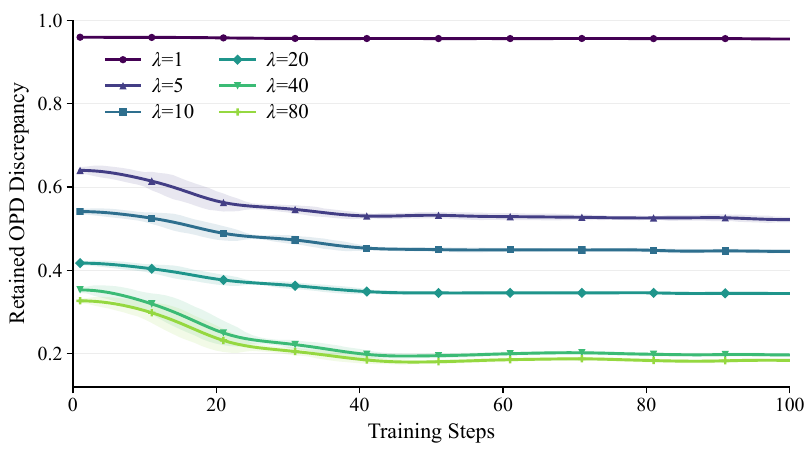}
        \\
        \small\textbf{(a) Impact of $\boldsymbol{\lambda}$ on Performance}
        &
        \small\textbf{(b) Impact of $\boldsymbol{\lambda}$ on Retained OPD Discrepancy}
    \end{tabular}

    \vspace{-1mm}
    \caption{
        Effect of the relaxation factor $\lambda$ on Cal-OPD performance and retained discrepancy ratio.
    }
    \label{fig:cal_lambda_analysis}
\end{figure}

\paragraph{Effect of the Relaxation Factor $\boldsymbol{\lambda}$.}
Figure~\ref{fig:cal_lambda_analysis} shows that Cal-OPD achieves optimal performance at $\lambda=5$, retaining approximately 52\% of the teacher--student discrepancy. Increasing $\lambda$ beyond 5 sharply degrades downstream performance, eventually dropping 1.7 points below the $\lambda=1$ baseline at $\lambda=80$. As $\lambda$ increases, the estimated TSD region expands too broadly, reducing the retained discrepancy to roughly 20\% and over-filtering task-relevant supervision. We further compare against a TSD-threshold filtering baseline matched in retained teacher--student discrepancy in Appendix~\ref{app:retention_matched_control}. Cal-OPD remains stronger, indicating that its gains are not due to signal attenuation alone.

\section{Conclusion}

We show that the teacher--student discrepancy used by standard OPD contains substantial teacher self-deviation (TSD), which emerges even without task-specific knowledge, remains largely insensitive to intervention semantics and correctness, and concentrates on surface-form tokens. Based on these findings, we introduce Calibrated On-Policy Distillation (Cal-OPD), which estimates the teacher's self-deviation region and removes the TSD-explained portion of the OPD signal before optimization. Across different teacher--student scales and mathematical reasoning benchmarks, Cal-OPD consistently outperforms standard OPD and its variants while retaining only 52--65\% of the original discrepancy and mitigating response-length expansion during training. These results suggest that effective on-policy distillation depends not on indiscriminately learning more teacher--student discrepancy, but on identifying which discrepancy is truly worth learning.
\bibliography{cal_opd_arxiv}
\bibliographystyle{plainnat}

\newpage

\appendix
\section{Appendix}

\subsection{Related Work}

\paragraph{Knowledge Distillation and On-Policy Distillation.}
Knowledge distillation (KD) transfers capabilities from a stronger teacher to a weaker student and has become a standard framework for model compression and capability transfer\citep{hinton2015distilling}. Sequence-level and pretrained language-model distillation extended this paradigm to generation and compression\citep{kim2016sequence,sanh2019distilbert,jiao2020tinybert}, while LLM-specific methods introduced reverse or student-aware divergence objectives and output-space alignment\citep{gu2024minillm,ko2024distillm,zhang2024dual}. For reasoning, teacher-generated rationales and reasoning strategies have been transferred through Fine-tune-CoT\citep{ho2023large}, Distilling Step-by-Step\citep{hsieh2023distilling}, SCoTD\citep{li2023symbolic}, MCC-KD\citep{chen2023mcc}, Orca~2\citep{mitra2023orca}, counterfactual distillation\citep{feng2024teaching}, and decomposed CoT distillation\citep{dai2024improve}; DeepSeek-R1 further demonstrates that CoT reasoning behaviors can be distilled into smaller models\citep{guo2025deepseek}.

Many sequence- and rationale-level distillation methods remain off-policy, training on teacher-generated or otherwise fixed trajectories. On-policy distillation instead provides teacher supervision on trajectories sampled from the student itself, reducing the training--inference distribution mismatch\citep{agarwal2024policy}, and has been adopted in large-scale reasoning post-training such as Qwen3\citep{yang2025qwen3}. Recent extensions modify the supervision signal through reward extrapolation\citep{yang2026learning}, entropy- or uncertainty-aware objectives\citep{jin2026entropy,ke2026respecting}, outcome-guided calibration\citep{hou2026uni}, divergence-adaptive supervision horizons\citep{hou2026dash}, and teacher-versus-base delta signals\citep{heo2026policy}. On-policy self-distillation further removes the need for a separate stronger teacher through privileged or behavior-conditioned self-teachers\citep{zhao2026self,sang2026policy}, self-generated supervision\citep{tan2026self}, and internal consistency\citep{li2026policy}, while on-policy context distillation internalizes teacher-only context such as prior experience or optimized system prompts\citep{ye2026policy}.

\paragraph{Privileged and Context-Augmented Distillation.}
Recent LLM distillation strengthens the teacher with information unavailable to the student at inference time. On-Policy Self-Distillation (OPSD) conditions a self-teacher on verified reference solutions\citep{zhao2026self}, SDPO conditions the self-teacher on environment feedback\citep{hubotter2026reinforcement}, and privileged-information distillation studies forms of training-only information\citep{penaloza2026privileged}. Related methods distill teacher-only experience or system prompts\citep{ye2026policy}, document-grounded evidence\citep{stein2026gates}, or behavioral instructions for reasoning compression\citep{sang2026policy}. For mathematical reasoning, Anti-Self-Distillation reverses harmful privileged-teacher guidance\citep{shen2026anti}, DOPD dynamically routes token-level supervision to mitigate privilege illusion\citep{yu2026dopd}, PHF transfers privileged hidden-state dynamics\citep{li2026phf}, AR-OPD anchors privileged guidance to a locally compatible view\citep{zhang2026beyond}, Purified OPSD removes reference-induced non-transferable components\citep{shen2026purified}, and DAPD introduces dual anchoring against information asymmetry\citep{wu2026dapd}. Recent analyses further show that privileged context can degrade thinking models\citep{kaur2026rethinking}, that references from other problems can retain comparable gains\citep{ichihara2026privileged}, that correct references do not provide consistent benefits\citep{shrestha2026rethinking}, and that structured privileged information can outperform raw solution traces\citep{zhao2026more}. Together, these findings suggest that privileged context changes not only the information available to the teacher, but also the teacher behavior induced by that context, complicating the interpretation of privileged teacher likelihoods as transferable knowledge.

\paragraph{Reliability of Teacher Supervision.}
Recent work has also questioned the assumption that teacher supervision is uniformly reliable across tokens and trajectories. Entropy-Aware OPD adapts distillation to teacher uncertainty\citep{jin2026entropy}, Uni-OPD calibrates teacher guidance using outcome-level order consistency\citep{hou2026uni}, and Position-Weighted OPSD shows that teacher-token reliability is strongly structured by reasoning position\citep{liu2026teacher}. OGLS-SD calibrates privileged teacher logits using outcome contrast\citep{yang2026ogls}, while BRTS selects teacher rollouts according to correctness and student alignment\citep{zhang2026policy}. Related studies further show that reasoning tokens remain relatively stable while stylistic or surface-form tokens are more sensitive to contextual interventions\citep{pan2026rlcsd,he2026not}, and that privileged context can induce shortcut behavior\citep{tian2026vicur}. More recently, \citet{ding2026does} reveal substantial, scale-dependent noise in OPD teacher supervision and show that removing noisy supervision can leave student performance largely unchanged. These studies establish that teacher supervision is heterogeneous, but characterize reliability primarily through uncertainty, outcomes, positions, trajectory quality, or specific privileged-reference effects. In contrast, the variation of teacher likelihoods under controlled contextual interventions has not been explicitly modeled as a finite-intervention self-deviation region for calibrating the teacher--student discrepancy in OPD.

\subsection{Data Collection and Preparation for TSD Analysis}
\label{app:tsd_data}

Our TSD analysis is conducted on a collection of student-generated reasoning trajectories derived from DAPO-17K~\citep{yu2026dapo}. Since DAPO-17K provides problems and ground-truth answers but does not include reference solutions, we first construct a solution-augmented subset, then generate on-policy student rollouts, and finally rescore the fixed trajectories under different teacher-side contextual interventions. Table~\ref{tab:tsd_data_config} summarizes the configurations used in this process.

\paragraph{Reference-Solution Preparation.}
We use Qwen3-235B-A22B-Instruct-2507~\citep{yang2025qwen3} to generate one reference solution for each problem in DAPO-17K. The model is used in its no-thinking mode with a maximum response length of 12,288 tokens, temperature $1.0$, and top-$p$ $1.0$. All generations use the following shared system prompt:
\begin{center}
\setlength{\fboxsep}{6pt}
\setlength{\fboxrule}{0.4pt}
\fbox{%
\begin{minipage}{0.94\linewidth}
\footnotesize
\textbf{System Prompt}
\vspace{3pt}

\ttfamily
You are a helpful math assistant. Please solve the math problem.

You must enclose your final answer exactly within \textbackslash boxed\{\}.
\end{minipage}%
}
\end{center}
We extract the final answer enclosed in \texttt{\textbackslash boxed\{\}} and compare it with the ground-truth answer provided by DAPO-17K. Solutions with incorrect final answers are discarded. After filtering, we retain 15,560 question--solution pairs, covering $90.18\%$ of the 17,255 problems, with 1,695 problems unsolved. The retained reference solutions contain an average of 7,636 characters, a median of 5,679 characters, a 95th percentile of 19,493 characters, and a range of 445--39,695 characters.

\paragraph{Student Rollout Collection.}
Starting from the 15,560 retained questions, we generate one on-policy response per question using Qwen3-1.7B~\citep{yang2025qwen3} in thinking mode. We set the maximum response length to 16,384 tokens, temperature to $1.0$, and top-$p$ to $1.0$, while using the same system prompt as above. Rollout generation is stopped once the cumulative number of student-generated response tokens reaches approximately 60 million. This results in 6,528 question--response pairs used for subsequent TSD analysis. The student responses contain an average of 9,270 tokens and a median of 7,758 tokens.

\paragraph{Training Data Usage.}
All distillation methods in our main experiments are trained on the same 15,560 question--solution pairs.

\begin{table}[t]
\centering
\footnotesize
\setlength{\tabcolsep}{4pt}
\renewcommand{\arraystretch}{1.10}

\begin{tabularx}{\linewidth}{
    @{}
    >{\raggedright\arraybackslash}p{0.19\linewidth}
    >{\raggedright\arraybackslash}X
    >{\centering\arraybackslash}p{0.13\linewidth}
    >{\centering\arraybackslash}p{0.12\linewidth}
    >{\centering\arraybackslash}p{0.09\linewidth}
    >{\centering\arraybackslash}p{0.09\linewidth}
    @{}
}
\toprule
\textbf{Stage}
&
\textbf{Model}
&
\textbf{Mode}
&
\textbf{Max Length}
&
\textbf{Temp.}
&
\textbf{Top-$p$}
\\
\midrule

Solution generation
&
Qwen3-235B-A22B-Instruct-2507
&
No-thinking
&
12,288
&
1.0
&
1.0
\\

Student rollout
&
Qwen3-1.7B
&
Thinking
&
16,384
&
1.0
&
1.0
\\

Teacher rescoring
&
Qwen3-8B
&
Thinking
&
Fixed rollout
&
--
&
--
\\

\bottomrule
\end{tabularx}

\caption{
Configurations used for preparing the data for TSD analysis. Solution generation and student rollout collection use one generation per question. Teacher rescoring does not involve sampling, since the student trajectory is kept fixed.
}
\label{tab:tsd_data_config}
\end{table}

\begin{figure}[!t]
    \centering

    \noindent
    \includegraphics[width=0.325\linewidth]
    {images/qwen3_8b_tsd_sensitive_token_fractions_tau_01.pdf}%
    \hfill
    \includegraphics[width=0.325\linewidth]
    {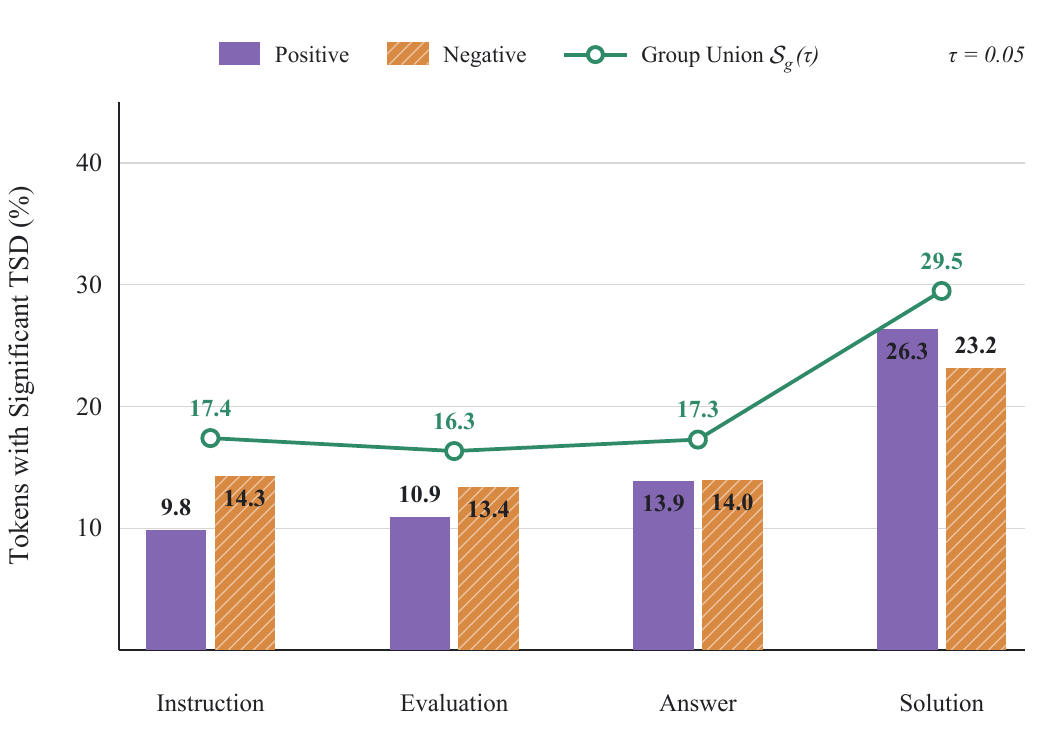}%
    \hfill
    \includegraphics[width=0.325\linewidth]
    {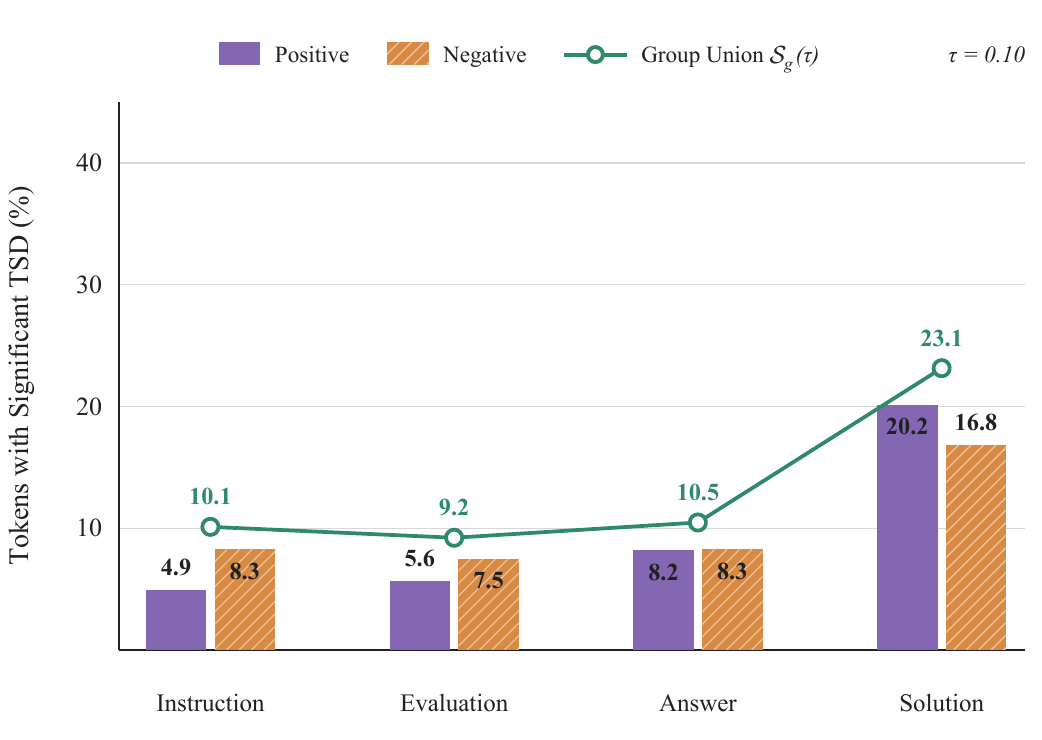}

    \vspace{-0.8mm}

    {\small\textbf{(a)} Qwen3-8B\par}

    \vspace{2.0mm}

    \noindent
    \includegraphics[width=0.325\linewidth]
    {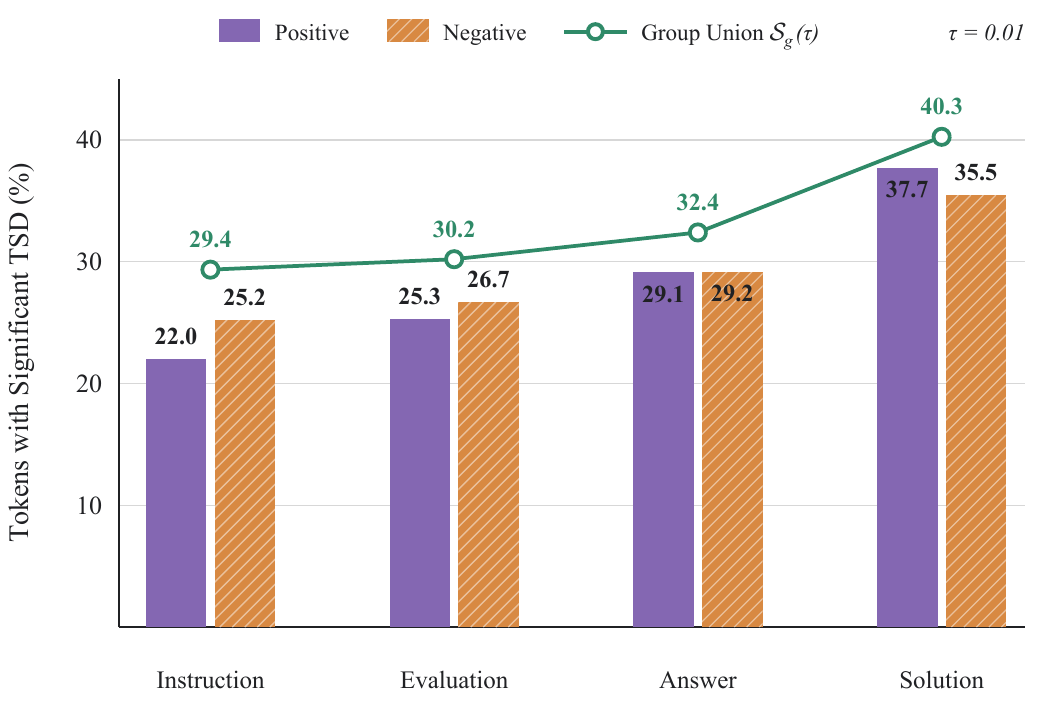}%
    \hfill
    \includegraphics[width=0.325\linewidth]
    {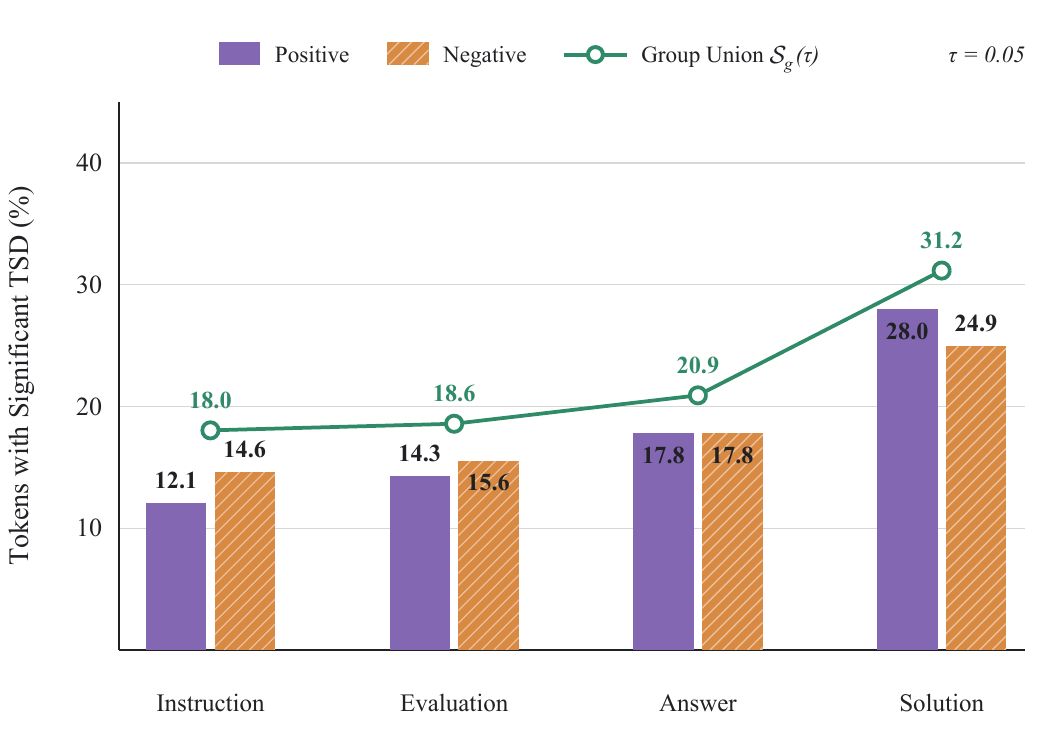}%
    \hfill
    \includegraphics[width=0.325\linewidth]
    {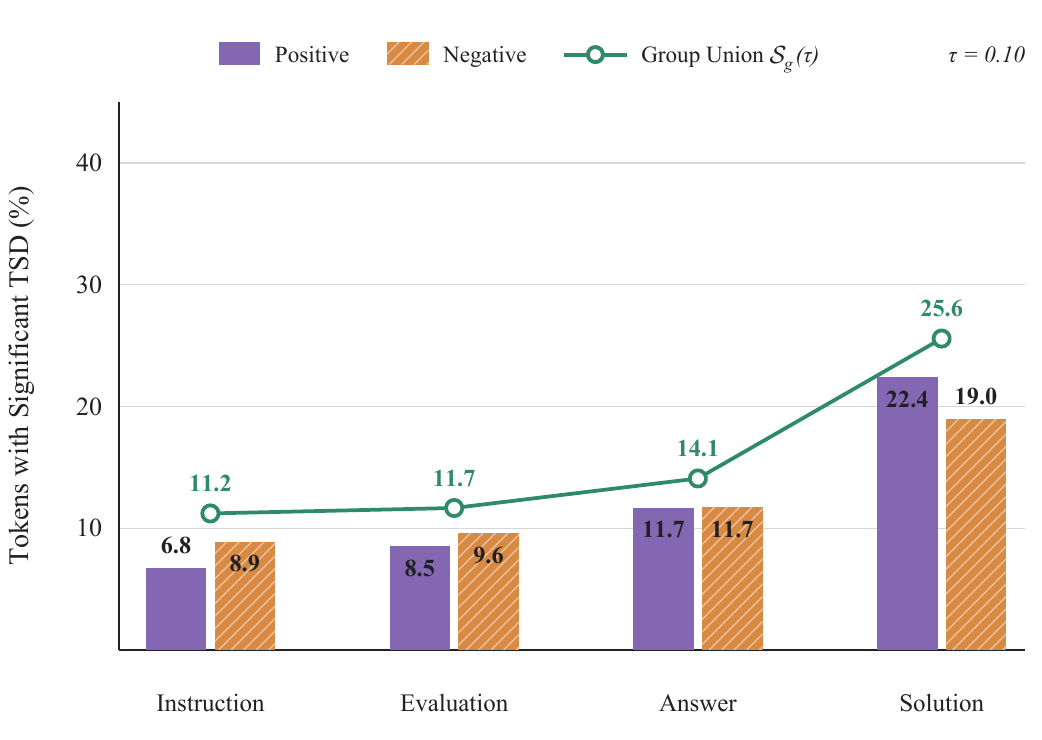}

    \vspace{-0.8mm}

    {\small\textbf{(b)} Qwen3-4B-Thinking-2507\par}

    \vspace{0.5mm}

    \caption{
        Prevalence of significant TSD across thresholds and teacher models.
        Although the absolute prevalence decreases with stricter thresholds,
        substantial TSD continues to emerge under task-agnostic interventions,
        while solution-level privilege consistently broadens the affected set.
    }
    \label{fig:app_tsd_prevalence}
\end{figure}

\begin{figure}[!t]
    \centering

    \noindent
    \includegraphics[width=0.325\linewidth]
    {images/qwen3_8b_tsd_group_retention_tau_01.pdf}%
    \hfill
    \includegraphics[width=0.325\linewidth]
    {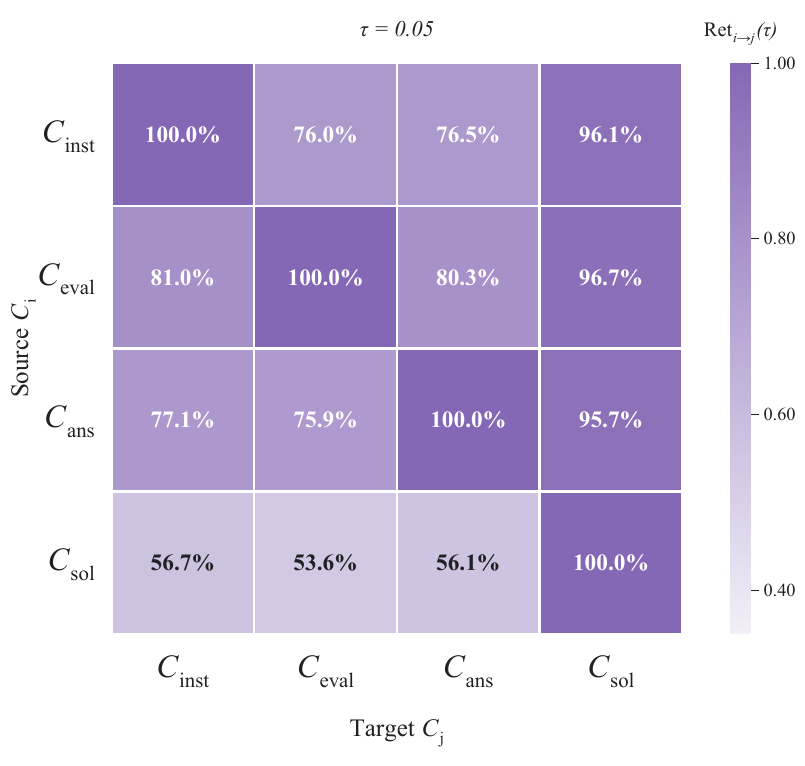}%
    \hfill
    \includegraphics[width=0.325\linewidth]
    {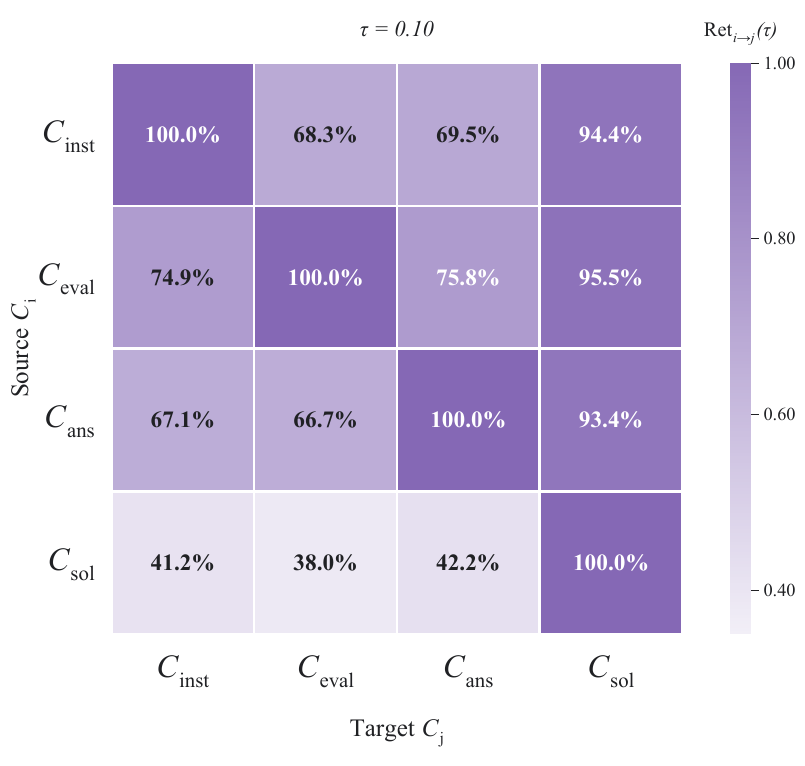}

    \vspace{-0.8mm}

    {\small\textbf{(a)} Qwen3-8B\par}

    \vspace{2.0mm}

    \noindent
    \includegraphics[width=0.325\linewidth]
    {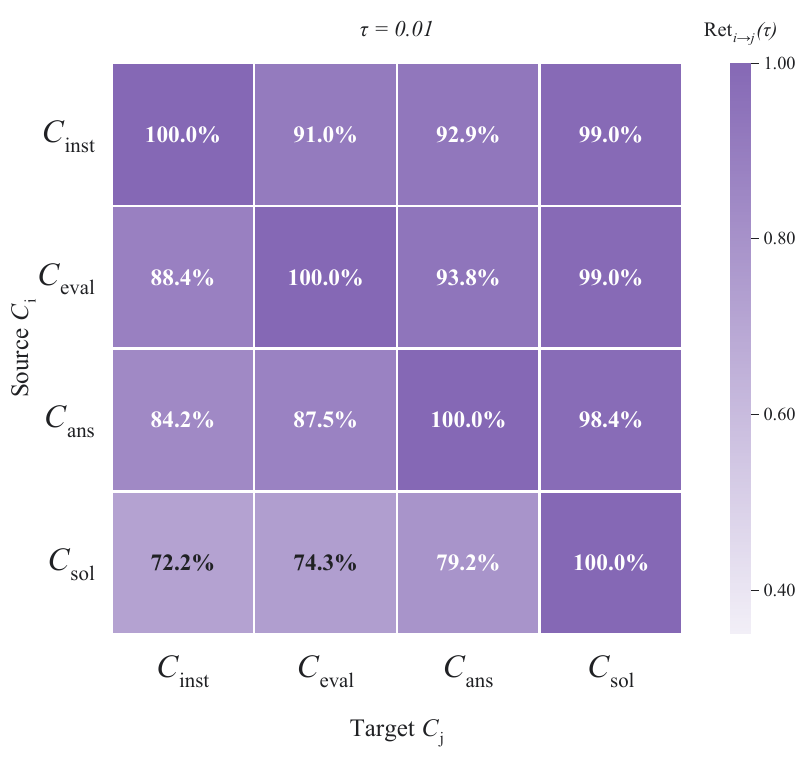}%
    \hfill
    \includegraphics[width=0.325\linewidth]
    {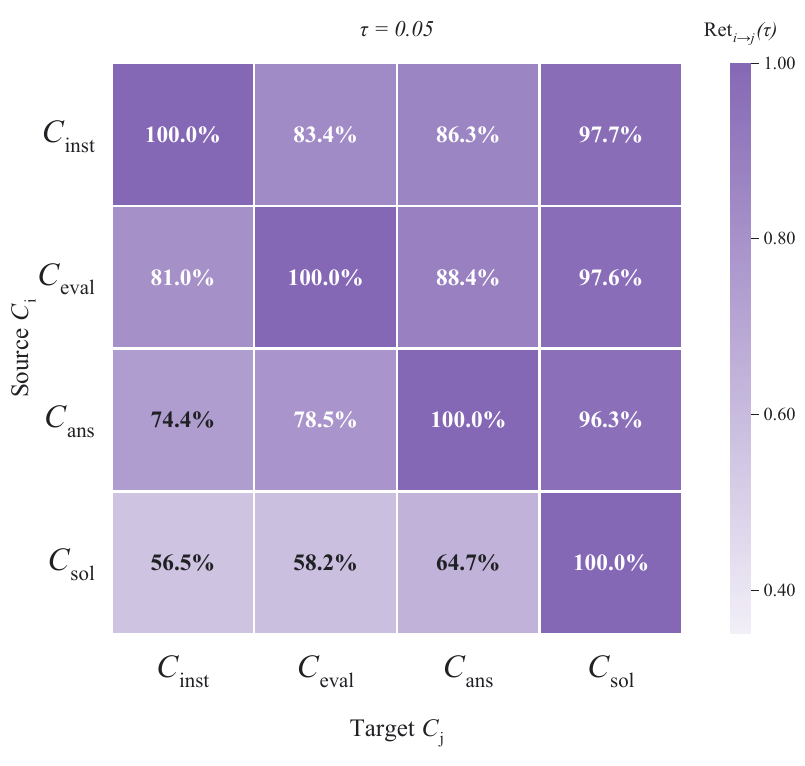}%
    \hfill
    \includegraphics[width=0.325\linewidth]
    {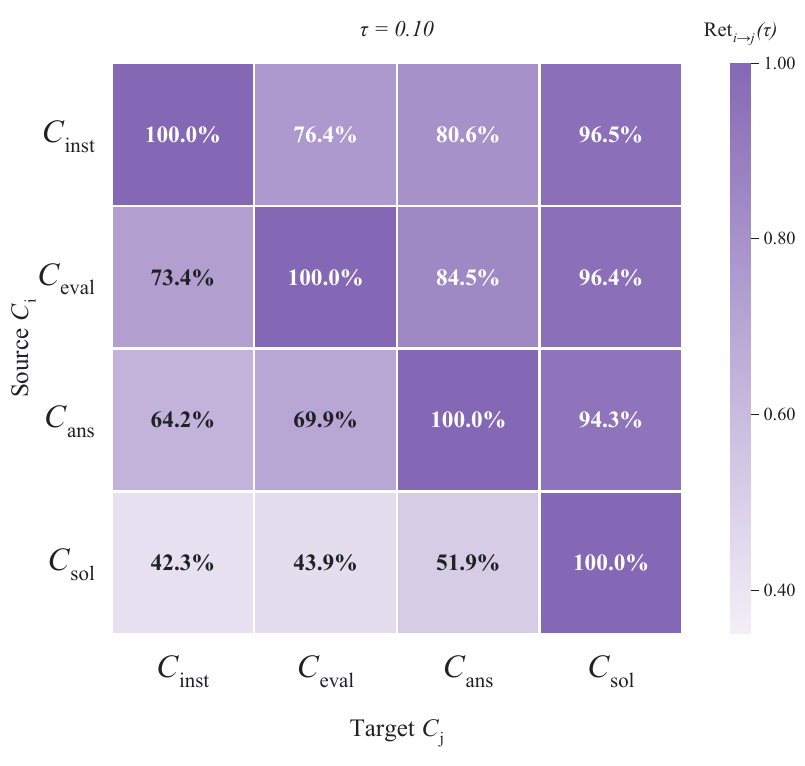}

    \vspace{-0.8mm}

    {\small\textbf{(b)} Qwen3-4B-Thinking-2507\par}

    \vspace{0.5mm}

    \caption{
        Retention of significant TSD across thresholds and teacher models.
        Less informative interventions retain nearly all affected token positions
        under solution-level privilege, whereas retention in the reverse direction
        is substantially lower.
    }
    \label{fig:app_tsd_retention}
\end{figure}

\paragraph{Construction of Privileged Contexts.}
For the answer-level interventions in Table~\ref{tab:contextual_interventions}, the positive variant uses the ground-truth answer associated with the current problem. To construct $c_{\mathrm{ans}}^{\mathrm{neg}}$, we randomly generate an incorrect answer with the same number of digits as the corresponding correct answer, thereby controlling for differences in answer length.

For the solution-level interventions, $c_{\mathrm{sol}}^{\mathrm{pos}}$ uses the verified reference solution generated for the current problem. To construct $c_{\mathrm{sol}}^{\mathrm{neg}}$, we randomly select the reference solution of a different problem whose tokenized length differs from that of the current reference solution by less than $10\%$. This length-matching constraint reduces the possibility that differences between positive and negative solution-level interventions are driven primarily by contextual length rather than content.

\begin{figure}[t]
    \centering

    \noindent
    \includegraphics[width=0.325\linewidth]
    {images/qwen3_8b_tsd_semantic_metrics_tau_01.pdf}%
    \hfill
    \includegraphics[width=0.325\linewidth]
    {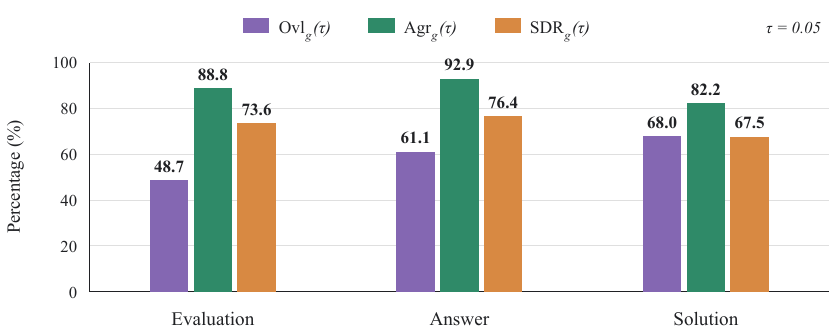}%
    \hfill
    \includegraphics[width=0.325\linewidth]
    {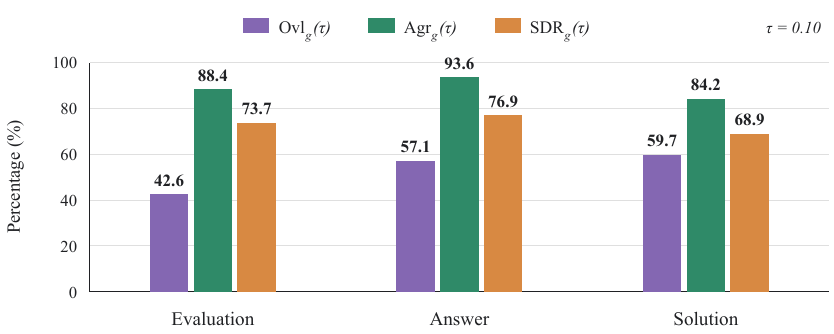}

    \vspace{-0.8mm}

    {\small\textbf{(a)} Qwen3-8B\par}

    \vspace{2.0mm}

    \noindent
    \includegraphics[width=0.325\linewidth]
    {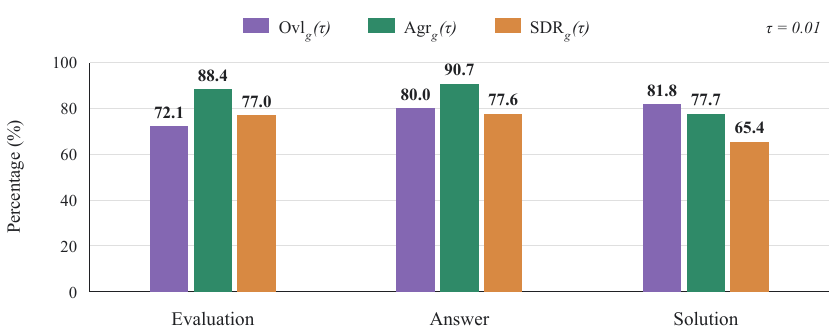}%
    \hfill
    \includegraphics[width=0.325\linewidth]
    {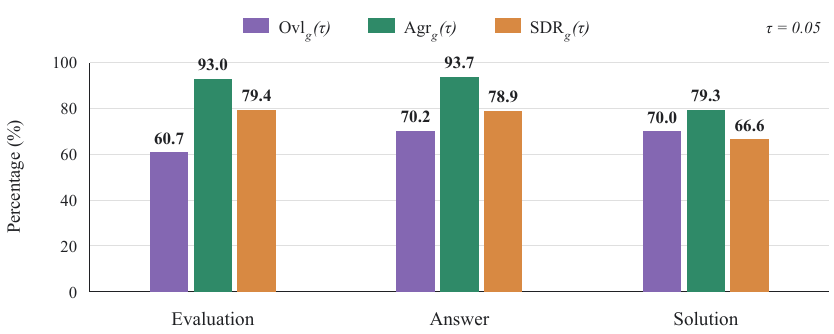}%
    \hfill
    \includegraphics[width=0.325\linewidth]
    {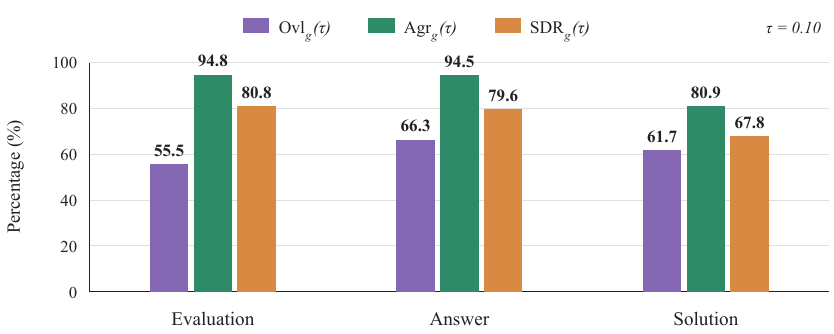}

    \vspace{-0.8mm}

    {\small\textbf{(b)} Qwen3-4B-Thinking-2507\par}

    \vspace{0.5mm}

    \caption{
        Semantic consistency of TSD across thresholds and teacher models.
        Contrasting interventions consistently exhibit substantial positional overlap,
        high directional agreement, and predominantly shared deviation across
        thresholds and model scales.
    }
    \label{fig:app_tsd_semantic}
\end{figure}

\paragraph{Teacher Likelihood Rescoring.}
We use Qwen3-8B~\citep{yang2025qwen3} as the teacher and evaluate each fixed Qwen3-1.7B rollout under nine teacher contexts:
\begin{equation}
\begin{gathered}
    c_0,\quad
    c_{\mathrm{inst}}^{\mathrm{pos}},\;
    c_{\mathrm{inst}}^{\mathrm{neg}},\
    c_{\mathrm{eval}}^{\mathrm{pos}},\;
    c_{\mathrm{eval}}^{\mathrm{neg}},\
    c_{\mathrm{ans}}^{\mathrm{pos}},\;
    c_{\mathrm{ans}}^{\mathrm{neg}},\
    c_{\mathrm{sol}}^{\mathrm{pos}},\;
    c_{\mathrm{sol}}^{\mathrm{neg}}.
\end{gathered}
\end{equation}
The shared system prompt above is included in every condition and is therefore not itself treated as a contextual intervention. For all nine contexts, the problem $x$, student rollout $y$, prefix $y_{<t}$, and evaluated token $y_t$ are kept identical; only the additional teacher-side context is changed. The teacher does not generate new trajectories. Instead, we recompute the token-level log-likelihood assigned to every student-generated token under each context, allowing the resulting likelihood differences to be attributed directly to the contextual intervention.

Consequently, the TSD analysis is based on approximately 60 million student-generated tokens, with each token rescored under the baseline context and eight contextual interventions.

\subsection{TSD Semantic Consistency Across Thresholds and Model Scales}
\label{app:tsd_semantic_consistency}

We further examine whether the TSD patterns reported in the main text persist across significance thresholds and teacher scales. Using the same Qwen3-1.7B student trajectories described in Appendix~\ref{app:tsd_data}, we repeat the analysis with Qwen3-8B and Qwen3-4B-Thinking-2507 as teachers under $\tau\in\{0.01,0.05,0.10\}$. We consider the prevalence of significant TSD, retention across intervention groups, and semantic consistency across the six teacher--threshold configurations.

\paragraph{Prevalence of Significant TSD.}
Figure~\ref{fig:app_tsd_prevalence} shows that increasing $\tau$ reduces the absolute prevalence of significant TSD, while preserving the relative pattern across intervention groups. For Qwen3-8B, the union prevalence under task-agnostic instructions decreases from $29.8\%$ at $\tau=0.01$ to $10.1\%$ at $\tau=0.10$, yet remains comparable to evaluative feedback at $29.1\%$ and $9.2\%$, respectively. The same pattern holds for Qwen3-4B-Thinking-2507, where task-agnostic instructions remain close to evaluative feedback across all thresholds. Solution-level privilege consistently produces the highest prevalence, supporting the conclusion that richer privileged context broadens the affected set rather than being necessary for TSD to emerge.

\paragraph{Retention Across Intervention Groups.}
Figure~\ref{fig:app_tsd_retention} exhibits a consistent asymmetric retention pattern across thresholds and teacher models. Token positions affected under instruction, evaluation, and answer interventions are largely retained under solution-level privilege. Even at $\tau=0.10$, their retention into the solution group remains at least $93.4\%$ for Qwen3-8B and $94.3\%$ for Qwen3-4B-Thinking-2507. The reverse retention is substantially lower and decreases as the threshold becomes stricter. This asymmetry indicates that richer privileged context predominantly extends an existing set of TSD-sensitive positions rather than replacing it with a distinct deviation pattern.

\paragraph{Semantic Consistency.}
Figure~\ref{fig:app_tsd_semantic} shows that the semantic-consistency pattern also persists across thresholds and teacher models, and across all three intervention groups. As $\tau$ increases, positional overlap generally decreases because fewer tokens remain significant under both contrasting interventions, while directional agreement remains consistently high. At $\tau=0.10$, answer-level agreement reaches $93.6\%$ for Qwen3-8B and $94.5\%$ for Qwen3-4B-Thinking-2507, with corresponding SDR values of $76.9\%$ and $79.6\%$. Evaluative interventions show the same trend, whereas solution-level interventions consistently exhibit higher positional overlap but lower SDR. Thus, stricter thresholds change which tokens remain significant without altering the broader finding that contrasting semantics induce strongly aligned and predominantly shared TSD.

Across all configurations, the qualitative conclusions of the main analysis remain unchanged. TSD emerges substantially without task-specific knowledge, richer privileged context largely preserves and expands previously affected token positions, and contrasting intervention semantics leave much of the resulting TSD directionally aligned and shared. \textbf{These results show that the observed TSD patterns are not specific to a particular significance threshold or teacher scale.}

\begin{table}[!t]
\caption{
Top-24 and bottom-24 token forms under
$c_{\mathrm{inst}}^{\mathrm{pos}}$ and
$c_{\mathrm{inst}}^{\mathrm{neg}}$ at $\tau=0.01$.
The final row reports unweighted means of
$\rho_{\tau}$ and $\overline{|\Delta|}$ over each 24-token subset.
}
\label{tab:app_tsd_tokens_inst}

\centering
\scriptsize
\setlength{\tabcolsep}{0pt}
\renewcommand{\arraystretch}{1.04}

\begin{tabular}{
@{}
r
@{\hspace{5pt}}
l @{\hspace{5pt}} r @{\hspace{9pt}} r
@{\hspace{10pt}}
l @{\hspace{5pt}} r @{\hspace{9pt}} r
@{\hspace{15pt}}
l @{\hspace{5pt}} r @{\hspace{9pt}} r
@{\hspace{10pt}}
l @{\hspace{5pt}} r @{\hspace{9pt}} r
@{}
}
\toprule
& \multicolumn{6}{c}{$c_{\mathrm{inst}}^{\mathrm{pos}}$}
& \multicolumn{6}{c}{$c_{\mathrm{inst}}^{\mathrm{neg}}$} \\
\cmidrule(lr){2-7}\cmidrule(lr){8-13}

\#
& \multicolumn{3}{c}{\textbf{Highest $\rho_{\tau}$}}
& \multicolumn{3}{c}{\textbf{Lowest $\rho_{\tau}$}}
& \multicolumn{3}{c}{\textbf{Highest $\rho_{\tau}$}}
& \multicolumn{3}{c}{\textbf{Lowest $\rho_{\tau}$}} \\
\cmidrule(lr){2-4}\cmidrule(lr){5-7}
\cmidrule(lr){8-10}\cmidrule(lr){11-13}

& \textbf{Token}
& {$\boldsymbol{\rho_{\tau}}$ (\%)}
& {$\mathbf{\overline{|\Delta|}}$}
& \textbf{Token}
& {$\boldsymbol{\rho_{\tau}}$ (\%)}
& {$\mathbf{\overline{|\Delta|}}$}
& \textbf{Token}
& {$\boldsymbol{\rho_{\tau}}$ (\%)}
& {$\mathbf{\overline{|\Delta|}}$}
& \textbf{Token}
& {$\boldsymbol{\rho_{\tau}}$ (\%)}
& {$\mathbf{\overline{|\Delta|}}$} \\
\midrule

1  & \texttt{maybe}      & 74.4 & 0.066
   & \texttt{}           & 1.0  & 0.069
   & \texttt{maybe}      & 84.4 & 0.110
   & \texttt{}           & 1.5  & 0.080 \\

2  & \texttt{consider}   & 73.1 & 0.071
   & \texttt{\#\#\#}     & 2.1  & 0.071
   & \texttt{consider}   & 84.2 & 0.106
   & \texttt{0}          & 2.9  & 0.116 \\

3  & \texttt{try}        & 66.1 & 0.086
   & \texttt{0}          & 2.2  & 0.095
   & \texttt{however}    & 82.8 & 0.157
   & \texttt{\#\#\#}     & 3.4  & 0.082 \\

4  & \texttt{however}    & 65.2 & 0.086
   & $^\circ$            & 2.6  & 0.068
   & \texttt{therefore}  & 82.1 & 0.170
   & $^\circ$            & 3.7  & 0.083 \\

5  & \texttt{since}      & 64.4 & 0.080
   & $\surd$             & 2.9  & 0.084
   & \texttt{earlier}    & 80.9 & 0.121
   & \texttt{\_}         & 3.7  & 0.089 \\

6  & \texttt{if}         & 63.9 & 0.072
   & \texttt{\_}         & 3.0  & 0.072
   & \texttt{try}        & 80.9 & 0.177
   & \texttt{9}          & 3.9  & 0.119 \\

7  & \texttt{earlier}    & 63.2 & 0.066
   & \texttt{9}          & 3.1  & 0.099
   & \texttt{now}        & 78.5 & 0.165
   & $\surd$             & 4.0  & 0.120 \\

8  & \texttt{now}        & 63.0 & 0.094
   & $\theta$            & 3.2  & 0.076
   & \texttt{how}        & 75.1 & 0.112
   & \texttt{8}          & 4.1  & 0.124 \\

9  & \texttt{how}        & 62.8 & 0.078
   & \texttt{8}          & 3.3  & 0.102
   & \texttt{because}    & 74.6 & 0.114
   & \texttt{frac}       & 4.2  & 0.073 \\

10 & \texttt{think}      & 61.8 & 0.070
   & \texttt{frac}       & 3.3  & 0.062
   & \texttt{another}    & 74.4 & 0.127
   & \texttt{6}          & 4.3  & 0.125 \\

11 & \texttt{another}    & 60.8 & 0.088
   & \texttt{6}          & 3.4  & 0.101
   & \texttt{if}         & 74.2 & 0.102
   & $\theta$            & 4.4  & 0.105 \\

12 & \texttt{check}      & 60.6 & 0.075
   & \texttt{7}          & 3.5  & 0.101
   & \texttt{since}      & 74.1 & 0.109
   & \texttt{7}          & 4.4  & 0.124 \\

13 & \texttt{because}    & 60.2 & 0.080
   & \texttt{5}          & 3.6  & 0.096
   & \texttt{think}      & 73.9 & 0.108
   & \texttt{5}          & 4.5  & 0.119 \\

14 & \texttt{therefore}  & 58.9 & 0.088
   & \texttt{\char123}   & 3.6  & 0.067
   & \texttt{let}        & 73.6 & 0.122
   & \texttt{\_i}        & 4.6  & 0.090 \\

15 & \texttt{when}       & 58.9 & 0.078
   & \texttt{2}          & 3.8  & 0.089
   & \texttt{so}         & 72.7 & 0.118
   & \texttt{\char123}   & 4.7  & 0.088 \\

16 & \texttt{there}      & 58.7 & 0.073
   & \texttt{\_i}        & 3.8  & 0.077
   & \texttt{but}        & 72.0 & 0.113
   & \texttt{2}          & 4.8  & 0.110 \\

17 & \texttt{let}        & 58.0 & 0.076
   & \texttt{)\char94}   & 4.0  & 0.078
   & \texttt{this}       & 71.8 & 0.117
   & \texttt{4}          & 5.1  & 0.119 \\

18 & \texttt{given}      & 57.2 & 0.105
   & \texttt{+}          & 4.0  & 0.089
   & \texttt{altern.}    & 71.7 & 0.123
   & \texttt{1}          & 5.3  & 0.109 \\

19 & \texttt{also}       & 56.8 & 0.083
   & \texttt{4}          & 4.0  & 0.095
   & \texttt{here}       & 71.2 & 0.115
   & \texttt{)\char94}   & 5.4  & 0.091 \\

20 & \texttt{use}        & 56.1 & 0.081
   & \texttt{1}          & 4.2  & 0.088
   & \texttt{when}       & 71.1 & 0.110
   & \texttt{3}          & 5.5  & 0.112 \\

21 & \texttt{so}         & 55.8 & 0.081
   & \texttt{3}          & 4.2  & 0.092
   & \texttt{there}      & 69.2 & 0.099
   & \texttt{+}          & 5.6  & 0.106 \\

22 & \texttt{but}        & 55.7 & 0.077
   & \texttt{'t}         & 4.6  & 0.088
   & \texttt{previous}   & 69.2 & 0.086
   & \texttt{'t}         & 5.7  & 0.101 \\

23 & \texttt{we}         & 55.2 & 0.065
   & \texttt{)/}         & 5.1  & 0.077
   & \texttt{it}         & 69.1 & 0.117
   & \texttt{cdot}       & 7.2  & 0.057 \\

24 & \texttt{yes}        & 55.1 & 0.096
   & \texttt{cdot}       & 5.7  & 0.051
   & \texttt{seems}      & 68.8 & 0.097
   & \texttt{)/}         & 7.4  & 0.100 \\

\midrule
\multicolumn{1}{c}{\textbf{Mean}}
& -- & 61.1 & 0.080
& -- & 3.5  & 0.083
& -- & 75.0 & 0.121
& -- & 4.6  & 0.102 \\
\bottomrule
\end{tabular}
\end{table}

\begin{table}[!t]
\caption{
Top-24 and bottom-24 token forms under
$c_{\mathrm{eval}}^{\mathrm{pos}}$ and
$c_{\mathrm{eval}}^{\mathrm{neg}}$ at $\tau=0.01$.
The final row reports unweighted means of
$\rho_{\tau}$ and $\overline{|\Delta|}$ over each 24-token subset.
}
\label{tab:app_tsd_tokens_eval}

\centering
\scriptsize
\setlength{\tabcolsep}{0pt}
\renewcommand{\arraystretch}{1.04}

\begin{tabular}{
@{}
r
@{\hspace{5pt}}
l @{\hspace{5pt}} r @{\hspace{9pt}} r
@{\hspace{10pt}}
l @{\hspace{5pt}} r @{\hspace{9pt}} r
@{\hspace{15pt}}
l @{\hspace{5pt}} r @{\hspace{9pt}} r
@{\hspace{10pt}}
l @{\hspace{5pt}} r @{\hspace{9pt}} r
@{}
}
\toprule
& \multicolumn{6}{c}{$c_{\mathrm{eval}}^{\mathrm{pos}}$}
& \multicolumn{6}{c}{$c_{\mathrm{eval}}^{\mathrm{neg}}$} \\
\cmidrule(lr){2-7}\cmidrule(lr){8-13}

\#
& \multicolumn{3}{c}{\textbf{Highest $\rho_{\tau}$}}
& \multicolumn{3}{c}{\textbf{Lowest $\rho_{\tau}$}}
& \multicolumn{3}{c}{\textbf{Highest $\rho_{\tau}$}}
& \multicolumn{3}{c}{\textbf{Lowest $\rho_{\tau}$}} \\
\cmidrule(lr){2-4}\cmidrule(lr){5-7}
\cmidrule(lr){8-10}\cmidrule(lr){11-13}

& \textbf{Token}
& {$\boldsymbol{\rho_{\tau}}$ (\%)}
& {$\mathbf{\overline{|\Delta|}}$}
& \textbf{Token}
& {$\boldsymbol{\rho_{\tau}}$ (\%)}
& {$\mathbf{\overline{|\Delta|}}$}
& \textbf{Token}
& {$\boldsymbol{\rho_{\tau}}$ (\%)}
& {$\mathbf{\overline{|\Delta|}}$}
& \textbf{Token}
& {$\boldsymbol{\rho_{\tau}}$ (\%)}
& {$\mathbf{\overline{|\Delta|}}$} \\
\midrule

1  & \texttt{maybe}      & 77.5 & 0.079
   & \texttt{\char125\char123} & 1.0 & 0.072
   & \texttt{maybe}      & 83.8 & 0.100
   & \texttt{\char125\char123} & 1.1 & 0.076 \\

2  & \texttt{consider}   & 76.2 & 0.074
   & \texttt{\#\#\#}     & 1.1 & 0.047
   & \texttt{however}    & 83.2 & 0.139
   & \texttt{\#\#\#}     & 1.3 & 0.053 \\

3  & \texttt{earlier}    & 72.4 & 0.082
   & \texttt{0}          & 2.5 & 0.111
   & \texttt{consider}   & 82.6 & 0.091
   & \texttt{0}          & 2.9 & 0.124 \\

4  & \texttt{however}    & 72.2 & 0.102
   & \texttt{frac}       & 2.8 & 0.058
   & \texttt{earlier}    & 79.9 & 0.113
   & \texttt{frac}       & 3.1 & 0.064 \\

5  & \texttt{try}        & 69.2 & 0.091
   & $^\circ$            & 2.9 & 0.070
   & \texttt{therefore}  & 77.9 & 0.131
   & \texttt{\_}         & 3.6 & 0.092 \\

6  & \texttt{therefore}  & 67.4 & 0.098
   & $\surd$             & 3.2 & 0.093
   & \texttt{try}        & 77.1 & 0.134
   & $^\circ$            & 3.7 & 0.068 \\

7  & \texttt{now}        & 66.5 & 0.102
   & \texttt{\_}         & 3.3 & 0.076
   & \texttt{says}       & 74.6 & 0.090
   & \texttt{9}          & 3.9 & 0.138 \\

8  & \texttt{since}      & 66.2 & 0.083
   & \texttt{9}          & 3.4 & 0.114
   & \texttt{another}    & 74.5 & 0.118
   & $\surd$             & 3.9 & 0.108 \\

9  & \texttt{how}        & 66.1 & 0.081
   & $\theta$            & 3.4 & 0.082
   & \texttt{since}      & 74.3 & 0.103
   & $\theta$            & 4.1 & 0.099 \\

10 & \texttt{if}         & 66.1 & 0.077
   & \texttt{8}          & 3.6 & 0.119
   & \texttt{think}      & 73.4 & 0.093
   & \texttt{8}          & 4.1 & 0.140 \\

11 & \texttt{think}      & 65.4 & 0.074
   & \texttt{6}          & 3.7 & 0.118
   & \texttt{if}         & 73.4 & 0.090
   & \texttt{6}          & 4.3 & 0.138 \\

12 & \texttt{because}    & 65.1 & 0.086
   & \texttt{\char123}   & 3.8 & 0.074
   & \texttt{how}        & 72.9 & 0.104
   & \texttt{\_i}        & 4.4 & 0.087 \\

13 & \texttt{another}    & 64.5 & 0.091
   & \texttt{7}          & 3.9 & 0.114
   & \texttt{altern.}    & 72.8 & 0.120
   & \texttt{\char123}   & 4.4 & 0.078 \\

14 & \texttt{previous}   & 63.6 & 0.071
   & \texttt{5}          & 3.9 & 0.111
   & \texttt{previous}   & 72.6 & 0.087
   & \texttt{7}          & 4.5 & 0.138 \\

15 & \texttt{says}       & 62.8 & 0.073
   & \texttt{\_i}        & 4.0 & 0.084
   & \texttt{now}        & 72.5 & 0.149
   & \texttt{5}          & 4.5 & 0.130 \\

16 & \texttt{when}       & 62.7 & 0.084
   & \texttt{2}          & 4.2 & 0.103
   & \texttt{because}    & 72.0 & 0.105
   & \texttt{cdot}       & 4.6 & 0.053 \\

17 & \texttt{there}      & 62.2 & 0.078
   & \texttt{cdot}       & 4.4 & 0.047
   & \texttt{let}        & 71.0 & 0.133
   & \texttt{2}          & 4.9 & 0.112 \\

18 & \texttt{so}         & 62.2 & 0.087
   & \texttt{4}          & 4.5 & 0.111
   & \texttt{but}        & 70.8 & 0.100
   & \texttt{4}          & 5.0 & 0.127 \\

19 & \texttt{but}        & 61.9 & 0.082
   & \texttt{)\char94}   & 4.5 & 0.087
   & \texttt{here}       & 70.0 & 0.103
   & \texttt{)\char94}   & 5.2 & 0.091 \\

20 & \texttt{let}        & 61.9 & 0.084
   & \texttt{+}          & 4.5 & 0.099
   & \texttt{check}      & 69.9 & 0.098
   & \texttt{1}          & 5.3 & 0.112 \\

21 & \texttt{this}       & 61.1 & 0.086
   & \texttt{1}          & 4.6 & 0.103
   & \texttt{seems}      & 69.9 & 0.090
   & \texttt{3}          & 5.4 & 0.130 \\

22 & \texttt{seems}      & 60.7 & 0.073
   & \texttt{3}          & 4.8 & 0.104
   & \texttt{so}         & 69.8 & 0.106
   & \texttt{'t}         & 5.5 & 0.124 \\

23 & \texttt{also}       & 60.3 & 0.087
   & \texttt{'t}         & 5.3 & 0.091
   & \texttt{when}       & 69.3 & 0.100
   & \texttt{+}          & 5.6 & 0.106 \\

24 & \texttt{altern.}    & 60.2 & 0.085
   & \texttt{)/}         & 5.8 & 0.087
   & \texttt{there}      & 68.9 & 0.095
   & \texttt{\char125}   & 6.7 & 0.069 \\

\midrule
\multicolumn{1}{c}{\textbf{Mean}}
& -- & 65.6 & 0.084
& -- & 3.7  & 0.091
& -- & 74.0 & 0.108
& -- & 4.3  & 0.102 \\
\bottomrule
\end{tabular}
\end{table}

\begin{table}[!t]
\caption{
Top-24 and bottom-24 token forms under
$c_{\mathrm{ans}}^{\mathrm{pos}}$ and
$c_{\mathrm{ans}}^{\mathrm{neg}}$ at $\tau=0.01$.
The final row reports unweighted means of
$\rho_{\tau}$ and $\overline{|\Delta|}$ over each 24-token subset.
}
\label{tab:app_tsd_tokens_ans}

\centering
\scriptsize
\setlength{\tabcolsep}{0pt}
\renewcommand{\arraystretch}{1.04}

\begin{tabular}{
@{}
r
@{\hspace{5pt}}
l @{\hspace{5pt}} r @{\hspace{9pt}} r
@{\hspace{10pt}}
l @{\hspace{5pt}} r @{\hspace{9pt}} r
@{\hspace{15pt}}
l @{\hspace{5pt}} r @{\hspace{9pt}} r
@{\hspace{10pt}}
l @{\hspace{5pt}} r @{\hspace{9pt}} r
@{}
}
\toprule
& \multicolumn{6}{c}{$c_{\mathrm{ans}}^{\mathrm{pos}}$}
& \multicolumn{6}{c}{$c_{\mathrm{ans}}^{\mathrm{neg}}$} \\
\cmidrule(lr){2-7}\cmidrule(lr){8-13}

\#
& \multicolumn{3}{c}{\textbf{Highest $\rho_{\tau}$}}
& \multicolumn{3}{c}{\textbf{Lowest $\rho_{\tau}$}}
& \multicolumn{3}{c}{\textbf{Highest $\rho_{\tau}$}}
& \multicolumn{3}{c}{\textbf{Lowest $\rho_{\tau}$}} \\
\cmidrule(lr){2-4}\cmidrule(lr){5-7}
\cmidrule(lr){8-10}\cmidrule(lr){11-13}

& \textbf{Token}
& {$\boldsymbol{\rho_{\tau}}$ (\%)}
& {$\mathbf{\overline{|\Delta|}}$}
& \textbf{Token}
& {$\boldsymbol{\rho_{\tau}}$ (\%)}
& {$\mathbf{\overline{|\Delta|}}$}
& \textbf{Token}
& {$\boldsymbol{\rho_{\tau}}$ (\%)}
& {$\mathbf{\overline{|\Delta|}}$}
& \textbf{Token}
& {$\boldsymbol{\rho_{\tau}}$ (\%)}
& {$\mathbf{\overline{|\Delta|}}$} \\
\midrule

1  & \texttt{maybe}      & 83.3 & 0.254
   & \texttt{\char125\char123} & 1.1 & 0.070
   & \texttt{maybe}      & 83.7 & 0.362
   & \texttt{\char125\char123} & 1.1 & 0.071 \\

2  & \texttt{however}    & 82.2 & 0.210
   & \texttt{\#\#\#}     & 1.6 & 0.058
   & \texttt{however}    & 82.3 & 0.163
   & \texttt{\#\#\#}     & 1.6 & 0.055 \\

3  & \texttt{therefore}  & 80.6 & 0.210
   & \texttt{0}          & 2.8 & 0.143
   & \texttt{consider}   & 81.0 & 0.090
   & \texttt{0}          & 2.8 & 0.131 \\

4  & \texttt{consider}   & 80.1 & 0.090
   & \texttt{frac}       & 3.1 & 0.066
   & \texttt{therefore}  & 80.6 & 0.358
   & \texttt{frac}       & 3.1 & 0.062 \\

5  & \texttt{earlier}    & 79.5 & 0.214
   & \texttt{\_}         & 3.7 & 0.088
   & \texttt{earlier}    & 79.5 & 0.226
   & \texttt{\_}         & 3.6 & 0.095 \\

6  & \texttt{if}         & 74.7 & 0.135
   & $\surd$             & 3.7 & 0.120
   & \texttt{if}         & 75.9 & 0.176
   & $\surd$             & 3.7 & 0.119 \\

7  & \texttt{try}        & 74.6 & 0.121
   & $\theta$            & 3.8 & 0.088
   & \texttt{another}    & 75.6 & 0.431
   & \texttt{9}          & 3.9 & 0.146 \\

8  & \texttt{seems}      & 74.6 & 0.458
   & \texttt{9}          & 3.9 & 0.155
   & \texttt{try}        & 75.1 & 0.126
   & $\theta$            & 3.9 & 0.086 \\

9  & \texttt{another}    & 73.9 & 0.211
   & \texttt{8}          & 4.2 & 0.165
   & \texttt{says}       & 74.4 & 0.111
   & \texttt{8}          & 4.1 & 0.152 \\

10 & \texttt{because}    & 73.2 & 0.127
   & $^\circ$            & 4.2 & 0.093
   & \texttt{seems}      & 74.0 & 0.487
   & $^\circ$            & 4.2 & 0.087 \\

11 & \texttt{now}        & 73.1 & 0.154
   & \texttt{6}          & 4.3 & 0.157
   & \texttt{how}        & 73.5 & 0.170
   & \texttt{6}          & 4.3 & 0.145 \\

12 & \texttt{how}        & 73.1 & 0.158
   & \texttt{\_i}        & 4.4 & 0.082
   & \texttt{think}      & 73.4 & 0.189
   & \texttt{\_i}        & 4.3 & 0.085 \\

13 & \texttt{says}       & 73.0 & 0.106
   & \texttt{\char123}   & 4.4 & 0.086
   & \texttt{because}    & 73.4 & 0.136
   & \texttt{\char123}   & 4.5 & 0.085 \\

14 & \texttt{think}      & 72.6 & 0.126
   & \texttt{5}          & 4.5 & 0.152
   & \texttt{since}      & 73.2 & 0.136
   & \texttt{5}          & 4.5 & 0.139 \\

15 & \texttt{since}      & 72.1 & 0.125
   & \texttt{7}          & 4.5 & 0.161
   & \texttt{check}      & 72.4 & 0.154
   & \texttt{7}          & 4.5 & 0.150 \\

16 & \texttt{but}        & 72.0 & 0.188
   & \texttt{2}          & 4.8 & 0.129
   & \texttt{there}      & 72.2 & 0.200
   & \texttt{2}          & 4.8 & 0.120 \\

17 & \texttt{so}         & 71.4 & 0.142
   & \texttt{)\char94}   & 5.0 & 0.087
   & \texttt{now}        & 72.1 & 0.151
   & \texttt{4}          & 5.0 & 0.136 \\

18 & \texttt{check}      & 71.1 & 0.122
   & \texttt{4}          & 5.1 & 0.141
   & \texttt{but}        & 71.8 & 0.136
   & \texttt{)\char94}   & 5.1 & 0.087 \\

19 & \texttt{let}        & 70.7 & 0.426
   & \texttt{cdot}       & 5.2 & 0.049
   & \texttt{so}         & 71.4 & 0.183
   & \texttt{cdot}       & 5.3 & 0.049 \\

20 & \texttt{there}      & 70.3 & 0.141
   & \texttt{1}          & 5.3 & 0.133
   & \texttt{altern.}    & 71.2 & 0.346
   & \texttt{1}          & 5.3 & 0.127 \\

21 & \texttt{altern.}    & 70.2 & 0.181
   & \texttt{3}          & 5.4 & 0.137
   & \texttt{let}        & 71.2 & 1.088
   & \texttt{3}          & 5.4 & 0.132 \\

22 & \texttt{previous}   & 70.1 & 0.125
   & \texttt{+}          & 5.7 & 0.105
   & \texttt{previous}   & 70.0 & 0.113
   & \texttt{+}          & 5.8 & 0.104 \\

23 & \texttt{when}       & 69.2 & 0.123
   & \texttt{'t}         & 5.9 & 0.120
   & \texttt{when}       & 69.9 & 0.150
   & \texttt{'t}         & 5.8 & 0.120 \\

24 & \texttt{wait}       & 68.1 & 0.221
   & \texttt{\char125}   & 7.0 & 0.090
   & \texttt{wait}       & 68.7 & 0.175
   & \texttt{\char125}   & 7.3 & 0.112 \\

\midrule
\multicolumn{1}{c}{\textbf{Mean}}
& -- & 73.9 & 0.182
& -- & 4.3  & 0.111
& -- & 74.4 & 0.244
& -- & 4.3  & 0.108 \\
\bottomrule
\end{tabular}
\end{table}


\subsection{Token-Level TSD Patterns Across Contextual Interventions}
\label{app:tsd_token_patterns}

To examine whether the token-level structure of TSD depends on a particular contextual intervention, we repeat the token-form analysis under all eight contextual interventions using Qwen3-8B at $\tau=0.01$. Following the main analysis, we merge tokenizer variants corresponding to the same normalized surface form and retain forms occurring more than 20,000 times. Tables~\ref{tab:app_tsd_tokens_inst}--\ref{tab:app_tsd_tokens_sol} report the top-24 and bottom-24 forms ranked by $\rho_{\tau}(v)$ for each positive--negative intervention pair, together with their significant-TSD rates and mean deviation magnitudes.

\paragraph{Stability Across Intervention Polarity.}
Positive and negative variants within each intervention group produce substantially overlapping token rankings. Their top-24 sets share $18/24$ forms for instruction, $22/24$ for evaluation, $24/24$ for answer-level privilege, and $23/24$ for solution-level privilege. The corresponding bottom-24 overlaps are $24/24$, $23/24$, $24/24$, and $24/24$. Thus, reversing intervention polarity or correctness generally leaves the token forms most prone to TSD largely unchanged. This is especially true for forms that remain comparatively stable across conditions, while the instruction-level top set shows somewhat more variation than the other pairs.

\paragraph{Stability Across Intervention Types.}
The same structure persists across different types of contextual intervention. Any two of the eight intervention conditions share at least $18/24$ top-ranked forms and $22/24$ bottom-ranked forms, while $16/24$ top forms and $22/24$ bottom forms appear in every ranking. Across all conditions, high-$\rho_{\tau}$ forms are consistently dominated by natural-language surface expressions such as \texttt{maybe}, \texttt{however}, \texttt{therefore}, \texttt{consider}, and \texttt{think}, whereas low-$\rho_{\tau}$ forms are dominated by digits, mathematical symbols, and notation. This separation is also quantitatively strong: the mean top-24 $\rho_{\tau}$ ranges from $61.1\%$ to $91.5\%$, compared with only $3.5\%$ to $8.2\%$ for the bottom-24, corresponding to an approximately $11$--$18\times$ gap within individual conditions.

\paragraph{Context Modulates Strength More Than Token Identity.}
While the ranking structure remains stable, the magnitude of TSD varies substantially across contextual interventions. Under positive interventions, the mean $\overline{|\Delta|}$ of the top-24 forms increases from $0.080$ under instruction and $0.084$ under evaluation to $0.182$ under answer-level privilege and $0.436$ under solution-level privilege. The bottom-24 exhibits the same overall trend, rising from $0.083$ and $0.091$ to $0.111$ and $0.290$, respectively. Negative interventions show a similar overall increase: the top-24 means are $0.121$, $0.108$, $0.244$, and $0.265$ for instruction, evaluation, answer-level privilege, and solution-level privilege, respectively, while the corresponding bottom-24 means are $0.102$, $0.102$, $0.108$, and $0.199$. Thus, richer privileged context can substantially amplify teacher-side likelihood shifts, particularly at the solution level. Crucially, however, this amplification does not substantially alter which token forms occupy the two ends of the ranking: largely the same surface-form tokens remain TSD-prone, while numerical and symbolic forms remain comparatively stable. These results indicate that contextual interventions primarily modulate the strength of teacher self-deviation, whereas its token-level concentration remains stable across intervention semantics and privilege types.

\begin{table}[!t]
\caption{
Top-24 and bottom-24 token forms under
$c_{\mathrm{sol}}^{\mathrm{pos}}$ and
$c_{\mathrm{sol}}^{\mathrm{neg}}$ at $\tau=0.01$.
The final row reports unweighted means of
$\rho_{\tau}$ and $\overline{|\Delta|}$ over each 24-token subset.
}
\label{tab:app_tsd_tokens_sol}

\centering
\scriptsize
\setlength{\tabcolsep}{0pt}
\renewcommand{\arraystretch}{1.04}

\begin{tabular}{
@{}
r
@{\hspace{5pt}}
l @{\hspace{5pt}} r @{\hspace{9pt}} r
@{\hspace{10pt}}
l @{\hspace{5pt}} r @{\hspace{9pt}} r
@{\hspace{15pt}}
l @{\hspace{5pt}} r @{\hspace{9pt}} r
@{\hspace{10pt}}
l @{\hspace{5pt}} r @{\hspace{9pt}} r
@{}
}
\toprule
& \multicolumn{6}{c}{$c_{\mathrm{sol}}^{\mathrm{pos}}$}
& \multicolumn{6}{c}{$c_{\mathrm{sol}}^{\mathrm{neg}}$} \\
\cmidrule(lr){2-7}\cmidrule(lr){8-13}

\#
& \multicolumn{3}{c}{\textbf{Highest $\rho_{\tau}$}}
& \multicolumn{3}{c}{\textbf{Lowest $\rho_{\tau}$}}
& \multicolumn{3}{c}{\textbf{Highest $\rho_{\tau}$}}
& \multicolumn{3}{c}{\textbf{Lowest $\rho_{\tau}$}} \\
\cmidrule(lr){2-4}\cmidrule(lr){5-7}
\cmidrule(lr){8-10}\cmidrule(lr){11-13}

& \textbf{Token}
& {$\boldsymbol{\rho_{\tau}}$ (\%)}
& {$\mathbf{\overline{|\Delta|}}$}
& \textbf{Token}
& {$\boldsymbol{\rho_{\tau}}$ (\%)}
& {$\mathbf{\overline{|\Delta|}}$}
& \textbf{Token}
& {$\boldsymbol{\rho_{\tau}}$ (\%)}
& {$\mathbf{\overline{|\Delta|}}$}
& \textbf{Token}
& {$\boldsymbol{\rho_{\tau}}$ (\%)}
& {$\mathbf{\overline{|\Delta|}}$} \\
\midrule

1  & \texttt{maybe}      & 97.4 & 0.594
   & \texttt{\char125\char123} & 2.8  & 0.251
   & \texttt{maybe}      & 96.2 & 0.305
   & \texttt{\char125\char123} & 2.0  & 0.141 \\

2  & \texttt{however}    & 97.4 & 0.625
   & \texttt{\#\#\#}     & 3.2  & 0.131
   & \texttt{however}    & 95.3 & 0.358
   & \texttt{\#\#\#}     & 2.9  & 0.102 \\

3  & \texttt{therefore}  & 95.9 & 0.460
   & \texttt{0}          & 5.0  & 0.356
   & \texttt{therefore}  & 95.0 & 0.351
   & \texttt{0}          & 4.3  & 0.250 \\

4  & \texttt{consider}   & 95.4 & 0.380
   & \texttt{9}          & 6.8  & 0.384
   & \texttt{consider}   & 93.5 & 0.230
   & \texttt{9}          & 5.9  & 0.270 \\

5  & \texttt{altern.}    & 94.2 & 0.576
   & \texttt{8}          & 7.2  & 0.395
   & \texttt{earlier}    & 92.5 & 0.304
   & \texttt{frac}       & 6.0  & 0.110 \\

6  & \texttt{earlier}    & 94.2 & 0.413
   & $\surd$             & 7.2  & 0.278
   & \texttt{altern.}    & 90.3 & 0.309
   & \texttt{\_}         & 6.0  & 0.172 \\

7  & \texttt{seems}      & 93.6 & 0.408
   & $^\circ$            & 7.3  & 0.170
   & \texttt{try}        & 90.2 & 0.281
   & $\surd$             & 6.2  & 0.205 \\

8  & \texttt{another}    & 92.9 & 0.504
   & \texttt{\_}         & 7.3  & 0.287
   & \texttt{another}    & 90.1 & 0.287
   & \texttt{8}          & 6.2  & 0.271 \\

9  & \texttt{since}      & 91.8 & 0.441
   & \texttt{6}          & 7.4  & 0.398
   & \texttt{seems}      & 90.0 & 0.229
   & $^\circ$            & 6.2  & 0.117 \\

10 & \texttt{try}        & 91.8 & 0.404
   & \texttt{7}          & 7.7  & 0.389
   & \texttt{think}      & 88.6 & 0.244
   & \texttt{6}          & 6.4  & 0.278 \\

11 & \texttt{think}      & 91.6 & 0.402
   & \texttt{5}          & 7.8  & 0.379
   & \texttt{says}       & 88.6 & 0.213
   & \texttt{7}          & 6.8  & 0.266 \\

12 & \texttt{says}       & 90.4 & 0.246
   & $\theta$            & 7.9  & 0.280
   & \texttt{since}      & 88.4 & 0.247
   & \texttt{\char123}   & 6.8  & 0.153 \\

13 & \texttt{wait}       & 90.0 & 0.415
   & \texttt{frac}       & 8.0  & 0.225
   & \texttt{here}       & 88.0 & 0.266
   & \texttt{5}          & 6.8  & 0.257 \\

14 & \texttt{check}      & 89.6 & 0.278
   & \texttt{\char123}   & 8.0  & 0.234
   & \texttt{if}         & 86.9 & 0.242
   & $\theta$            & 6.9  & 0.215 \\

15 & \texttt{let}        & 89.5 & 0.504
   & \texttt{4}          & 8.8  & 0.372
   & \texttt{because}    & 86.7 & 0.246
   & \texttt{\_i}        & 7.6  & 0.181 \\

16 & \texttt{here}       & 89.4 & 0.364
   & \texttt{2}          & 8.8  & 0.323
   & \texttt{wait}       & 86.7 & 0.226
   & \texttt{4}          & 7.6  & 0.256 \\

17 & \texttt{because}    & 89.4 & 0.396
   & \texttt{3}          & 9.2  & 0.362
   & \texttt{previous}   & 86.3 & 0.219
   & \texttt{2}          & 7.7  & 0.219 \\

18 & \texttt{now}        & 89.2 & 0.485
   & \texttt{\_i}        & 9.2  & 0.283
   & \texttt{how}        & 86.2 & 0.264
   & \texttt{3}          & 8.1  & 0.259 \\

19 & \texttt{also}       & 89.2 & 0.470
   & \texttt{1}          & 9.4  & 0.330
   & \texttt{so}         & 86.2 & 0.263
   & \texttt{1}          & 8.2  & 0.218 \\

20 & \texttt{if}         & 89.2 & 0.364
   & \texttt{'t}         & 10.0 & 0.302
   & \texttt{check}      & 86.0 & 0.210
   & \texttt{'t}         & 8.8  & 0.238 \\

21 & \texttt{similarly}  & 89.0 & 0.532
   & \texttt{)\char94}   & 10.5 & 0.208
   & \texttt{also}       & 85.8 & 0.271
   & \texttt{)\char94}   & 9.0  & 0.166 \\

22 & \texttt{but}        & 88.7 & 0.439
   & \texttt{+}          & 10.7 & 0.276
   & \texttt{but}        & 85.7 & 0.221
   & \texttt{cdot}       & 9.4  & 0.094 \\

23 & \texttt{so}         & 88.3 & 0.366
   & \texttt{cdot}       & 12.8 & 0.159
   & \texttt{now}        & 85.5 & 0.310
   & \texttt{+}          & 9.5  & 0.203 \\

24 & \texttt{how}        & 88.3 & 0.408
   & \texttt{\char125}   & 12.9 & 0.188
   & \texttt{let}        & 84.8 & 0.267
   & \texttt{\char125}   & 11.1 & 0.130 \\

\midrule
\multicolumn{1}{c}{\textbf{Mean}}
& -- & 91.5 & 0.436
& -- & 8.2  & 0.290
& -- & 88.9 & 0.265
& -- & 6.9  & 0.199 \\
\bottomrule
\end{tabular}
\end{table}

\subsection{Trajectory-Level Case Study of TSD}
\label{app:tsd_trajectory_case}

We complement the aggregate analysis with two trajectory-level examples illustrating how TSD is distributed within complete reasoning traces. We use the same Qwen3-1.7B student and Qwen3-8B teacher configuration and visualize TSD under $c_{\mathrm{sol}}^{\mathrm{pos}}$. Red and blue indicate decreases and increases in teacher log-likelihood, respectively, with darker shades denoting larger $|\Delta_t^T|$; tokens with $|\Delta_t^T|\leq0.1$ are left uncolored. Since the student trajectory is fixed, all highlighted variation is induced solely by the additional reference solution.


\newcommand{\tsdNegI}[1]{\begingroup\setlength{\fboxsep}{0.45pt}\colorbox{red!20}{\strut #1}\endgroup}
\newcommand{\tsdNegII}[1]{\begingroup\setlength{\fboxsep}{0.45pt}\colorbox{red!30}{\strut #1}\endgroup}
\newcommand{\tsdNegIII}[1]{\begingroup\setlength{\fboxsep}{0.45pt}\colorbox{red!42}{\strut #1}\endgroup}
\newcommand{\tsdNegIV}[1]{\begingroup\setlength{\fboxsep}{0.45pt}\colorbox{red!54}{\strut #1}\endgroup}
\newcommand{\tsdNegV}[1]{\begingroup\setlength{\fboxsep}{0.45pt}\colorbox{red!66}{\strut #1}\endgroup}

\newcommand{\tsdPosI}[1]{\begingroup\setlength{\fboxsep}{0.45pt}\colorbox{blue!20}{\strut #1}\endgroup}
\newcommand{\tsdPosII}[1]{\begingroup\setlength{\fboxsep}{0.45pt}\colorbox{blue!30}{\strut #1}\endgroup}
\newcommand{\tsdPosIII}[1]{\begingroup\setlength{\fboxsep}{0.45pt}\colorbox{blue!42}{\strut #1}\endgroup}
\newcommand{\tsdPosIV}[1]{\begingroup\setlength{\fboxsep}{0.45pt}\colorbox{blue!54}{\strut #1}\endgroup}
\newcommand{\tsdPosV}[1]{\begingroup\setlength{\fboxsep}{0.45pt}\colorbox{blue!66}{\strut #1}\endgroup}

\begingroup
\scriptsize
\raggedright
\sloppy
\setlength{\parindent}{0pt}
\noindent\hrule\medskip

\noindent
\textbf{Case 1: Correct Student Rollout}
\hfill
\textit{Outcome: Correct}

\vspace{2pt}
\hrule
\vspace{4pt}

\noindent
\textbf{Problem.}
Calculate the least integer greater than
$5^{(-6)(-5)(-4)\cdots(2)(3)(4)}$.

\vspace{3pt}

\noindent
\textbf{Answer.}
The student rollout outputs $\boxed{2}$, which matches the gold answer.

\vspace{3pt}

\noindent
\textbf{TSD Legend.}\quad
\tsdNegI{\phantom{xx}} $[-0.3,-0.1)$\quad
\tsdNegII{\phantom{xx}} $[-0.7,-0.3)$\quad
\tsdNegIII{\phantom{xx}} $[-1.5,-0.7)$\quad
\tsdNegIV{\phantom{xx}} $[-3,-1.5)$\quad
\tsdNegV{\phantom{xx}} $<-3$

\par\noindent
\tsdPosI{\phantom{xx}} $(0.1,0.3]$\quad
\tsdPosII{\phantom{xx}} $(0.3,0.7]$\quad
\tsdPosIII{\phantom{xx}} $(0.7,1.5]$\quad
\tsdPosIV{\phantom{xx}} $(1.5,3]$\quad
\tsdPosV{\phantom{xx}} $>3$

\vspace{3pt}

\noindent

\vspace{4pt}
\hrule
\vspace{4pt}

\noindent
\textbf{Student trajectory excerpt under $c_{\mathrm{sol}}^{\mathrm{pos}}$.}

\vspace{3pt}

\noindent
{\ttfamily
Okay, \tsdNegII{so} I need to find the least integer greater than \tsdNegV{\textdollar{}}5\textasciicircum{}{\normalfont\char123}(-6)(-5)(-4\tsdNegIII{)...}(2)(3)(4){\normalfont\char125}\textdollar{}. \tsdNegIV{Hmm}, let me \tsdPosII{start} by understanding \tsdPosII{the} \tsdPosI{exponent} \tsdNegIV{here}. The exponent is \tsdNegIII{the} product of \tsdPosI{all} integers from -6 up to 4\tsdNegIII{.} \tsdNegII{Wait}, \tsdNegIII{let} me confirm\tsdNegIV{:} the \tsdNegV{exponent} is \tsdNegV{(-}6\tsdNegI{)(}-5)(-4)...(2)(3)(4). \tsdPosI{So} \tsdPosI{that}\tsdNegII{'s} \tsdNegIII{multiplying} all \tsdNegII{integers} \tsdNegIV{from} -6 \tsdPosI{to} 4\tsdPosII{,} \tsdNegV{but} \tsdNegIV{I} need to \tsdNegI{check} if \tsdPosII{that}\tsdNegIV{'s} \tsdPosIII{correct}\tsdNegII{.} \tsdNegII{Let} me \tsdNegII{count} \tsdPosII{how} many \tsdPosIV{numbers} \tsdNegI{are} \tsdPosII{there}\tsdPosI{.}

\par\noindent
Starting from -6 \tsdNegIV{up} to 4\tsdNegI{.} Let\tsdPosI{'s} \tsdNegIII{see}: \tsdPosII{from} -6 to\tsdPosIV{\hspace{0.32em}}\tsdNegI{4} \tsdNegI{inclusive}. \tsdNegII{The} \tsdNegII{number} of \tsdNegIII{terms} \tsdPosI{would} be 4 - (-6) + 1 = 11 terms\tsdPosIII{.} \tsdNegIV{But} \tsdNegIII{wait}, \tsdPosII{let} me \tsdNegIV{check}:

\par\noindent
\tsdPosI{Numbers}\tsdNegIII{:} -6, -5, -4, -3, -2, -1, 0, 1, 2, 3, 4. \tsdNegII{Yes}, that's 11 numbers. \tsdNegV{But} \tsdNegIII{wait}, \tsdNegI{the} \tsdNegII{problem} \tsdNegI{says} \tsdPosIII{"}(-6)(-5)(-4)...(2)(3)(4\tsdNegI{)".} \tsdNegIII{So} \tsdPosII{the} \tsdNegI{product} \tsdPosII{is} from -6 \tsdPosI{to} 4\tsdPosII{,} \tsdPosIII{including}\tsdNegIII{\hspace{0.32em}}0\tsdNegII{?} \tsdNegIV{Wait}, \tsdNegIV{but}\tsdNegIV{\hspace{0.32em}}0 is \tsdPosIV{included} \tsdPosII{here}\tsdNegI{.} \tsdNegV{However}, \tsdPosV{when} you multiply \tsdNegV{by}\tsdNegII{\hspace{0.32em}}0, the entire product becomes 0. \tsdNegV{But} \tsdNegIV{then} \tsdPosII{the} exponent \tsdNegIII{would} be\tsdNegI{\hspace{0.32em}}0, so\tsdPosI{\hspace{0.32em}}5\textasciicircum{}0 is 1\tsdPosI{.} \tsdPosI{Then} the least integer greater than 1 is 2\tsdPosIII{.} \tsdNegIII{But} \tsdNegIV{that} seems \tsdNegIII{too} \tsdPosII{straightforward}. \tsdNegIV{Maybe} I \tsdNegIII{made} a mistake here\tsdNegIII{.}

\par\noindent
Wait, \tsdPosIII{but} \tsdNegIV{hold} on. Let me \tsdNegII{check} \tsdPosI{again}. The \tsdPosI{problem} \tsdNegI{says} \tsdNegI{"}\tsdNegI{(-}6)(-5)(-4)...(2)(3)(4)". \tsdNegII{Is} \tsdNegIII{the} product \tsdPosI{from} -6 \tsdNegII{up} to 4\tsdNegII{,} \tsdNegI{including}\tsdNegIII{\hspace{0.32em}}0? \tsdNegIV{But} \tsdNegIII{if} \tsdNegIII{that}'s the case, \tsdPosI{then} \tsdNegIII{the} \tsdPosI{product} is zero\tsdPosI{,} so \tsdPosI{the} \tsdNegI{exponent} is zero, \tsdNegII{and} 5\textasciicircum{}0 \tsdPosI{is} 1\tsdPosI{.} Therefore, the least integer greater than 1 is 2. \tsdNegI{But} \tsdNegIV{that} \tsdNegI{seems} \tsdNegI{too} \tsdNegI{easy}\tsdPosI{.} \tsdNegII{Maybe} \tsdPosI{there}'s a misunderstanding here\tsdNegI{.}

\par\noindent
\tsdNegII{Alternatively}, maybe the \tsdPosIII{product} \tsdNegI{is} \tsdNegII{from} -6 \tsdPosI{to} 4, \tsdNegIII{excluding}\tsdNegII{\hspace{0.32em}}0? But the \tsdNegII{problem} \tsdNegII{didn}'t specify\tsdNegIV{.} Wait, \tsdPosIII{the} \tsdPosI{original} \tsdPosI{problem} \tsdPosII{says} \tsdNegI{"}(-6)(-5)(-4)...(2)(3)(4)". \tsdPosII{The} ellipsis \tsdNegI{is} between -4 and 2\tsdPosII{,} \tsdNegI{so} maybe \tsdPosI{the} \tsdNegI{product} is \tsdNegI{from} -6 \tsdNegIII{up} to 4, but \tsdNegV{skipping}\tsdNegIII{\hspace{0.32em}}0? \tsdPosI{Wait}, \tsdNegI{but} \tsdNegII{how}\tsdPosI{?} \tsdNegI{If} you have \tsdNegII{-}6, -5, -4, \tsdNegII{...,}\tsdNegII{\hspace{0.32em}}2, 3, 4. \tsdNegII{So} \tsdNegII{if} you start at -6 \tsdNegI{and} go up to 4, \tsdNegII{but} \tsdNegIII{not} \tsdNegIII{including}\tsdNegI{\hspace{0.32em}}0? But\tsdNegIII{\hspace{0.32em}}0 is between -\tsdPosI{6} and 4. \tsdNegI{So} unless \tsdPosI{there}'s a typo, \tsdPosIII{but} \tsdPosIII{the} \tsdNegIV{problem} \tsdNegIII{is} \tsdNegI{written} as \tsdNegI{(-}6)(-5)(-4)...(2)(3)(4). So \tsdNegII{maybe} the \tsdNegI{product} is from -6 to 4, \tsdPosII{including}\tsdNegII{\hspace{0.32em}}0\tsdNegII{?} \tsdNegIII{But} \tsdNegI{then} the product \tsdNegI{is} \tsdPosI{zero}. But \tsdNegI{maybe} \tsdPosII{there}'s a \tsdPosI{mistake} \tsdPosI{here}.

\par\noindent
\tsdPosII{Wait}, \tsdPosIV{but} let \tsdNegI{me} \tsdNegI{check} \tsdNegIII{if} \tsdPosI{the} \tsdPosIII{product} \tsdNegII{is} \tsdPosI{from} -6 to 4\tsdNegII{,} \tsdNegII{but} \tsdNegI{not} including\tsdNegI{\hspace{0.32em}}0\tsdPosII{.} If \tsdPosI{that}'s the case, then \tsdPosI{the} \tsdPosI{product} \tsdNegII{is} \tsdNegIII{(-}6)(-5)(-4\tsdNegIII{)...}\tsdNegI{(-}1)(1)(2)(3)(4). \tsdPosII{Let}\tsdPosIII{'s} \tsdNegII{check} \tsdNegIII{how} \tsdNegII{many} terms that would be. From -6 to -1 \tsdPosI{is} 6 \tsdPosII{numbers}, then\tsdPosI{\hspace{0.32em}}1\tsdNegI{,}\tsdNegI{\hspace{0.32em}}2, 3, 4\tsdPosI{.} So total of 6 + 4 = 10 terms. \tsdPosI{But} \tsdPosI{then} the \tsdPosII{product} would \tsdNegI{be} \tsdNegII{(-}6)(-5\tsdPosII{)...}(-1\tsdPosII{)} \tsdNegII{*}\tsdNegII{\hspace{0.32em}}1 \tsdNegI{*} 2 * 3 * 4. \tsdNegI{Let} me compute that\tsdNegI{.}

\par\noindent
\tsdPosI{First}, \tsdPosI{the} product \tsdPosI{from} -6 to -1 \tsdNegII{is} (-\tsdNegIV{1})\textasciicircum{}6 \tsdNegI{*} 6! = 1 * 720 = 720. Then \tsdPosII{multiply} by 1*2*3*4 = 24. So total product is 720 * 24 \tsdPosI{=} 17\tsdNegI{,}280. \tsdPosII{Then}\tsdPosII{\hspace{0.32em}}5\tsdPosI{\textasciicircum{}}17280 \tsdPosI{is} \tsdNegII{an} \tsdPosIII{extremely} large number\tsdNegI{.} \tsdPosII{Then} the least integer greater than that \tsdPosI{would} \tsdPosI{be} 5\textasciicircum{}17280 + 1\tsdPosII{,} \tsdNegI{but} that \tsdNegII{seems} \tsdPosI{way} \tsdNegI{too} big. \tsdPosI{But} the \tsdNegII{problem} \tsdNegI{says} \tsdNegI{"}the least integer greater than\tsdNegI{\hspace{0.32em}}5\tsdNegII{\textasciicircum{}{\normalfont\char123}}...\tsdPosIII{{\normalfont\char125}"} \tsdNegII{so} \tsdNegI{maybe} \tsdPosI{the} \tsdPosIII{answer} \tsdNegI{is} \tsdNegI{something} else\tsdPosI{.} \tsdPosI{But} \tsdPosIV{the} \tsdPosIII{original} \tsdNegII{problem} \tsdNegIV{is} written \tsdNegI{as} (-6)(-5\tsdPosIV{)...}(2)(3)(4). So \tsdNegI{if} \tsdPosII{the} product includes 0, \tsdPosI{then} \tsdPosII{it}'s \tsdPosI{zero}, \tsdPosII{so} 5\textasciicircum{}0 \tsdPosII{is} 1\tsdNegII{.} \tsdPosIII{Then} \tsdNegI{the} \tsdNegIII{least} integer greater than 1 is 2. But \tsdNegI{maybe} the \tsdNegI{problem} is written \tsdPosII{with} a \tsdNegIII{different} \tsdPosI{intention}\tsdNegII{.}

\par\noindent
\tsdPosII{Wait}, \tsdNegII{maybe} the \tsdNegI{exponent} is \tsdPosII{from} -6 to 4, but not including\tsdNegI{\hspace{0.32em}}0? But how\tsdNegI{?} \tsdNegI{If} \tsdNegII{you} have \tsdNegII{from} -6 to 4, \tsdNegII{but} \tsdPosI{not} including\tsdNegI{\hspace{0.32em}}0, then \tsdPosI{the} product \tsdNegII{is} \tsdNegI{(-}6)(-5\tsdPosI{)...}(-1)(1)(2\tsdPosI{)(}3)(4). \tsdNegIII{Which} is \tsdNegII{what} I \tsdPosI{thought} \tsdPosI{earlier}. But then the \tsdNegIII{exponent} \tsdNegI{is} 17\tsdNegI{2}80\tsdPosI{,} so 5\textasciicircum{}17280 is a \tsdNegIII{gigantic} number, \tsdPosII{and} the least integer greater than that \tsdPosII{would} be 5\textasciicircum{}17280 + 1. But that \tsdNegII{seems} \tsdNegIII{like} an \tsdNegII{unreasonable} answer\tsdPosI{.} \tsdPosII{However}, \tsdNegIII{maybe} \tsdNegII{there}'s a \tsdPosII{mistake} in \tsdNegII{my} \tsdPosI{calculation}.

\par\noindent
Wait, \tsdNegI{let} me check the product again. \tsdPosI{If} \tsdPosII{the} \tsdPosI{product} is from -6 to 4\tsdNegI{,} \tsdNegIII{excluding} 0\tsdNegI{,} then\tsdNegI{:}

\par\noindent
First, \tsdPosI{the} \tsdPosII{product} \tsdPosI{from} -6 to -1 \tsdPosI{is} (-6)(-5)(-4)(-3)(-2)(-1). Let \tsdNegI{me} compute that\tsdPosI{:}

\par\noindent
(-6\tsdPosIII{)} * (-5) = 30

\par\noindent
30 * (-4) = -120

\par\noindent
-120 * (-3) = 360

\par\noindent
360 * (-2) = -720

\par\noindent
-720 * (-1) = 720. \tsdPosI{So} that's 720.

\par\noindent
}

\medskip
\noindent
\textit{[Remaining trajectory omitted.]}

\par\medskip\noindent\hrule
\endgroup


\providecommand{\tsdNegI}[1]{\begingroup\setlength{\fboxsep}{0.45pt}\colorbox{red!20}{\strut #1}\endgroup}
\providecommand{\tsdNegII}[1]{\begingroup\setlength{\fboxsep}{0.45pt}\colorbox{red!30}{\strut #1}\endgroup}
\providecommand{\tsdNegIII}[1]{\begingroup\setlength{\fboxsep}{0.45pt}\colorbox{red!42}{\strut #1}\endgroup}
\providecommand{\tsdNegIV}[1]{\begingroup\setlength{\fboxsep}{0.45pt}\colorbox{red!54}{\strut #1}\endgroup}
\providecommand{\tsdNegV}[1]{\begingroup\setlength{\fboxsep}{0.45pt}\colorbox{red!66}{\strut #1}\endgroup}

\providecommand{\tsdPosI}[1]{\begingroup\setlength{\fboxsep}{0.45pt}\colorbox{blue!20}{\strut #1}\endgroup}
\providecommand{\tsdPosII}[1]{\begingroup\setlength{\fboxsep}{0.45pt}\colorbox{blue!30}{\strut #1}\endgroup}
\providecommand{\tsdPosIII}[1]{\begingroup\setlength{\fboxsep}{0.45pt}\colorbox{blue!42}{\strut #1}\endgroup}
\providecommand{\tsdPosIV}[1]{\begingroup\setlength{\fboxsep}{0.45pt}\colorbox{blue!54}{\strut #1}\endgroup}
\providecommand{\tsdPosV}[1]{\begingroup\setlength{\fboxsep}{0.45pt}\colorbox{blue!66}{\strut #1}\endgroup}

\begingroup
\scriptsize
\raggedright
\sloppy
\setlength{\parindent}{0pt}
\noindent\hrule\medskip

\noindent
\textbf{Case 2: Incorrect Student Rollout}
\hfill
\textit{Outcome: Incorrect}

\vspace{2pt}
\hrule
\vspace{4pt}

\noindent
\textbf{Problem.}
Alex the Kat has written $61$ problems for a math contest, and there are a
total of $187$ problems submitted. How many more problems does he need to
write and submit before he has written half of the total problems?

\vspace{3pt}

\noindent
\textbf{Answer and error.}
The student outputs $93.5$, whereas the gold answer is $65$.
It incorrectly keeps the total fixed at $187$ and treats
$187/2=93.5$ as the required additional count.

\vspace{3pt}

\noindent
\textbf{TSD Legend.}\quad
\tsdNegI{\phantom{xx}} $[-0.3,-0.1)$\quad
\tsdNegII{\phantom{xx}} $[-0.7,-0.3)$\quad
\tsdNegIII{\phantom{xx}} $[-1.5,-0.7)$\quad
\tsdNegIV{\phantom{xx}} $[-3,-1.5)$\quad
\tsdNegV{\phantom{xx}} $<-3$

\par\noindent
\tsdPosI{\phantom{xx}} $(0.1,0.3]$\quad
\tsdPosII{\phantom{xx}} $(0.3,0.7]$\quad
\tsdPosIII{\phantom{xx}} $(0.7,1.5]$\quad
\tsdPosIV{\phantom{xx}} $(1.5,3]$\quad
\tsdPosV{\phantom{xx}} $>3$

\vspace{3pt}

\noindent

\vspace{4pt}
\hrule
\vspace{4pt}

\noindent
\textbf{Student trajectory excerpt under $c_{\mathrm{sol}}^{\mathrm{pos}}$.}

\vspace{3pt}

\noindent
{\ttfamily
Okay, \tsdPosII{let}\tsdPosII{'s} \tsdPosIII{try} to figure out \tsdNegIII{this} problem \tsdPosIII{step} by step. So, Alex the Kat has already written 61 problems \tsdNegIV{for} a math contest\tsdPosV{.} \tsdNegV{The} total number of problems \tsdNegI{submitted} \tsdNegV{is} 187. \tsdNegI{The} question is asking how many more problems \tsdPosIV{he} needs to write and submit \tsdNegV{before} he has written half of the total problems.\tsdPosII{\hspace{0.32em}}

\par\noindent
\tsdNegIV{First}, \tsdPosII{I} need to \tsdNegIII{understand} \tsdNegI{what} exactly \tsdNegII{is} being asked here. The \tsdNegIV{total} \tsdPosI{number} of problems \tsdNegV{submitted} is\tsdNegI{\hspace{0.32em}}187\tsdPosIII{,} but \tsdPosII{I} \tsdPosII{think} \tsdPosI{that} \tsdNegV{might} \tsdPosIV{not} \tsdPosI{be} \tsdNegV{directly} \tsdNegIV{relevant}\tsdNegIII{.} Wait, \tsdPosIII{actually}, \tsdPosIV{the} problem says \tsdNegV{Alex} has written 61 \tsdNegII{problems}, and there are \tsdNegII{a} total of 187 problems submitted. \tsdNegIII{Hmm}\tsdPosI{,} \tsdPosI{maybe} \tsdNegI{I} need to clarify \tsdNegI{that}.\tsdPosII{\hspace{0.32em}}

\par\noindent
\tsdPosI{Wait}, \tsdNegIII{maybe} the \tsdNegI{total} number of problems in the contest \tsdNegI{is}\tsdNegII{\hspace{0.32em}}187? Or is 187 the \tsdPosV{number} of problems \tsdNegIV{submitted} by \tsdPosII{others}? \tsdPosI{The} \tsdNegI{wording} \tsdNegI{is} a bit \tsdPosI{confusing}. Let me \tsdNegI{read} \tsdNegIII{it} again\tsdNegIII{:} "Alex the Kat has written 61 problems for a math contest, and there are a total of 187 problems submitted\tsdNegIII{.} How many more problems does he need to write and submit before he has written half of the total problems\tsdPosI{?"}

\par\noindent
\tsdNegI{So}, \tsdPosII{the} total \tsdPosII{number} of problems submitted is 187. \tsdPosI{But} \tsdNegIII{Alex} has written 61\tsdPosIII{.} \tsdPosII{Wait}, \tsdPosII{maybe} \tsdPosIII{"}\tsdPosII{submitted}" refers to the total \tsdNegII{number} of problems \tsdPosI{that} have been submitted \tsdNegI{by} \tsdNegI{everyone}\tsdPosI{,} including Alex? \tsdPosI{Or} is\tsdPosII{\hspace{0.32em}}187 the total number of problems \tsdPosIII{in} the contest\tsdNegI{?}

\par\noindent
\tsdNegIV{This} is \tsdNegI{a} bit ambiguous\tsdNegI{.} \tsdPosIV{But} \tsdNegIV{let}\tsdPosII{'s} \tsdNegIV{think}\tsdNegIII{.} \tsdNegIII{If} \tsdNegI{Alex} \tsdNegI{has} written 61 \tsdNegII{problems}, and \tsdNegI{there} are\tsdPosI{\hspace{0.32em}}187 problems submitted in total, \tsdNegI{maybe} \tsdNegIII{the} \tsdNegII{total} \tsdNegI{number} of problems \tsdNegI{in} the contest is 187\tsdPosI{?} \tsdPosII{But} then he \tsdNegI{has} written 61\tsdPosI{,} so he needs to write \tsdPosI{half} of\tsdPosII{\hspace{0.32em}}187\tsdPosIV{,} which is 93.5\tsdNegIV{?} \tsdNegI{But} \tsdPosII{that} \tsdPosII{doesn}'t make sense \tsdNegI{because} you can't \tsdPosII{write} half a problem. \tsdNegIII{Wait}, \tsdNegIII{maybe} \tsdNegII{the} \tsdNegI{total} number of problems \tsdPosI{is} \tsdNegV{different}\tsdPosIII{.}\tsdPosI{\hspace{0.32em}}

\par\noindent
\tsdNegIII{Alternatively}, \tsdNegI{maybe} the \tsdNegI{total} number of problems \tsdPosIII{in} the contest is\tsdNegIV{\hspace{0.32em}}187, \tsdNegI{and} Alex has written 61\tsdPosII{.} \tsdPosIII{Then} \tsdPosI{he} needs to write \tsdNegII{half} of \tsdNegI{the} total\tsdPosIII{,} which \tsdPosI{is} \tsdNegII{1}87\tsdPosII{/}2 \tsdPosI{=} 93.5\tsdNegI{.} \tsdPosI{But} \tsdNegI{since} \tsdNegIII{you} can't \tsdNegI{have} half a problem, \tsdNegI{maybe} \tsdNegII{it}'s\tsdPosII{\hspace{0.32em}}94? But \tsdNegII{the} \tsdNegI{problem} \tsdPosI{says} "\tsdNegIV{how} many more problems does he need to write and submit\tsdNegV{,"} so maybe \tsdNegII{it}'s \tsdNegIII{expecting} \tsdNegIII{a} \tsdNegIV{fractional} answer? \tsdPosIII{But} \tsdNegII{that} seems \tsdNegI{odd}.

\par\noindent
\tsdPosI{Wait}, \tsdNegIV{maybe} \tsdPosI{the} \tsdNegII{total} number of problems is not\tsdNegV{\hspace{0.32em}}187. \tsdPosII{Let} me \tsdNegI{re}-examine the problem\tsdNegI{.}

\par\noindent
\tsdNegIV{"}Alex the Kat has written 61 problems for a math contest, and there are a total of 187 problems submitted. How many more problems does he need to write and submit before he has written half of the total problems?"

\par\noindent
\tsdNegIII{Hmm}\tsdNegI{.} \tsdNegIII{Maybe} "\tsdPosIII{submitted}" \tsdNegII{refers} to the total number of problems that have been submitted by all participants, \tsdPosI{including} Alex\tsdNegII{?} So, \tsdPosI{Alex} has written 61 \tsdNegII{problems}, and \tsdNegII{there} are 187 problems \tsdPosI{submitted} in total. \tsdPosI{So}\tsdNegI{,} \tsdNegI{the} total number of problems \tsdNegII{in} the contest is 187\tsdPosI{?} \tsdPosI{But} \tsdNegI{then} Alex has \tsdPosI{written} 61\tsdPosI{,} so he needs to write half of 187\tsdPosI{,} which is 93.5. But \tsdNegIII{since} \tsdNegII{you} can't write half a problem, \tsdNegI{maybe} the \tsdPosIII{answer} is\tsdNegI{\hspace{0.32em}}94? But \tsdPosI{the} \tsdNegIII{problem} is \tsdNegI{probably} \tsdNegII{expecting} \tsdNegI{an} \tsdNegIII{exact} \tsdPosI{answer}\tsdNegII{.}\tsdPosI{\hspace{0.32em}}

\par\noindent
\tsdNegII{Alternatively}, \tsdNegI{maybe} the total number of problems \tsdPosII{in} the contest is \tsdPosII{not}\tsdNegV{\hspace{0.32em}}187. \tsdNegII{Maybe}\tsdPosII{\hspace{0.32em}}187 is the \tsdNegIII{total} number of problems \tsdNegI{that} have been submitted, \tsdNegII{and} Alex \tsdNegII{has} written 61. \tsdNegII{Then}, the \tsdPosI{total} number of problems \tsdPosI{in} the contest \tsdPosI{is}\tsdPosII{\hspace{0.32em}}187\tsdNegII{?} \tsdNegI{But} \tsdNegI{then} \tsdNegI{he} needs to write half of that, \tsdPosI{which} is 93.5. But again, \tsdPosI{this} \tsdPosI{seems} odd.

\par\noindent
\tsdPosI{Wait}, \tsdNegIII{maybe} \tsdPosI{the} problem is \tsdPosI{that} \tsdNegIV{Alex} has written 61 \tsdNegI{problems}, and \tsdNegI{there} are 187 problems \tsdNegIII{in} total. \tsdNegI{So}\tsdNegI{,} \tsdNegI{he} needs to write half of 187\tsdNegI{,} which is 93.5\tsdPosI{.} But \tsdNegII{since} he \tsdNegIV{can}'t write half a problem, maybe \tsdNegIII{we} need to \tsdNegII{round} up\tsdNegI{?} \tsdPosI{But} the \tsdNegII{problem} might \tsdNegI{be} \tsdNegIII{expecting} \tsdNegIII{an} exact answer\tsdNegI{.}\tsdPosI{\hspace{0.32em}}

\par\noindent
\tsdNegII{Alternatively}, \tsdNegI{maybe} \tsdPosI{the} \tsdPosII{problem} is that \tsdPosI{the} total number of problems submitted \tsdPosII{is} 187, \tsdNegI{and} Alex \tsdNegI{has} written 61. So, the total number of problems \tsdNegII{that} \tsdPosIV{have} been submitted \tsdNegII{is} 187\tsdPosI{.} \tsdPosII{But} \tsdPosI{Alex} \tsdNegII{needs} to write \tsdNegIII{half} of the total \tsdNegI{problems}. \tsdNegI{Wait}, \tsdPosI{but} \tsdNegII{if} \tsdNegII{the} total \tsdNegI{problems} submitted is 187, \tsdNegI{then} \tsdNegIV{half} of that is 93.5. \tsdNegI{So}\tsdNegI{,} he \tsdPosIII{needs} to \tsdNegI{write} 93.5 \tsdNegII{more} \tsdNegI{problems}? \tsdNegI{But} \tsdPosI{that}'s not possible.

\par\noindent
Wait, \tsdNegII{maybe} \tsdPosIII{I} need to think differently. \tsdNegIV{Maybe} \tsdPosI{the} total number of problems \tsdNegI{in} the contest is\tsdNegIII{\hspace{0.32em}}187. \tsdNegII{Alex} has written 61\tsdPosI{,} \tsdNegII{so} he needs to write \tsdNegI{half} of\tsdNegI{\hspace{0.32em}}187, which is 93.5. \tsdNegIV{Therefore}, he needs to write 93.5 \tsdPosI{more} \tsdNegI{problems}. But \tsdNegII{since} \tsdNegI{you} can't write half a problem, \tsdNegI{maybe} the \tsdPosI{answer} is 94\tsdNegI{?} \tsdPosI{But} the \tsdNegI{problem} \tsdPosI{says} "how many more problems does he need to write and submit\tsdNegIII{,"} \tsdNegI{so} maybe \tsdNegI{it}'s expecting \tsdNegII{a} \tsdNegII{fractional} answer\tsdNegI{?} \tsdPosII{But} \tsdNegII{in} \tsdNegIII{reality}, \tsdNegI{problems} are \tsdNegI{whole} numbers.

\par\noindent
}

\medskip
\noindent
\textit{[Remaining trajectory omitted.]}

\par\medskip\noindent\hrule
\endgroup

\paragraph{TSD Throughout a Correct Trajectory.}
Case~1 shows a rollout that correctly concludes $\boxed{2}$ after recognizing that the exponent contains zero. Despite reaching the correct reasoning early, the student repeatedly revisits its interpretation. TSD is distributed throughout this process rather than being confined to task-critical mathematical steps. Strong deviations frequently occur on discourse and reasoning-management expressions such as \texttt{Hmm}, \texttt{Wait}, \texttt{But}, \texttt{However}, \texttt{Maybe}, and \texttt{Alternatively}, whereas much of the numerical and symbolic content remains comparatively stable. This qualitatively matches the token-level pattern in Table~\ref{tab:top_bottom_tsd_tokens}.

\paragraph{TSD Is Not Necessarily Localized Around Reasoning Errors.}
Case~2 shows an incorrect rollout in which the student repeatedly treats $187$ as a fixed total and reasons from $187/2=93.5$. The correct relation is
\begin{equation}
    61+x=\frac{187+x}{2},
\end{equation}
where $x$ denotes the number of additional problems Alex needs to write, which yields $x=65$. However, the TSD pattern does not concentrate around this identifiable conceptual error or the repeated occurrence of $93.5$. Instead, substantial deviations again appear broadly on expressions such as \texttt{First}, \texttt{Hmm}, \texttt{Wait}, \texttt{Alternatively}, and \texttt{Therefore}. Thus, even when the privileged solution directly resolves the student's mistake, the induced likelihood variation does not behave as a localized reasoning-error signal.

\paragraph{Takeaway.}
Across both correct and incorrect trajectories, TSD is broadly distributed and particularly pronounced on tokens that organize or redirect reasoning, rather than selectively aligning with mathematical correctness or error locations. These examples provide a trajectory-level illustration of why context-induced teacher variation should not be indiscriminately treated as transferable supervision, further motivating the calibration mechanism in Cal-OPD.

\subsection{Training Details and Hyperparameters}
\label{app:training_details}

We implement all methods in the \texttt{verl} framework~\citep{sheng2025hybridflow} and train on 8 NVIDIA H20 GPUs, with 4 GPUs hosting the student and 4 hosting the teacher. All methods share a common training recipe unless otherwise specified. We first summarize the shared configuration, then describe method-specific hyperparameters and the algorithmic role each plays.

\paragraph{Shared Training Configuration.}
Table~\ref{tab:shared_hyperparams} lists the optimization and training-loop settings common to all baselines and Cal-OPD. All methods are trained for 100 steps, with a maximum response length of 16,384 tokens during training, temperature $1.0$, and top-$p$ $1.0$. We use AdamW with a learning rate of $1\times10^{-6}$, no learning-rate warmup, weight decay $0.01$, a cosine schedule, and gradient clipping at $1.0$. Each training step processes 256 on-policy trajectories, with one rollout per question, one PPO epoch, and a mini-batch size of 256. The entropy coefficient and KL regularization are set to zero because the OPD distillation objective replaces the standard RL KL penalty. We use GRPO as the advantage estimator with standard deviation normalization within each group~\citep{yu2026dapo}.

\begin{table}[t]
\centering
\small
\setlength{\tabcolsep}{6pt}
\renewcommand{\arraystretch}{1.10}
\begin{tabular}{@{}lll@{}}
\toprule
\textbf{Category} & \textbf{Hyperparameter} & \textbf{Value} \\
\midrule
\multicolumn{3}{@{}l}{\textit{Optimizer}} \\
& Learning rate & $1\times10^{-6}$ \\
& LR warmup ratio & 0.0 \\
& Weight decay & 0.01 \\
& LR scheduler & cosine \\
& Gradient clip & 1.0 \\
\midrule
\multicolumn{3}{@{}l}{\textit{Training loop}} \\
& Training steps & 100 \\
& Max prompt length & 2,048 \\
& Max response length & 16,384 \\
& Temperature & 1.0 \\
& Top-$p$ & 1.0 \\
& PPO epochs & 1 \\
& Train batch size & 256 trajectories/step \\
& PPO mini-batch size & 256 \\
& PPO micro-batch size & 1 + dynamic batching \\
& Entropy coefficient & 0.0 \\
& KL regularization & disabled \\
\midrule
\multicolumn{3}{@{}l}{\textit{Distillation}} \\
& Student chunk size & 1024 \\
& Teacher chunk size & 128 \\
& Teacher max logprobs & 16 \\
& Token selection ratio & 1.0 \\
& Policy loss mode & reinforce \\
\midrule
\multicolumn{3}{@{}l}{\textit{Advantage and loss}} \\
& Loss max clamp & 10 \\
& Advantage estimator & GRPO \\
\bottomrule
\end{tabular}
\caption{
Shared training configuration for all methods.
}
\label{tab:shared_hyperparams}
\end{table}

\paragraph{Distillation-Specific Configuration.}
The distillation module operates on student-generated trajectories and queries the teacher for token-level log-probabilities. We set the student chunk size to 1024 and the teacher chunk size to 128, which controls the number of tokens processed per forward pass and is chosen to balance GPU memory against throughput. The teacher returns log-probabilities over its top-16 tokens by default, providing a dense yet bounded supervision signal. All student-generated tokens participate in the distillation loss without filtering, corresponding to a token selection ratio of 1.0 with random selection. The policy loss is computed in \texttt{reinforce} mode, and the per-token loss is clamped at a maximum value of 10 to prevent destabilizing updates from outlier advantages.

\paragraph{Method-Specific Hyperparameters.}
Table~\ref{tab:method_hyperparams} summarizes the hyperparameters that differ across baselines. Below, we describe the algorithmic role of each.

\begin{table}[t]
\centering
\small
\setlength{\tabcolsep}{5pt}
\renewcommand{\arraystretch}{1.15}
\begin{tabularx}{\linewidth}{
@{}
>{\raggedright\arraybackslash}p{0.15\linewidth}
>{\raggedright\arraybackslash}X
@{}
}
\toprule
\textbf{Method} & \textbf{Method-specific hyperparameters} \\
\midrule

OPD
&
None beyond the shared configuration.
\\

\midrule

ExOPD
&
\texttt{exopd\_lambda}=1.25;\quad
\texttt{loss\_max\_clamp}=null;\quad
\texttt{loss\_agg\_mode}=token-mean.
\newline
LoRA: rank=64, alpha=128, target=all-linear.
\\

\midrule

EOPD
&
\texttt{eopd\_entropy\_threshold}=0.8;\quad
\texttt{eopd\_alpha}=1.0;\quad
teacher max logprobs=17.
\\

\midrule

Uni-OPD
&
Batch structure: 16 questions per step, 16 rollouts per question
(256 trajectories per step); mini-batch size 16 questions.
\newline
Loss: \texttt{loss\_agg\_mode}=token-mean;\quad
\texttt{loss\_max\_clamp}=null.
\newline
Margin calibration: \texttt{target\_correct\_ratio}=0.5;\quad
\texttt{margin\_scope}=group;\quad
\texttt{trajectory\_reduce}=mean;
\newline
\texttt{margin\_direction}=spread;\quad
\texttt{margin\_delta}=0.4.
\\

\midrule

Privileged-OPD
&
Max prompt length=12,288 to accommodate reference solutions;
longer solutions are truncated.
Otherwise identical to OPD.
\\

\bottomrule
\end{tabularx}
\caption{
Method-specific hyperparameters. Entries marked ``none'' use the shared configuration unchanged.
}
\label{tab:method_hyperparams}
\end{table}

\paragraph{ExOPD.}
ExOPD extrapolates the teacher-derived reward beyond the teacher's own performance level with a scaling factor $\lambda_{\mathrm{exo}}>1$~\citep{yang2026learning}; we set $\lambda_{\mathrm{exo}}=1.25$. It maintains both the current policy and the pre-RL base model. To avoid storing duplicate weights, we parameterize the student with LoRA adapters (rank 64, alpha 128, all linear layers) and switch between base and policy by disabling or enabling the adapter. We set \texttt{loss\_max\_clamp}=null and \texttt{loss\_agg\_mode}=token-mean.

\paragraph{EOPD.}
EOPD augments the reverse-KL distillation objective with a forward-KL term applied selectively to high-entropy teacher tokens~\citep{jin2026entropy}. The entropy threshold $\tau_{\mathrm{ent}}=0.8$ determines which tokens receive the forward-KL treatment: tokens whose teacher entropy exceeds this threshold are supervised with forward KL to preserve generation diversity, while lower-entropy tokens retain the standard reverse-KL objective. The mixing coefficient $\alpha_{\mathrm{ent}}=1.0$ controls the weight of the forward-KL term relative to the reverse-KL term. To compute teacher entropy, EOPD requires one additional log-probability slot beyond the top-16 used by other methods; we therefore set teacher max logprobs to 17 for EOPD only.

\paragraph{Uni-OPD.}
Uni-OPD introduces a dual-perspective recipe that combines student-side data balancing with teacher-side outcome-guided margin calibration~\citep{hou2026uni}. On the student side, an online correctness-aware filter reshapes each training batch to maintain a target correct-to-incorrect ratio $\rho_{\mathrm{corr}}=0.5$; we implement this by training on 16 questions per step with $n{=}16$ rollouts per question, yielding 256 trajectories per step, and applying sample filtering to achieve the target ratio. The mini-batch size is correspondingly set to 16 questions. On the teacher side, Uni-OPD calibrates token-level teacher margins against trajectory-level outcome rewards. The margin scope is set to \texttt{group}, meaning calibration is performed within each group of rollouts for the same question. The trajectory reduction is \texttt{mean}, the margin direction is \texttt{spread} (which spreads apart the margins of correct and incorrect trajectories), and the target margin is $\delta_{\mathrm{margin}}=0.4$. The loss aggregation mode is token-mean, and the loss is not clamped.

\paragraph{Privileged-OPD.}
Privileged-OPD conditions the teacher on additional training-time information, specifically the verified reference solution for each problem~\citep{ye2026policy,kaur2026rethinking}. Because reference solutions are substantially longer than the problem statements, we increase the maximum prompt length from the shared default to 12,288 tokens to accommodate the concatenation of problem, reference solution, and student rollout prefix. Solutions exceeding this length are truncated. All other settings follow the standard OPD configuration. In our implementation, the privileged context is prepended to the problem and student prefix in the teacher's input, while the student receives only the problem and its own rollout, preserving the information asymmetry that Privileged-OPD exploits.

\paragraph{Cal-OPD.}
Cal-OPD follows the shared configuration above. It additionally uses the evaluative feedback interventions $c_{\mathrm{eval}}^{\mathrm{pos}}$ and $c_{\mathrm{eval}}^{\mathrm{neg}}$ to probe the teacher's self-deviation region, with relaxation factor $\lambda=5$. These interventions are used only to estimate the TSD region and are not directly distilled into the student. The distillation loss then operates on the calibrated advantage $A_t^{\mathrm{Cal}}$ in place of the raw $A_t^{\mathrm{OPD}}$, as described in Section~\ref{sec:cal_opd}.

\subsection{Additional Training Dynamics}
\label{app:training_dynamics}

\begin{figure}[t]
    \centering

    \begin{minipage}[t]{0.48\linewidth}
        \centering
        \includegraphics[width=\linewidth]{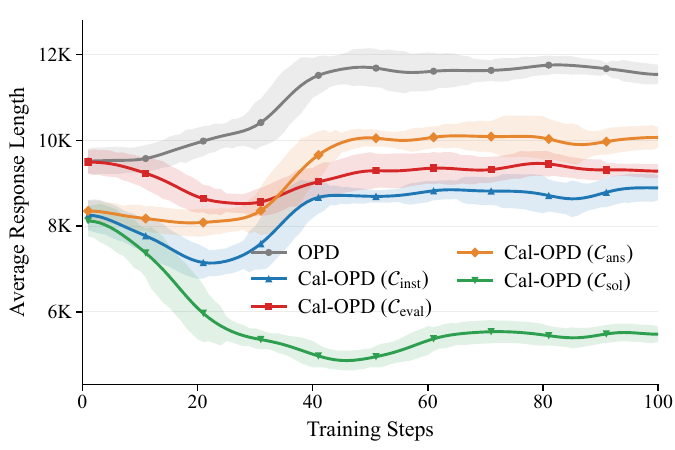}
        \small (a) Average Response Length
    \end{minipage}
    \hfill
    \begin{minipage}[t]{0.48\linewidth}
        \centering
        \includegraphics[width=\linewidth]{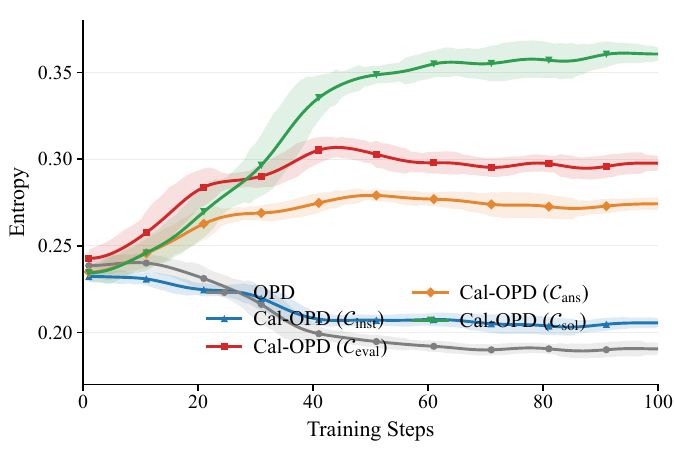}
        \small (b) Entropy
    \end{minipage}

    \vspace{0.5em}

    \begin{minipage}[t]{0.48\linewidth}
        \centering
        \includegraphics[width=\linewidth]{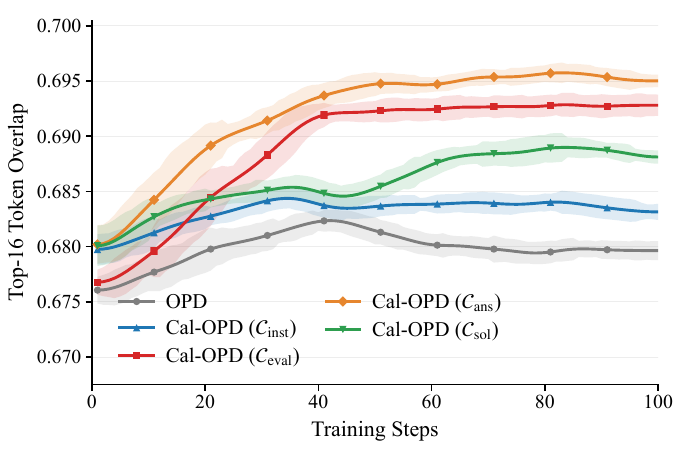}
        \small (c) Top-16 Token Overlap
    \end{minipage}
    \hfill
    \begin{minipage}[t]{0.48\linewidth}
        \centering
        \includegraphics[width=\linewidth]{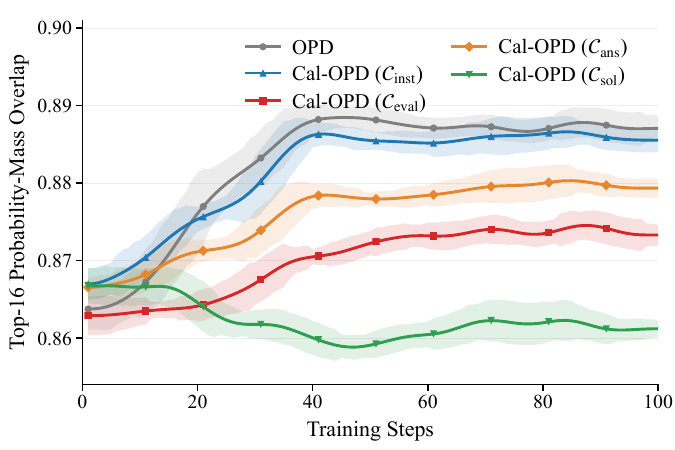}
        \small (d) Top-16 Probability-Mass Overlap
    \end{minipage}

    \caption{Training dynamics of OPD and Cal-OPD on Qwen3-1.7B.}
    \label{fig:additional_training_dynamics}
\end{figure}

\paragraph{OPD vs.\ Cal-OPD with evaluative calibration.}
Figure~\ref{fig:additional_training_dynamics} reveals qualitatively different training dynamics between OPD and Cal-OPD with $\mathcal{C}_{\mathrm{eval}}$. OPD progressively increases response length while reducing student entropy, whereas Cal-OPD maintains substantially shorter responses and higher entropy throughout training. This contrast is also reflected in how each method aligns with the teacher's top-16 distribution. Let $H$ denote the student entropy, $\mathrm{Top16Tok}$ the top-16 token overlap, and $\mathrm{Top16Mass}$ the top-16 probability-mass overlap. The dynamics can be summarized as
\begin{equation}
\begin{aligned}
H_{\mathrm{Cal\text{-}OPD}} &> H_{\mathrm{OPD}}, \\
\mathrm{Top16Tok}_{\mathrm{Cal\text{-}OPD}} &> \mathrm{Top16Tok}_{\mathrm{OPD}}, \\
\mathrm{Top16Mass}_{\mathrm{Cal\text{-}OPD}} &< \mathrm{Top16Mass}_{\mathrm{OPD}}.
\end{aligned}
\end{equation}
In particular, the top-16 token overlap under Cal-OPD steadily increases to approximately $0.693$, whereas OPD peaks earlier and then slightly declines. Meanwhile, OPD achieves a higher top-16 probability-mass overlap, indicating stronger matching of the teacher's probability allocation.

\textbf{Together, these dynamics suggest that OPD increasingly fits the teacher's dominant probability pattern, whereas Cal-OPD preserves a broader student distribution while covering more teacher-supported token patterns, enabling the student to capture multiple plausible reasoning modes rather than overfitting to a particular one.}

\paragraph{Cal-OPD with instruction-level calibration.}
In contrast to evaluative calibration, instruction-level calibration with $\mathcal{C}_{\mathrm{inst}}$ yields dynamics that \textbf{closely resemble standard OPD}, indicating that its calibration signal is too weak to meaningfully separate TSD from transferable knowledge. As shown in Figure~\ref{fig:additional_training_dynamics}, the student entropy under $\mathcal{C}_{\mathrm{inst}}$ follows the same declining trend as OPD and stabilizes around $0.20$, whereas the other Cal-OPD variants maintain substantially higher entropy. Its top-16 probability-mass overlap is also nearly identical to that of OPD, stabilizing around $0.886$, which shows that it continues to match the teacher's dominant probability allocation as strongly as the uncalibrated objective. Similarly, its top-16 token overlap remains low at approximately $0.684$, again close to OPD and well below the levels reached by $\mathcal{C}_{\mathrm{eval}}$, $\mathcal{C}_{\mathrm{ans}}$, and $\mathcal{C}_{\mathrm{sol}}$. These trends indicate that, without task-specific information or strong semantic contrast, the estimated TSD region is too narrow to filter the teacher--student discrepancy effectively, and Cal-OPD largely degenerates to standard OPD behavior.

\paragraph{Cal-OPD with solution-level calibration.}
Solution-level calibration with $\mathcal{C}_{\mathrm{sol}}$ lies at the opposite extreme: it over-filters the teacher--student discrepancy and causes the supervision signal to collapse. As shown in Figure~\ref{fig:additional_training_dynamics}, the student entropy under $\mathcal{C}_{\mathrm{sol}}$ is the highest among all methods and continues to rise throughout training, stabilizing near $0.36$, while its average response length is the lowest and steadily decreases to roughly $5.5$K tokens. More strikingly, its top-16 probability-mass overlap is not only the lowest among all variants but also falls below its initial value, stabilizing around $0.862$. Its top-16 token overlap remains moderate at approximately $0.688$, below $\mathcal{C}_{\mathrm{eval}}$ and $\mathcal{C}_{\mathrm{ans}}$, but above OPD. These trends indicate that the TSD region estimated from solution-level privilege is excessively broad, so that a large fraction of the teacher--student discrepancy is classified as TSD and zeroed out. As a result, the student receives too few effective supervision signals to follow the teacher distribution, and instead drifts toward a high-entropy, short-response regime. \textbf{Over-calibration therefore degrades the signal more severely than no calibration at all}, which is consistent with its lowest downstream performance among all variants.

\begin{figure}[t]
    \centering

    \begin{minipage}[t]{0.32\linewidth}
        \centering
        \includegraphics[width=\linewidth]{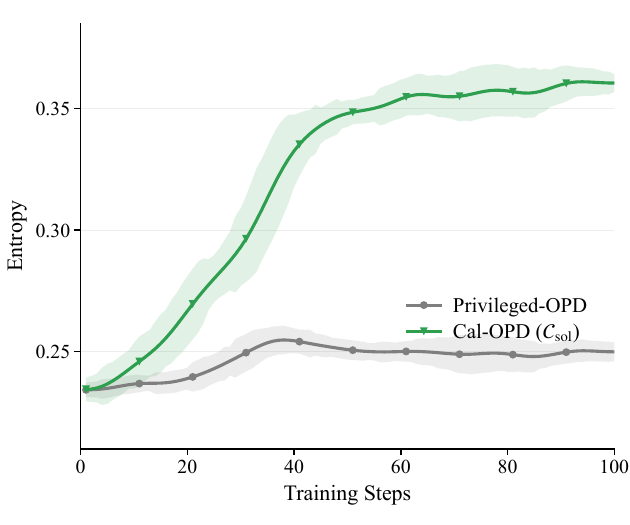}
        \small (a) Entropy
    \end{minipage}
    \hfill
    \begin{minipage}[t]{0.32\linewidth}
        \centering
        \includegraphics[width=\linewidth]{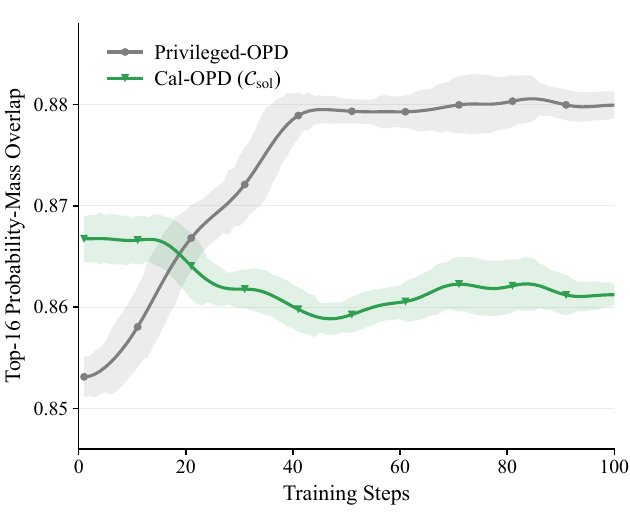}
        \small (b) Top-16 Probability-Mass Overlap
    \end{minipage}
    \hfill
    \begin{minipage}[t]{0.32\linewidth}
        \centering
        \includegraphics[width=\linewidth]{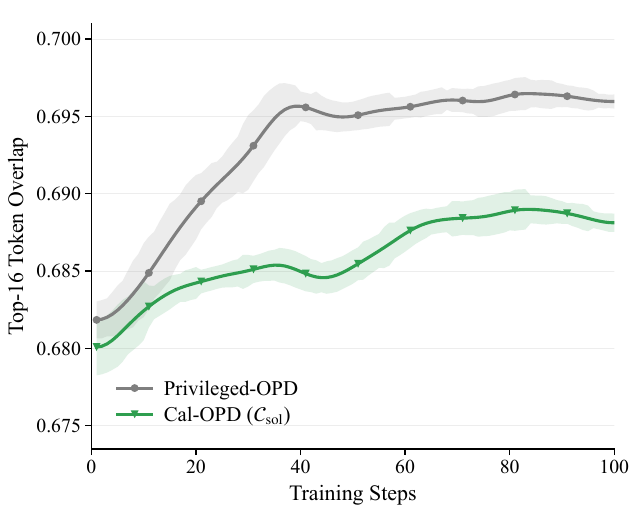}
        \small (c) Top-16 Token Overlap
    \end{minipage}

    \caption{Training dynamics comparison between privileged OPD and Cal-OPD with solution-level calibration.}
    \label{fig:privileged_vs_cal_sol_dynamics}
\end{figure}

\paragraph{Failure Modes of Privileged-OPD and Cal-OPD with Solution-Level Calibration.}
Both methods shorten responses and degrade performance, but for different reasons. All top-16 metrics are computed against the unprivileged teacher. Privileged-OPD keeps student entropy low and stable (around $0.25$) while its top-16 probability-mass overlap rises to about $0.880$, indicating that the student becomes overconfident and concentrates on a narrow token set. In contrast, Cal-OPD with $\mathcal{C}_{\mathrm{sol}}$ drives entropy up to about $0.36$, and its top-16 probability-mass overlap even drops to roughly $0.859$, showing that it fails to match the unprivileged teacher. We attribute this to two distinct failure modes: Privileged-OPD suffers from \textbf{overconfident shortcut collapse}, where the student imitates the privileged teacher's sharp distribution but loses diversity; Cal-OPD with $\mathcal{C}_{\mathrm{sol}}$ suffers from \textbf{signal collapse}, where an excessively broad TSD region zeros out useful supervision, causing the student to drift into a high-entropy, short-response regime.

\subsection{Efficiency Analysis}
\label{app:efficiency}

\begin{table}[t]
\centering
\small
\setlength{\tabcolsep}{6pt}
\renewcommand{\arraystretch}{1.10}
\begin{tabular}{@{}llcc@{}}
\toprule
\textbf{Method} & \textbf{Transfer} & \textbf{Avg. time per step (min)} & \textbf{Total time for 100 steps (h)} \\
\midrule
Cal-OPD & 30B-A3B-2507 $\rightarrow$ 8B   & 50.05 & $\sim$83.4 \\
OPD     & 30B-A3B-2507 $\rightarrow$ 8B   & 35.19 & $\sim$58.7 \\
\midrule
Cal-OPD & 30B-A3B-2507 $\rightarrow$ 4B   & 48.76 & $\sim$81.3 \\
OPD     & 30B-A3B-2507 $\rightarrow$ 4B   & 31.26 & $\sim$52.1 \\
\midrule
Cal-OPD & 4B-2507 $\rightarrow$ 1.7B      & 14.90 & $\sim$24.8 \\
OPD     & 4B-2507 $\rightarrow$ 1.7B      & 18.80 & $\sim$31.3 \\
\bottomrule
\end{tabular}
\caption{
Efficiency comparison between Cal-OPD and OPD across different teacher--student configurations.
}
\label{tab:efficiency}
\end{table}

\paragraph{Efficiency of Cal-OPD.}
Cal-OPD requires two additional teacher forward passes per step to probe the TSD region, which introduces extra computation. Table~\ref{tab:efficiency} compares the wall-clock time per step and the total time for 100 steps between Cal-OPD and OPD across three teacher--student configurations. For the Qwen3-4B-Thinking-2507$\rightarrow$Qwen3-1.7B setting, Cal-OPD is actually faster than OPD (14.90 vs.\ 18.80 minutes per step), because Cal-OPD avoids the response-length expansion that OPD exhibits, and the teacher is not large enough for the extra forward passes to dominate. In the other two configurations, where the teacher (30B-A3B) is substantially larger than the student (8B or 4B), the two additional teacher forward passes become the main computational overhead, making Cal-OPD slower per step (50.05 vs.\ 35.19 and 48.76 vs.\ 31.26 minutes). Overall, Cal-OPD remains practically efficient: its extra cost is bounded by two teacher forward passes, and it can even reduce total training time when the teacher--student size gap is moderate.

\subsection{Retention-Matched Control on Teacher--Student Discrepancy}
\label{app:retention_matched_control}

\begin{figure}[t]
    \centering

    \begin{minipage}[t]{0.48\linewidth}
        \centering
        \includegraphics[width=\linewidth]{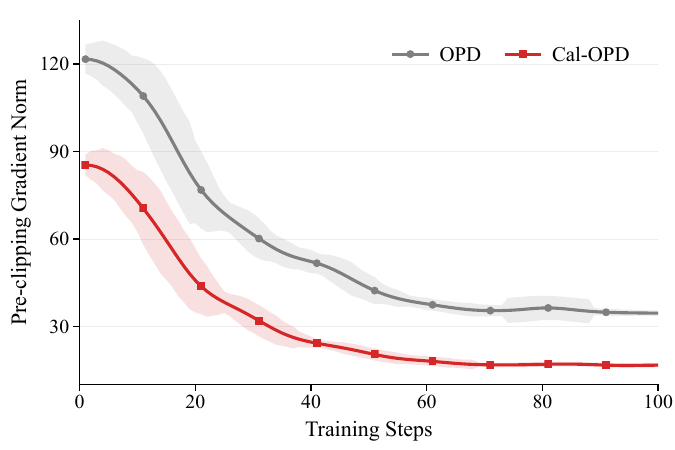}
        \small (a) Qwen3-4B-Thinking-2507 $\rightarrow$ Qwen3-1.7B
    \end{minipage}
    \hfill
    \begin{minipage}[t]{0.48\linewidth}
        \centering
        \includegraphics[width=\linewidth]{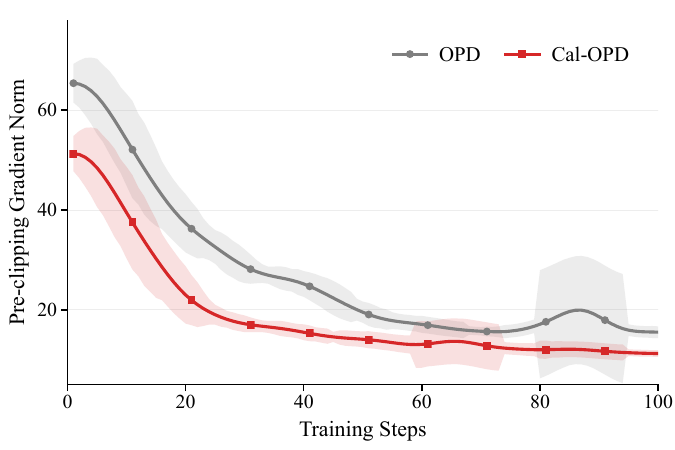}
        \small (b) Qwen3-30B-A3B-Thinking-2507 $\rightarrow$ Qwen3-4B
    \end{minipage}

    \caption{Gradient norm dynamics of OPD training across different teacher--student pairs.}
    \label{fig:grad_norm_dynamics}
\end{figure}

Figure~\ref{fig:grad_norm_dynamics} reports the pre-clipping gradient norm during OPD training. In both teacher--student configurations, the norm is initially large, around $120$ and $65$, and gradually decreases to approximately $35$ and $15$. All methods apply gradient clipping with threshold $C=1$, so the update is computed as
\begin{equation}
\tilde{\mathbf{g}} = \mathbf{g} \cdot \min\!\left(1, \frac{C}{\|\mathbf{g}\|}\right).
\end{equation}
Since $\|\mathbf{g}\|$ is far above $1$ throughout training, the raw gradient is scaled by roughly $C/\|\mathbf{g}\|$, corresponding to an attenuation of one to two orders of magnitude. \textbf{Thus, substantial signal attenuation is already inherent to standard OPD, and the gains of Cal-OPD cannot be attributed to a smaller overall gradient magnitude. Cal-OPD instead changes the \emph{relative} contribution of individual token advantages by removing the TSD-explained component, while leaving the global clipping behavior unchanged.}

To rule out the possibility that Cal-OPD gains merely from attenuating the optimization signal, we construct three retention-matched baselines that mimic its signal reduction without TSD calibration.

\begin{itemize}
    \item \textbf{Advantage-Sync (Advantage-Level Synchronization).} At each training step $k$, we read Cal-OPD's advantage retention ratio $r_k$, defined as the ratio between the sum of absolute calibrated advantages and the sum of absolute original teacher--student discrepancies. We then scale standard OPD's token-level advantages by the same factor, $\tilde{A}_t^{\mathrm{OPD}} = r_k \cdot A_t^{\mathrm{OPD}}$, so that the global advantage magnitude matches Cal-OPD step by step.
    
    \item \textbf{Token-Sync (Token-Level Synchronization).} At each training step $k$, we read Cal-OPD's zero-advantage token ratio $z_k$, i.e., the fraction of tokens whose calibrated advantage is zero. We then randomly select the same fraction $z_k$ of tokens in standard OPD and set their advantages to zero, keeping the remaining advantages unchanged. This matches Cal-OPD's sparsity pattern in terms of the number of discarded tokens, but not which tokens are discarded.
    
    \item \textbf{TSD-Filter (TSD-Threshold Filtering).} For each token, we compute the maximum absolute deviation induced by the positive and negative privileged interventions, and discard the token entirely if this deviation exceeds a threshold $\tau_{\mathrm{TSD}}$. We evaluate three thresholds, $\tau_{\mathrm{TSD}} \in \{0.1, 0.05, 0.01\}$, corresponding to increasingly aggressive filtering. Unlike Cal-OPD, which retains the residual discrepancy beyond the estimated TSD region, this baseline removes the token from the loss altogether.
\end{itemize}

These controls match Cal-OPD in global advantage scale, token sparsity, and TSD-based token selection, respectively, but omit the calibrated residual.

\begin{table}[t]
\caption{
Retention-matched control on teacher--student discrepancy for
Qwen3-4B-Thinking-2507$\rightarrow$Qwen3-1.7B.
}
\label{tab:retention_matched_control}

\centering
\small
\setlength{\tabcolsep}{2.6pt}
\renewcommand{\arraystretch}{1.08}

\begin{tabularx}{\linewidth}{
@{}
>{\raggedright\arraybackslash}p{0.27\linewidth}
>{\centering\arraybackslash}X
>{\centering\arraybackslash}X
>{\centering\arraybackslash}X
>{\centering\arraybackslash}X
>{\centering\arraybackslash}X
>{\centering\arraybackslash}X
>{\centering\arraybackslash}X
@{}
}
\toprule

\textbf{Method}
& \textbf{AMC23}
& \textbf{AIME24}
& \textbf{AIME25}
& \textbf{AIME26}
& \textbf{HMMT26}
& \textbf{MATH500}
& \textbf{Avg.}
\\

\midrule

OPD
& 82.0 & 39.6 & 35.0 & 34.6 & 22.9 & 90.6 & 50.8 \\

Advantage-Sync
& 81.7 & 39.8 & 35.4 & 35.0 & 22.2 & 90.5 & 50.8 \\

Token-Sync
& 81.3 & 38.3 & 34.6 & 34.0 & 22.3 & 89.8 & 50.1 \\

TSD-Filter ($\tau_{\mathrm{TSD}}=0.1$)
& 83.6 & 44.0 & 37.1 & 36.0 & 24.4 & 90.9 & 52.7 \\

TSD-Filter ($\tau_{\mathrm{TSD}}=0.05$)
& 81.7 & 41.7 & 35.4 & 34.8 & 22.9 & 90.1 & 51.1 \\

TSD-Filter ($\tau_{\mathrm{TSD}}=0.01$)
& 80.3 & 38.8 & 33.5 & 32.9 & 21.6 & 89.3 & 49.4 \\

Cal-OPD
& 84.5 & 45.0 & 37.1 & 36.3 & 24.8 & 90.8 & \textbf{53.1} \\

\bottomrule
\end{tabularx}
\end{table}

Table~\ref{tab:retention_matched_control} compares Cal-OPD against three retention-matched controls. \textbf{Advantage-Sync} applies the same global scaling to OPD's advantages and achieves $50.8$, identical to OPD, showing that reducing the overall advantage magnitude alone brings no benefit. \textbf{Token-Sync} randomly masks the same fraction of tokens and even drops to $50.1$, indicating that matching the sparsity level without selecting the right tokens is insufficient. \textbf{TSD-Filter} with $\tau_{\mathrm{TSD}}=0.1$ reaches $52.7$, the strongest control, but still falls short of Cal-OPD; more aggressive thresholds of $0.05$ and $0.01$ degrade to $51.1$ and $49.4$, respectively. These results show that neither global attenuation, nor random token removal, nor hard TSD-based filtering can reproduce the gains of Cal-OPD. \textbf{The advantage of Cal-OPD therefore stems from its soft calibration mechanism, which retains the discrepancy beyond the estimated TSD region, rather than from signal attenuation or token sparsity alone.}

\end{document}